\documentclass[10pt,journal,compsoc]{IEEEtran}
\ifCLASSOPTIONcompsoc
  \usepackage[nocompress]{cite}
\else
  \usepackage{cite}
\fi
\ifCLASSINFOpdf
\else
\fi

\usepackage{times}
\usepackage{epsfig}
\usepackage{graphicx}
\usepackage{amsmath}
\usepackage{amssymb}
\usepackage{mathtools}
\usepackage{booktabs}
\usepackage{xcolor}
\usepackage{multirow}
\usepackage{float}
\usepackage{amsfonts}
\usepackage{comment}
\usepackage{url}
\usepackage[caption=false]{subfig}
\usepackage{wrapfig}
\newcommand{\parsection}[1]{\vspace{1mm}\noindent\textbf{#1:}~}
\usepackage[pagebackref=true,breaklinks=true,colorlinks,bookmarks=false]{hyperref}

\usepackage{xspace}
\makeatletter
\DeclareRobustCommand\onedot{\futurelet\@let@token\@onedot}
\def\@onedot{\ifx\@let@token.\else.\null\fi\xspace}
\def\eg{\emph{e.g}\onedot} 
\def\ie{\emph{i.e}\onedot}

\def\etal{\emph{et al}\onedot}

\newcommand{\reals}{\mathbb{R}}
\newcommand{\tp}{^\text{T}}

\begin{document}
%
\title{PDC-Net+: Enhanced Probabilistic Dense Correspondence Network}
%
%
%
%

\author{Prune Truong,
        Martin Danelljan, Radu Timofte
        and Luc Van Gool
\IEEEcompsocitemizethanks{\IEEEcompsocthanksitem All authors are with the Computer Vision Lab of ETH Zurich.\protect\\
E-mails: \{prune.truong, martin.danelljan, vangool, radu.timofte\}\protect\\@vision.ee.ethz.ch}
}
\IEEEtitleabstractindextext{%
\begin{abstract}
Establishing robust and accurate correspondences between a pair of images is a long-standing computer vision problem with numerous applications. While classically dominated by sparse methods, emerging dense approaches offer a compelling alternative paradigm that avoids the keypoint detection step.   
However, dense flow estimation is often inaccurate in the case of large displacements, occlusions, or homogeneous regions. In order to apply dense methods to real-world applications, such as pose estimation, image manipulation, or 3D reconstruction, it is therefore crucial to estimate the confidence of the predicted matches. 

We propose the Enhanced Probabilistic Dense Correspondence Network, PDC-Net+, capable of estimating accurate dense correspondences along with a reliable confidence map.
We develop a flexible probabilistic approach that jointly learns the flow prediction and its uncertainty. In particular, we parametrize the predictive distribution as a constrained mixture model, ensuring better modelling of both accurate flow predictions and outliers. Moreover, we develop an architecture and an enhanced training strategy tailored for robust and generalizable uncertainty prediction in the context of self-supervised training. 
Our approach obtains state-of-the-art results on multiple challenging geometric matching and optical flow datasets. We further validate the usefulness of our probabilistic confidence estimation for the tasks of pose estimation, 3D reconstruction, image-based localization, and image retrieval. Code and models are available at \url{https://github.com/PruneTruong/DenseMatching}.

\end{abstract}

}

\maketitle

\IEEEdisplaynontitleabstractindextext

%
\IEEEpeerreviewmaketitle

\IEEEraisesectionheading{\section{Introduction}\label{intro}}

Finding correspondences between pairs of images is a fundamental computer vision problem with numerous applications, including image alignment~\cite{BrownL07, GLAMpoint, shrivastava-sa11}, video analysis~\cite{SimonyanZ14}, image manipulation~\cite{HaCohenSGL11, LiuYT11}, Structure-from-Motion (SfM)~\cite{Wu13, SchonbergerF16}, and Simultaneous Localization and Mapping (SLAM)~\cite{EngelKC18}. 
Correspondence estimation has traditionally been dominated by sparse approaches~\cite{SIFT, SURF, Brief, ORB, Belousov2017, superpoint, Dusmanu2019CVPR, OnoTFY18, R2D2descriptor, DELF}, which first detect local keypoints in salient regions that are then matched. However, recent years have seen a growing interest in dense methods~\cite{Melekhov2019, Rocco2018a, GLUNet}. 
By predicting a match for every single pixel in the image, these methods open the door to additional applications, such as texture or style transfer~\cite{Kim2019, Liao2017}.
Moreover, dense methods do not require detection of salient and repeatable keypoints, which itself is a challenging problem.

Dense correspondence estimation has most commonly been addressed in the context of optical flow~\cite{Baker2011, Horn1981, Hur2020OpticalFE,RAFT}, where the image pairs represent consecutive frames in a video. While these methods excel in the case of small appearance changes and limited displacements, they cannot cope with the challenges posed by the more general geometric matching task. In geometric matching, the images can stem from radically different views of the same scene, often captured by different cameras and at different occasions.
This leads to large displacements and significant appearance transformations between the frames. 
In contrast to optical flow, the more general dense correspondence problem has received much less attention~\cite{Melekhov2019,Rocco2018b,RANSAC-flow, GLUNet, pdcnet}. 

\begin{figure*}[t]
\centering%
\newcommand{\wid}{0.89\textwidth}%
\begin{tabular}{c}
\small{
    \hspace{1.3cm} (a) Query image \hspace{1.7cm} (b) Reference image \hspace{1.7cm} (c) Baseline  \hspace{1.9cm} (d) \textbf{PDC-Net+} (Ours) \hspace{2cm}}
\end{tabular}
\includegraphics[width=\wid]{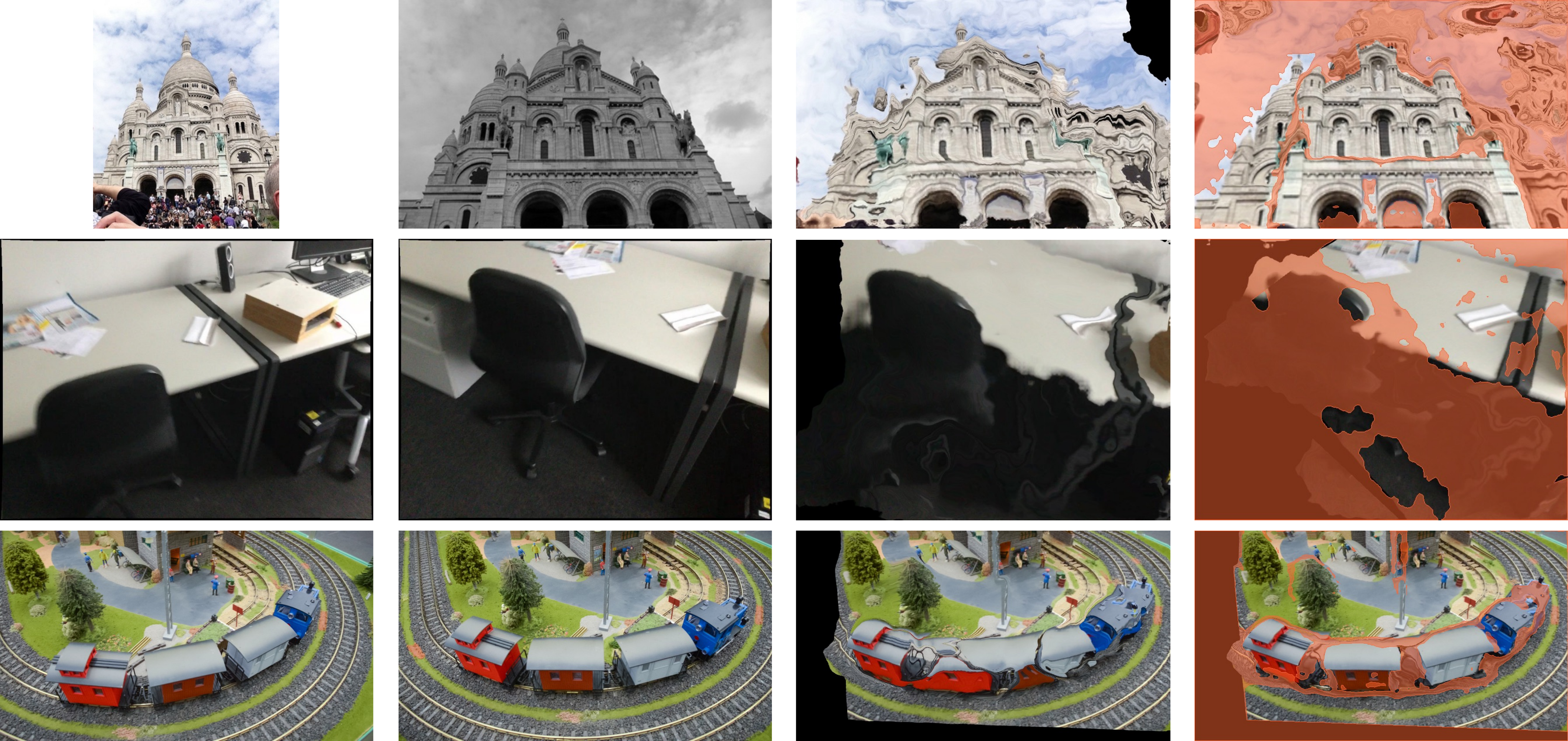}
\vspace{-3mm}\caption{ 
Estimating dense correspondences between the query (a) and the reference (b) image. The query is warped according to the predicted flows (c)-(d).
The baseline (c) does not estimate an uncertainty map and is therefore unable to filter out the inaccurate predictions at \eg occluded and homogeneous regions. In contrast, our PDC-Net+ (d) not only learns more accurate correspondences, but also when to trust them. It predicts a robust uncertainty map that identifies accurate matches and excludes incorrect and unmatched pixels (red).
}\vspace{-2mm}
\label{fig:intro}
\end{figure*}

Dense flow estimation is prone to errors in the presence of large displacements, appearance changes, or homogeneous regions. It is also ill-defined in case of occlusions or in \eg the sky, where predictions are bound to be inaccurate (Fig.~\ref{fig:intro}c). For geometric matching applications, it is thus crucial to know when and where to trust the estimated correspondences. 
The identification of inaccurate or incorrect matches is particularly important in, for instance, dense 3D reconstruction~\cite{SchonbergerF16}, high quality image alignment~\cite{shrivastava-sa11,GLAMpoint}, and multi-frame image restoration~\cite{burstsr}.
Moreover, dense confidence estimation bridges the gap between the application domains of the dense and sparse correspondence estimation paradigms. It enables the selection of robust and accurate matches from the dense output, to be utilized in, \eg, pose estimation and image-based localization. 
Uncertainty estimation is also indispensable for safety-critical tasks, such as autonomous driving and medical imaging. 
In this work, we set out to widen the application domain of dense correspondence estimation by learning to predict reliable confidence values (Fig.~\ref{fig:intro}d).

We propose the Enhanced Probabilistic Dense Correspondence Network, PDC-Net+, for joint learning of dense flow estimation along with its uncertainties. It is applicable even for extreme appearance and view-point changes, often encountered in geometric matching scenarios. 
Our model learns to predict the conditional probability density of the dense flow between two images. In order to accurately capture the uncertainty of inlier and outlier flow predictions, we introduce a \emph{constrained mixture model}. In contrast to predicting a single variance, our formulation allows the network to directly assess the probability of a match being an inlier or outlier. By constraining the variances, we further enforce the components to focus on separate uncertainty intervals, which effectively resolves the ambiguity caused by the permutation invariance of the mixture model components. 

Learning reliable and generalizable uncertainties without densely annotated real-world training data is a highly challenging problem. 
We tackle this issue from the architecture and the data perspective in the context of self-supervised training. 
Directly predicting the uncertainties using the flow decoder leads to highly over-confident predictions in, for instance, texture-less regions of real scenes. This stems from the network's ability to extrapolate neighboring matches during self-supervised training. We alleviate this problem by introducing an architecture that processes each spatial location of the correlation volume independently, leading to robust and generalizable uncertainty estimates.
We also revisit the data generation problem for self-supervised learning. We find that current strategies yield too predictable flow fields, which leads to inaccurate uncertainty estimates on real data. To this end, we introduce random perturbations in the synthetic ground-truth flow fields. 
Since our strategy encourages the network to focus on the local appearance rather than simplistic smoothness priors, it improves the uncertainty prediction in, for example, homogeneous regions.

To better simulate moving objects and occlusions encountered in real scenes, we further improve upon the self-supervised data generation pipeline by iteratively adding multiple independently moving objects onto a base image pair. 
However, since the base image pair is related by a simple background transformation, the network tends to primarily focus on its flow, at the expense of the object motion.
To better encourage the network to learn the more challenging object flows, we introduce an injective criterion for masking out regions from the objective. 
Our approach only masks out occluded regions that violate a one-to-one ground-truth mapping. This allows the network to focus on flow estimation of \emph{visible} moving objects as opposed to occluded background regions, while simultaneously learning vital interpolation and extrapolation capabilities.

Our final PDC-Net+ approach generates a predictive distribution, from which we extract the mean flow field and its corresponding confidence map. To ensure its versatility in different domains and applications, we design multiple inference strategies. In particular, we utilize our confidence estimation to further improve the final flow prediction in scenarios with extreme view-point changes, by proposing both a multi-stage and a multi-scale approach. 
We apply PDC-Net+ to a variety of tasks and datasets. Our approach sets a new state-of-the-art on the Megadepth~\cite{megadepth} and the RobotCar~\cite{RobotCar} geometric matching datasets.
Without any task-specific fine-tuning, PDC-Net+ generalizes to the optical flow domain by outperforming recent state-of-the-art methods~\cite{RAFT} on the KITTI-2015 training set~\cite{Geiger2013}.
We further apply our approach to pose estimation on both the outdoor YFCC100M~\cite{YFCC} and the indoor ScanNet~\cite{scannet} datasets, outperforming previous dense methods and ultimately closing the gap to state-of-the-art sparse approaches.
We also validate our method for image-based localization and dense 3D reconstruction on the Aachen dataset~\cite{SattlerMTTHSSOP18, SattlerWLK12}. Lastly, we demonstrate that the confidence estimation provided by PDC-Net+ can be directly used for robust image retrieval. 

The rest of the manuscript is organized as follows. We review related work in Section~\ref{sec:related-work}. Section~\ref{sec:method} introduces our approach, PDC-Net+. Extensive experimental results are presented in Section~\ref{sec:exp}. Finally, we discuss conclusions and future work in Section~\ref{sec:conclusion}.

\section{Related work}
\label{sec:related-work}

\subsection{Correspondence estimation}

\parsection{Sparse matching} Sparse methods generally consist of three stages: keypoint detection, feature description, and feature matching. Keypoint detectors and descriptors are either hand-crafted~\cite{SIFT, SURF, Brief, ORB} or learned~\cite{GLAMpoint, Belousov2017, superpoint, Dusmanu2019CVPR, OnoTFY18, R2D2descriptor, DELF}. 
Feature matching is generally treated as a separate stage, where descriptors are matched exhaustively. This is followed by heuristics, such as the ratio test~\cite{SIFT}, or robust matchers and filtering methods. The filtering methods are either hand-crafted~\cite{abs-1803-07469, ransac} or learned~\cite{BrachmannR19, YiTOLSF18, OANet, SarlinDMR20}. Our approach instead is learned end-to-end and directly predicts dense correspondences from an image pair, without the need for detecting local keypoints.

\parsection{Sparse-to-Dense matching} While sparse methods have yielded very competitive results, they rely on the detection of stable and repeatable keypoints across images, which is a highly challenging problem. 
Recently, Germain~\etal~\cite{GermainBL19} proposed to transform the sparse-to-sparse paradigm into a sparse-to-dense approach. Instead of trying to detect repeatable feature points across images, feature detection is performed asymmetrically and correspondences are searched exhaustively in the other image. The more recent work S2DNet~\cite{GermainBL20} casts the correspondence learning problem as a supervised classification task and learns multi-scale feature maps.

\parsection{Dense-to-sparse matching} These approaches start from a correlation volume, that densely matches feature vectors between two feature maps at low resolution. These matches are then sparsified and processed at higher resolution to obtain a final set of refined matches. In SparseNC-Net, Rocco~\etal~\cite{Rocco20} sparsify the 4D correlation by projecting it onto a sub-manifold. Dual-RC-Net~\cite{DualRCNet} and XRC-Net~\cite{Xreo} instead use a coarse-to-fine re-weighting mechanism to guide the search for the best match in a fine resolution correlation map. Recently, Sun~\etal~\cite{LOFTR} introduced LoFTR, a Transformer-based architecture which also first establishes pixel-wise dense matches at a coarse level and later refines the good matches at a finer scale.

\parsection{Dense matching} Dense methods instead directly predict dense matches. 
Rocco~\etal~\cite{Rocco2018b} rely on a correlation layer to perform the matching and further propose an end-to-end trainable neighborhood consensus network, NC-Net. 
However, it is very memory expensive, which makes it difficult to scale up to higher image resolutions, and thus limits the accuracy of the resulting matches. 
Similarly, Wiles~\etal~\cite{D2D} learn dense descriptors conditioned on an image pair, which are then matched with a feature correlation layer. Other related approaches seek to learn correspondences on a semantic level, between instances of the same class~\cite{Rocco2017a, Rocco2018a, ArbiconNet, SeoLJHC18, GLUNet, MinLPC20, Kim2019}.

Most related to our work are dense flow regression methods, that predict a dense correspondence map or flow field relating an image pair. 
While dense regression methods were originally designed for the optical flow task~\cite{Dosovitskiy2015, Ilg2017a, Hui2018, Hui2019, Sun2018, Sun2019, RAFT, DDFlow, SefFlow, ARFlow}, recent works have extended such approaches to the geometric matching scenario, in order to handle large geometric and appearance transformations. 
Melekhov~\etal~\cite{Melekhov2019} introduced DGC-Net, a coarse-to-fine Convolutional Neural Network (CNN)-based framework that generates dense correspondences between image pairs. It relies on a global cost volume constructed at the coarsest resolution. However, this method restricts the network input resolution to a fixed low dimension.
Truong~\etal~\cite{GLUNet} proposed GLU-Net to learn dense correspondences without such a constraint on the input resolution, by integrating both global and local correlation layers. The authors further introduced GOCor~\cite{GOCor}, an online optimization-based matching module acting as a direct replacement to the feature correlation layer. It significantly improves the accuracy and robustness of the predicted dense correspondences.
Shen~\etal~\cite{RANSAC-flow} proposed RANSAC-Flow, a two-stage image alignment method. It performs coarse alignment with multiple homographies using RANSAC on off-the-shelf deep features, followed by a fine-grained alignment. Recently, Huang~\etal~\cite{LIFE} adopt the RAFT architecture~\cite{RAFT}, originally designed for optical flow, and train it for pairs with large lighting variations in a weakly-supervised framework based on epipolar constraints. COTR~\cite{COTR} uses a Transformer-based architecture to retrieve matches at any queried locations, which are later densified. However, the process is computationally costly, making it unpractical for many applications. 
In contrast to these works, we propose a unified network that estimates the flow field along with probabilistic uncertainties.

\subsection{Confidence estimation}

\parsection{Confidence estimation in geometric matching}
Only very few works have explored confidence estimation in the context of dense geometric or semantic matching. 
Novotny~\etal~\cite{NovotnyCVPR18Self} estimate the reliability of their trained descriptors by using a self-supervised probabilistic matching loss for the task of semantic matching.
A few approaches~\cite{DCCNet, Rocco20, Rocco2018b, DualRCNet, Xreo} represent the final correspondences as a 4D correspondence volume, thus inherently encoding a confidence score for each tentative match. However, generating one final reliable confidence value for each match is difficult since multiple high-scoring alternatives often co-occur.
Similarly, Wiles~\etal~\cite{D2D} predict a distinctiveness score along with learned descriptors. However, it is trained with hand-crafted heuristics. In contrast, we do not need to generate annotations to train our uncertainty prediction nor to make assumptions on what it should capture. Instead we learn it solely from correspondence ground-truth in a probabilistic manner. 
In DGC-Net, Melekhov~\etal~\cite{Melekhov2019} predict both dense correspondences and a matchability map. However, the matchability map is only trained to identify out-of-view pixels rather than to reflect the actual reliability of the matches. Recently, RANSAC-Flow~\cite{RANSAC-flow} also learns a matchability mask using a combination of losses.
In contrast, we introduce a probabilistic formulation that is learned with a single unified loss -- the negative log-likelihood.

\parsection{Uncertainty estimation in optical flow} 
While optical flow has been a long-standing subject of active research, only a handful of methods provide uncertainty estimates. 
A few approaches~\cite{Aodha2013LearningAC, Barron94performanceof, KondermannKJG07, KondermannMG08, KybicN11} treat the uncertainty estimation as a post-processing step. 
Recently, some works propose probabilistic frameworks for joint optical flow and uncertainty prediction. They either estimate the model uncertainty~\cite{GalG15, IlgCGKMHB18}, also known as epistemic uncertainty~\cite{KendallG17}, or focus on the uncertainty from the observation itself, referred to as aleatoric uncertainty~\cite{KendallG17}. 
Following recent works~\cite{Gast018, YinDY19}, we aim at capturing aleatoric uncertainty. 
Yet, the uncertainty estimates have to generalize to real scenes, which is particularly challenging in the context of self-supervised learning. 
Wannenwetsch~\etal~\cite{ProbFlow} introduced ProbFlow, a probabilistic approach applicable to energy-based optical flow algorithms~\cite{Barron94performanceof, RevaudWHS15, Sun2014}. 
Gast~\etal~\cite{Gast018} proposed probabilistic output layers that require only minimal changes to existing networks. 
Yin~\etal~\cite{YinDY19} introduced HD$^3$F, a method which estimates uncertainty locally, at multiple spatial scales, and further aggregates the results. 
Whereas these approaches are carefully designed for optical flow data and restricted to small displacements, we consider the more general setting of estimating reliable confidence values for dense geometric matching, applicable to \eg pose estimation and 3D reconstruction. This brings additional challenges, including coping with significant appearance changes and large geometric transformations.

\subsection{Differences from the preliminary version \cite{pdcnet}}

This paper extends our work PDC-Net~\cite{pdcnet}, which was published at CVPR 2021.
Our extended paper, contains several new additions compared to its preliminary version.
\newcommand{\bp}[1]{\textbf{#1}}
(i) We introduce an injective criterion for masking out occluded regions that violate a one-to-one ground-truth flow. It allows the network to better learn matching of independently moving objects, while still learning vital interpolation and extrapolation capabilities in occluded regions. 
(ii) We propose an enhanced self-supervised data generation pipeline by introducing multiple independently moving objects to better model challenges encountered in real scenes.
(iii) We present additional ablation studies, in particular analyzing the effectiveness of our injective mask and self-supervised training strategy. 
(iv) We demonstrate the superiority of our confidence estimation over additional baselines, such as variance and forward-backward consistency error~\cite{Meister2017}. 
(v) We present experiments on the indoor pose estimation dataset ScanNet~\cite{scannet}, which demonstrates the generalization properties of our approach, solely trained on outdoor data. 
(vi) We evaluate our dense approach on the Homography dataset HPatches~\cite{Lenc}, in both the dense and sparse settings. 
(vii) We further validate our dense flow and confidence estimation for image-based localization on the Aachen dataset~\cite{SattlerMTTHSSOP18, SattlerWLK12}. 
(viii) We propose an approach for image retrieval that is fully based on the confidence estimates provided by PDC-Net+, and evaluate its performance against state-of-the-art global descriptor retrieval methods on the Aachen dataset~\cite{SattlerMTTHSSOP18, SattlerWLK12}. 
(ix) We introduce a strategy for employing PDC-Net+ to establish matches given sets of sparse keypoints. Its effectiveness is directly validated on HPatches~\cite{Lenc} and for image-based localization on the Aachen dataset~\cite{SattlerMTTHSSOP18, SattlerWLK12}.

\section{Our Approach}
\label{sec:method} 

We introduce PDC-Net+, a method for estimating the dense flow field relating two images, coupled with a robust pixel-wise confidence map. The later indicates the reliability and accuracy of the flow prediction, which is indispensable for applications such as pose estimation, image manipulation, and 3D reconstruction.

\subsection{Probabilistic Flow Regression}
\label{subsec:proba-model}

We formulate dense correspondence estimation with a probabilistic model, which provides a unified framework to learn both the flow and its confidence.
For a given image pair $X = \left(I^q, I^r \right)$ of spatial size $H \times W$, the aim of dense matching is to estimate a flow field $Y \in \mathbb{R}^{H \times W \times 2}$ relating the reference $I^r$ to the query $I^q$. Most learning-based methods address this problem by training a network $F$ with parameters $\theta$ that directly predicts the flow as $Y = F(X; \theta)$. However, this does not provide any information about the confidence of the prediction. 

Instead of generating a single flow prediction $Y$, our goal is to learn the conditional probability density $p(Y| X; \theta)$ of a flow $Y$ given the input image pair $X = \left(I^q, I^r \right)$. This is generally achieved by letting a network predict the parameters $\Phi(X; \theta)$ of a family of distributions $p(Y | X; \theta) = p(Y|\Phi(X; \theta)) = \prod_{ij} p(y_{ij} | \varphi_{ij}(X; \theta))$. To ensure a tractable estimation of the dense flow, conditional independence of the predictions at different spatial locations $(i,j)$ is generally assumed. We use $y_{ij} \in \reals^2$ and $\varphi_{ij} \in \reals^n$ to denote the flow $Y$ and predicted parameters $\Phi$ respectively, at the spatial location $(i,j)$. In the following, we generally drop the sub-script $ij$ to avoid clutter. 

Compared to the direct approach $Y = F(X; \theta)$, the generated parameters $\Phi(X; \theta)$ of the predictive distribution can encode more information about the flow prediction, including its uncertainty. In probabilistic regression techniques for optical flow~\cite{Gast018, IlgCGKMHB18} and a variety of other tasks~\cite{KendallG17, Shen2020, Walz2020}, this is most commonly performed by predicting the \emph{variance} of the estimate $y$.
In these cases, the predictive density $p(y|\varphi)$ is modeled using Gaussian or Laplace distributions. In the latter case, the density is given by,
\begin{equation}
\label{eq:laplace}
\mathcal{L}(y| \mu, \sigma^2) =\frac{1}{\sqrt{2 \sigma_u^2}} e^{-\sqrt{\frac{2}{\sigma_u^2}}|u-\mu_u|} . \frac{1}{\sqrt{2 \sigma_v^2}}
e^{-\sqrt{\frac{2}{\sigma_v^2}}|v-\mu_v|}
\end{equation}
where the components $u$ and $v$ of the flow vector $y=(u,v) \in \mathbb{R}^2$ are modelled with two conditionally independent Laplace distributions. The mean $\mu=[\mu_u, \mu_v]^T \in \mathbb{R}^2$ and variance $\sigma^2=[\sigma^2_u, \sigma^2_v]^T \in \mathbb{R}^2_+$ of the distribution $p(y|\varphi) = \mathcal{L}(y | \mu, \sigma^2)$ are predicted by the network as $(\mu, \sigma^2) = \varphi(X; \theta)$ at every spatial location.

\subsection{Constrained Mixture Model Prediction}
\label{sec:constained-mixture}

\begin{figure}[t]
\centering%
\includegraphics[width=0.40\textwidth]{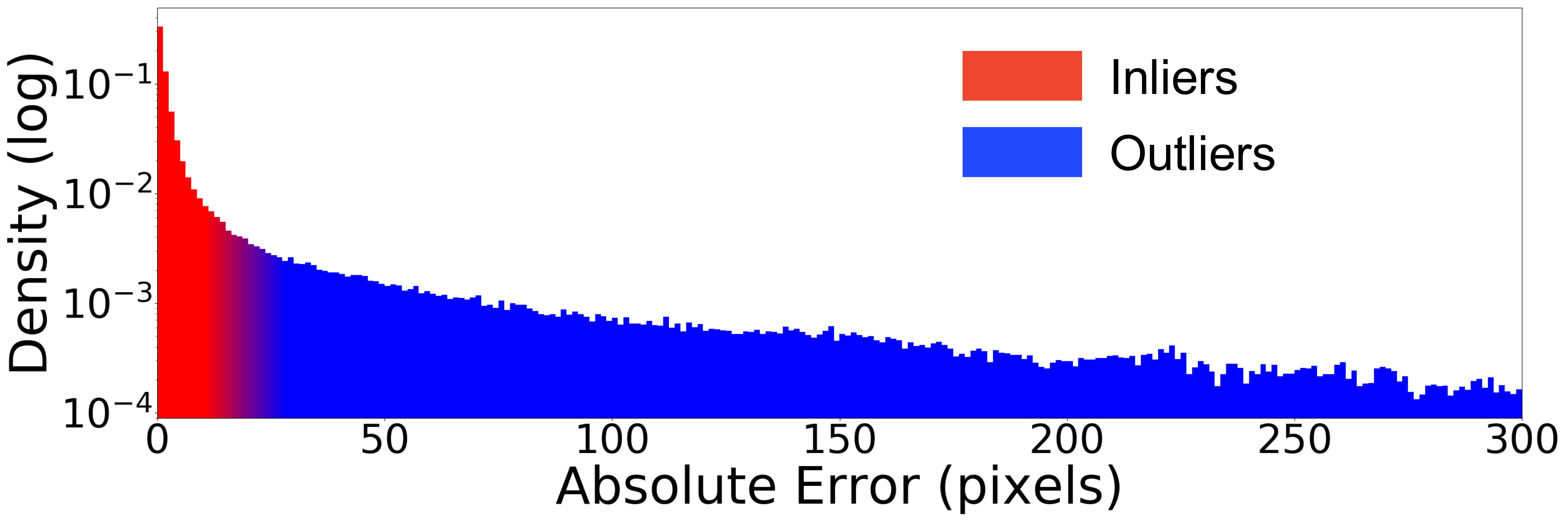}
\vspace{-2mm}
\caption{Distribution of errors 
$|\widehat{y}-y|$ on MegaDepth~\cite{megadepth} between the flow $\widehat{y}$ estimated by GLU-Net~\cite{GLUNet} and the ground-truth $y$. 
}
\vspace{-4mm}
\label{fig:distribution}
\end{figure}

Fundamentally, the goal of probabilistic deep learning is to achieve a predictive model $p(y|X;\theta)$ that coincides with empirical probabilities as well as possible. We can get important insights into this problem by studying the empirical error distribution of a state-of-the-art matching model, in this case GLU-Net~\cite{GLUNet}.  As visualized in Fig.~\ref{fig:distribution},
Errors can be categorized into two populations: inliers (in red) and outliers (in blue). Current probabilistic methods~\cite{Gast018, IlgCGKMHB18,abs-2010-04367} mostly rely on a Laplacian model \eqref{eq:laplace} of $p(y|X;\theta)$. Such a model is effective for correspondences that are easily estimated to be either inliers or outliers with \emph{high certainty}, since their distributions are captured by predicting a low or high variance respectively. 
However, often the network is not certain whether a match is an inlier or an outlier. A single Laplace can only predict an intermediate variance, which does not faithfully represent the more complicated uncertainty pattern in this case.

\begin{figure}[b]
\vspace{-3mm}
\centering%
{\includegraphics[width=0.63\columnwidth]{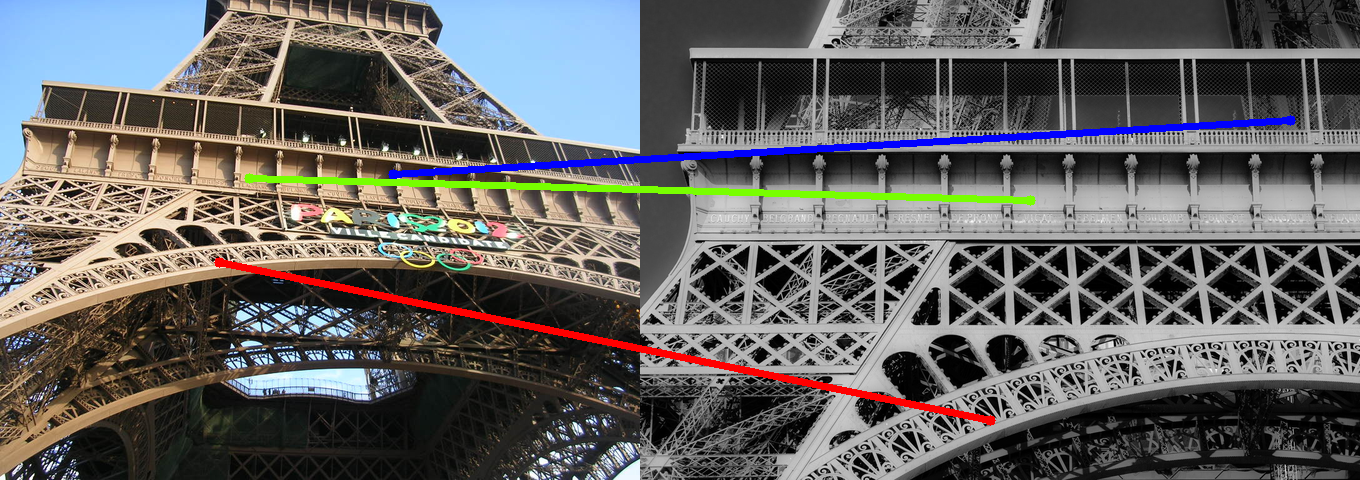}}~%
{\includegraphics[width=0.35\columnwidth]{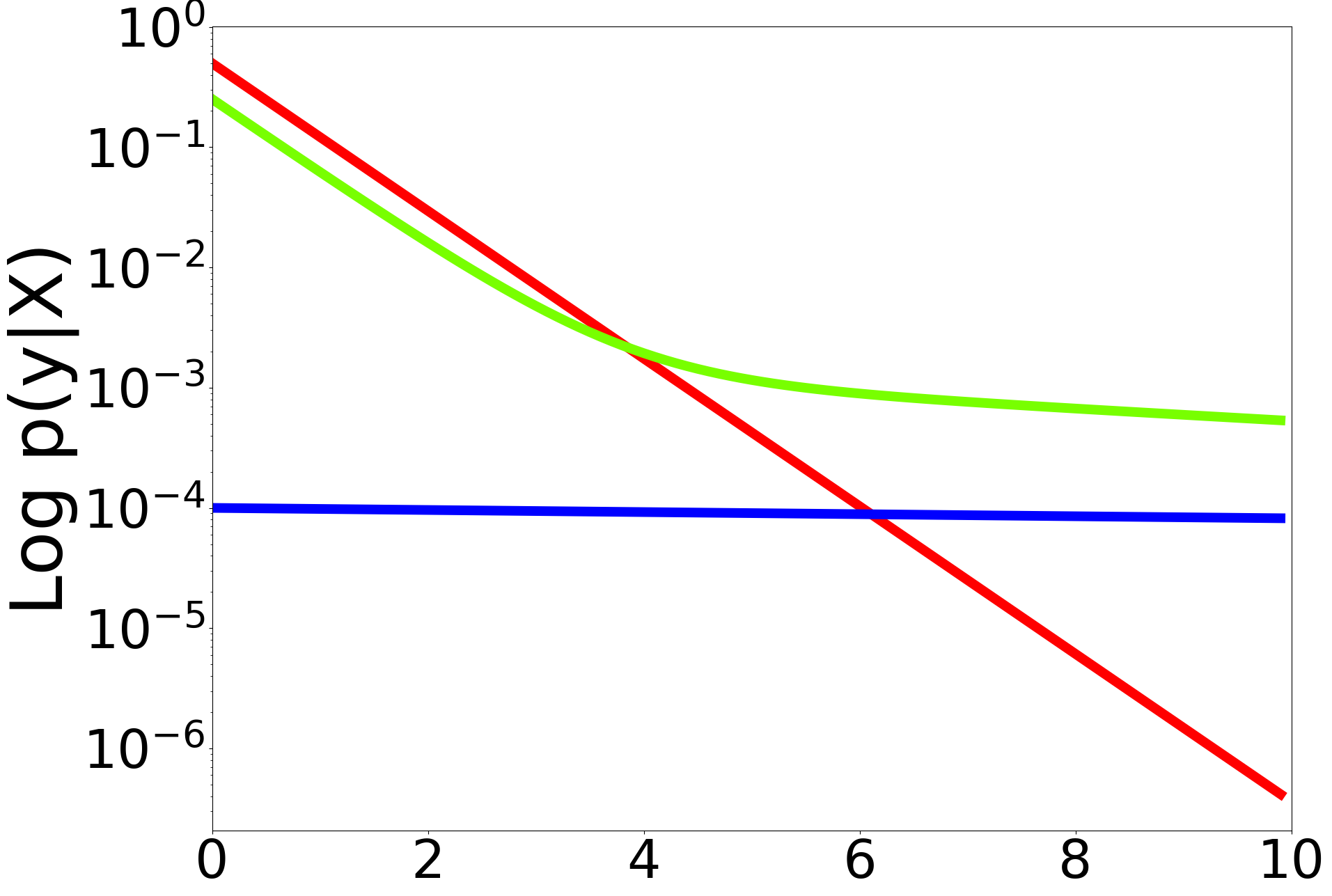}}
\vspace{-2mm}
\caption{Predictive log-density $\log p(y|X)$ \eqref{eq:mixture}-\eqref{eq:constraint} for an inlier (red), outlier (blue), and ambiguous (green) match. Our mixture model faithfully represents the uncertainty also in the latter case.
}
\label{fig:predictive}
\end{figure}

\begin{figure*}[t]
\centering%
\newcommand{\wid}{\textwidth}
\includegraphics[width=\textwidth]{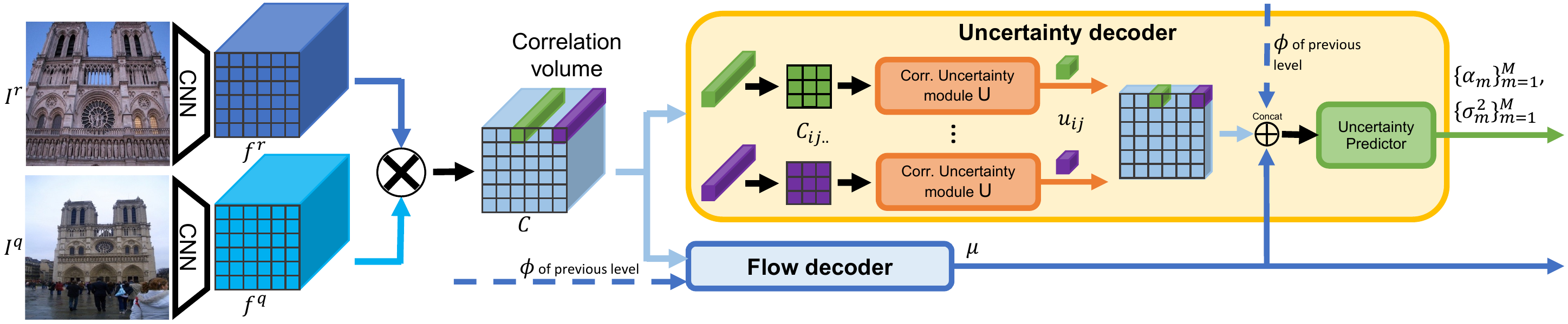}
\vspace{-5mm}\caption{The proposed architecture for flow and uncertainty estimation. The correlation uncertainty module $U_\theta$ independently processes each 2D-slice $C_{ij\cdot\cdot}$ of the correlation volume. Its output is combined with the estimated mean flow $\mu$ and the mixture model parameters $\phi$ from the previous scale level. These are then given to the uncertainty predictor, which finally estimates the weight $\{\alpha_m\}_1^M$ and variance $\{\sigma^2_m\}_1^M$ parameters of our constrained mixture model \eqref{eq:mixture}-\eqref{eq:scalepred}.}\vspace{-4mm} 
\label{fig:arch}
\end{figure*}

\parsection{Mixture model}
To achieve a flexible model capable of fitting more complex distributions, we parametrize $p(y | X; \theta)$ with a mixture model. In general, we consider a distribution consisting of $M$ components, 
\begin{equation}
\label{eq:mixture}
p\left(y | \varphi \right)=\sum_{m=1}^{M} \alpha_{m}  \mathcal{L}\left(y |\mu, \sigma^2_m\right) \,.
\end{equation}
While we have here chosen Laplacian components \eqref{eq:laplace}, any simple density function can be used. The scalars $\alpha_{m} \geq 0$ control the weight of each component, satisfying $\sum_{m=1}^{M} \alpha_{m} = 1$. Note that all components share the same mean $\mu$, which can thus be interpreted as the estimated flow vector. However, each component has a different variance $\sigma^2_m$. 
The distribution \eqref{eq:mixture} is therefore unimodal, but can capture more complex uncertainty patterns. In particular, it allows to model the inlier (red) and outlier (blue) populations in Fig.~\ref{fig:distribution} using separate Laplace components. The network can then predict the probability of a match being an inlier or outlier through the corresponding mixture weights $\alpha_m$. This is visualized in Fig.~\ref{fig:predictive} for a mixture with $M=2$ components. The red and blue matches are with certainty predicted as inlier and outlier respectively, thus requiring only a single active component. In ambiguous cases (green), our mixture model \eqref{eq:mixture} predicts the probability of inlier vs.\ outlier, each modeled with a separate component, giving a better fit compared to the single-component alternative.

\parsection{Mixture constraints}
To employ the mixture model \eqref{eq:mixture}, the network $\Phi$ needs, for each pixel location, to predict the mean flow $\mu$ along with the variance $\sigma^2_m$ and weight $\alpha_m$ of each component, as $\big( \mu, (\alpha_m )_{m=1}^M, ( \sigma^2_m  )_{m=1}^M  \big) = \varphi(X; \theta)$. 
However, an issue when predicting the parameters of a mixture model is its permutation invariance. That is, the predicted distribution \eqref{eq:mixture} is unchanged even if we change the order of the individual components. This can cause confusion in the learning, since the network first needs to \emph{decide} what each component should model before estimating the individual weights $\alpha_m$ and variances $\sigma^2_m$. As shown by our experiments in Sec.~\ref{subsec:ablation-study}, this problem severely degrades the quality of the estimated uncertainties.

We therefore propose a model that breaks the permutation invariance of the mixture \eqref{eq:mixture}. It simplifies the learning and greatly improves the robustness of the estimated uncertainties. In essence, each component $m$ is tasked with modeling a specified range of variances $\sigma^2_m$. We achieve this by constraining the mixture \eqref{eq:mixture} as,
\begin{equation}
\label{eq:constraint}
0 < \beta_1^- \leq \sigma^2_1 \leq \beta_1^+ \leq \beta_2^- \leq \sigma^2_2 \leq \ldots \leq \sigma^2_M \leq \beta_M^+ 
\end{equation}
For simplicity, we here assume a single variance parameter $\sigma^2_m$ for both the $u$ and $v$ directions in \eqref{eq:laplace}.
The constants $\beta_m^-, \beta_m^+$ specify the range of variances $\sigma^2_m$. Intuitively, each component is thus responsible for a different range of uncertainties, roughly corresponding to different regions in the error distribution in Fig.~\ref{fig:distribution}. In particular, component $m=1$ accounts for the most accurate predictions, while component $m=M$ models the largest errors and outliers. 

To enforce the constraint \eqref{eq:constraint}, we first predict an unconstrained value $h_m \in \reals$, which is then mapped to the given range as,
\begin{equation}
\label{eq:scalepred}
\sigma^2_m = \beta_{m}^- + (\beta_m^+ - \beta_{m}^-)\, \text{Sigmoid}(h_m) .
\end{equation}
The constraint values $\beta_m^+, \beta_m^-$ can either be treated as hyper-parameters or learned end-to-end alongside $\theta$. 

Lastly, we emphasize an interesting interpretation of our constrained mixture formulation \eqref{eq:mixture}-\eqref{eq:constraint}. Note that the predicted weights $\alpha_m$, in practice obtained through a final SoftMax layer, represent the probabilities of each component $m$. The network therefore effectively \emph{classifies} the flow prediction at each pixel into the separate uncertainty intervals \eqref{eq:constraint}. 
In fact, our network learns this ability without any extra supervision as detailed next.

\begin{figure*}[t]
\centering%
\newcommand{\wid}{0.16\textwidth}
\vspace{-3mm}
\subfloat[Query image \label{fig:arch-visual-query}]{\includegraphics[width=\wid]{  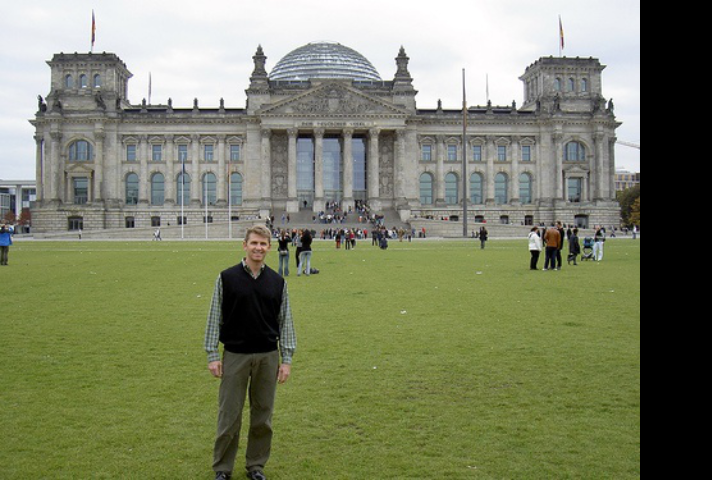}}~%
\subfloat[Reference image \label{fig:arch-visual-ref}]{\includegraphics[width=\wid]{  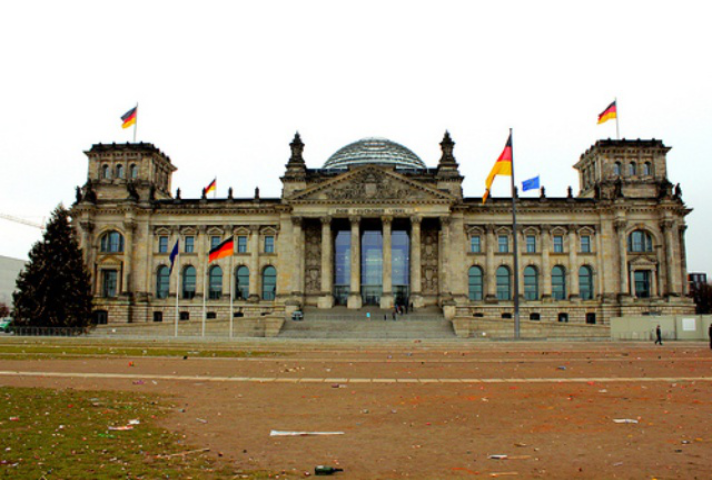}}~%
\subfloat[\centering Common decoder\label{fig:arch-visual-common}]{\includegraphics[width=\wid]{  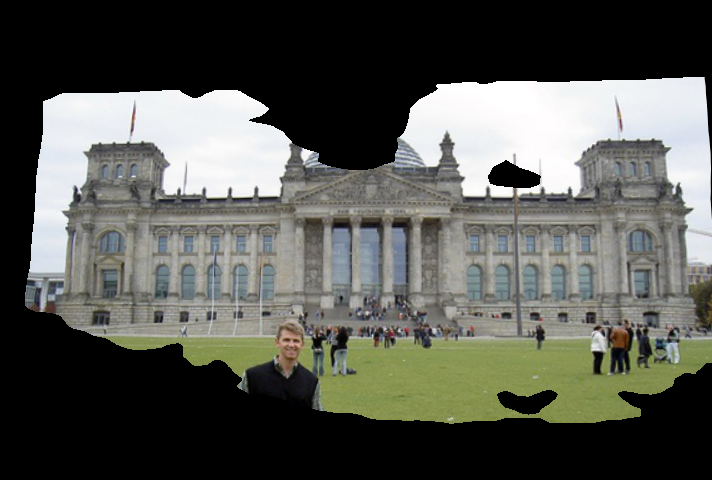}}~%
\subfloat[\centering Our decoder \label{fig:arch-visual-sep}]{\includegraphics[width=\wid]{  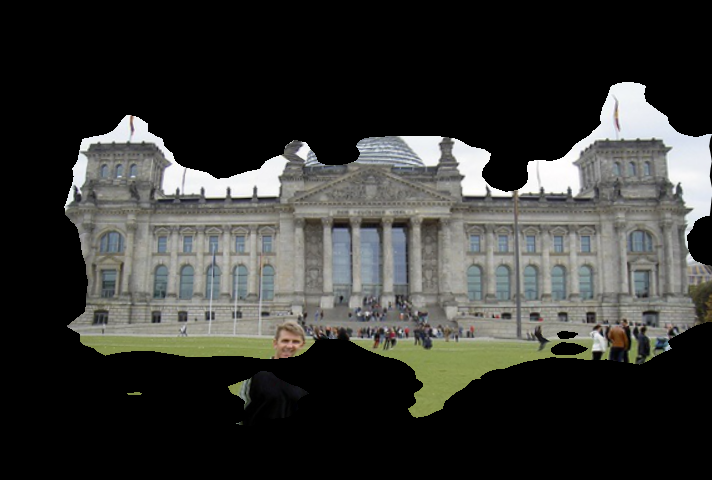}}~%
\subfloat[\centering Our decoder and data \label{fig:arch-visual-sep-pertur}]{\includegraphics[width=\wid]{  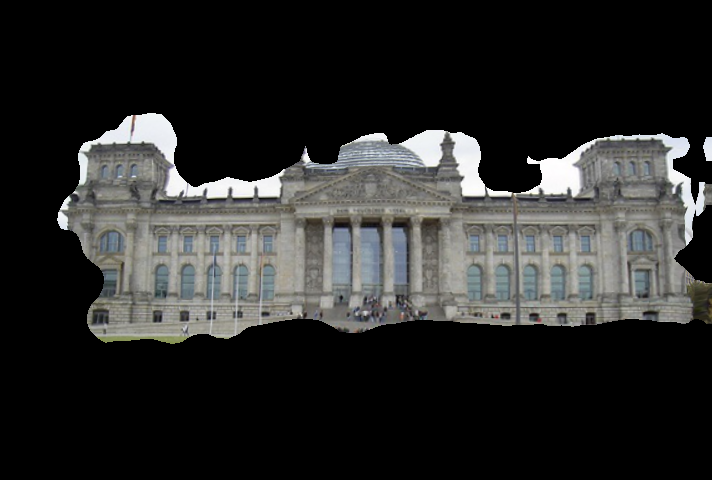}}~%
\subfloat[\centering RANSAC-Flow \label{fig:arch-visual-ransac}]{\includegraphics[width=\wid]{  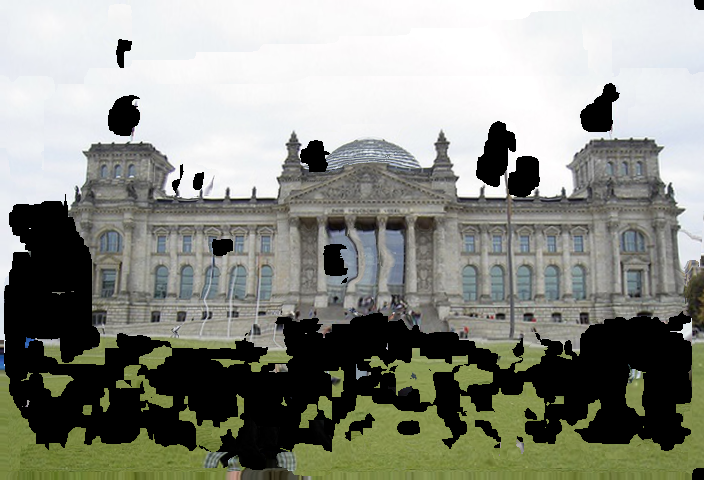}} 
\vspace{-1mm}\caption{Visualization of the estimated uncertainties by masking the warped query image to only show the confident flow predictions. The standard approach \protect\subref{fig:arch-visual-common} uses a common decoder for both flow and uncertainty estimation. It generates overly confident predictions in the sky and grass. The uncertainty estimates are substantially improved in \protect\subref{fig:arch-visual-sep}, when using the proposed architecture described in Sec.~\ref{sec:uncertainty-arch}. 
Adding the flow perturbations for self-supervised training (Sec.~\ref{subsec:perturbed-data}) further improves the robustness and generalization of the uncertainties \protect\subref{fig:arch-visual-sep-pertur}. For reference, we also visualize the flow and confidence mask \protect\subref{fig:arch-visual-ransac} predicted by the recent state-of-the-art approach RANSAC-Flow~\cite{RANSAC-flow}.
}\vspace{-4mm}
\label{fig:arch-visual}
\end{figure*}

\parsection{Training objective}  As customary in probabilistic regression~\cite{prdimp, Gast018, ebmregECCV2020, IlgCGKMHB18, KendallG17, Shen2020, VarameshT20, Walz2020}, we train our method using the negative log-likelihood as the only objective. For one input image pair $X = \left(I^q, I^r \right)$ and corresponding ground-truth flow $Y$, the objective is given by
\begin{equation}
\label{eq:nll}
    - \log p\big(Y|\Phi(X;\theta)\big) = - \sum_{ij} \log p\big(y_{ij} | \varphi_{ij}(X;\theta)\big)  
\end{equation}
In Appendix~B.1, we provide efficient analytic expressions of the loss \eqref{eq:nll} for our constrained mixture \eqref{eq:mixture}-\eqref{eq:constraint}, that also ensure numerical stability.

\parsection{Bounding the variance}
Our constrained formulation \eqref{eq:constraint} allows us to circumvent another issue with the direct application of the mixture model \eqref{eq:mixture}. When trained with the standard negative log-likelihood objective \eqref{eq:nll}, an unconstrained model encourages the network to primarily focus on easy correspondences. By accurately predicting such matches with high confidence (low variance), the network can arbitrarily reduce the loss during training. Generating fewer, but highly accurate predictions then dominates during training at the expense of more challenging regions. Fundamentally, this problem appears since the negative log-likelihood loss is theoretically unbounded in this case. We solve this issue by simply setting a non-zero lower bound $0 < \beta_1^-$ for the predicted variances in our formulation \eqref{eq:constraint}. This effectively provides a lower bound on the negative-log likelihood loss itself, leading to a well-behaved objective. Such a constraint can be seen as adding a prior term in order to regularize the likelihood. In our experiments, we simply set the lower bound to be a standard deviation of one pixel, i.e.\ $1 = \beta_1^- \leq \sigma^2_1$.

\subsection{Uncertainty Prediction Architecture}
\label{sec:uncertainty-arch}

Our aim is to predict an uncertainty value that quantifies the \emph{reliability} of a proposed correspondence or flow vector. Crucially, the uncertainty prediction needs to \emph{generalize} well to real scenarios, not seen during training. 
However, this is particularly challenging in the context of self-supervised training, which relies on synthetically warped images or animated data. Specifically, when trained on simple synthetic motion patterns, such as homography transformations, the network learns to heavily rely on global smoothness assumptions, which do not generalize well to more complex settings. As a result, the network learns to \emph{confidently} interpolate and extrapolate the flow field to regions where no robust match can be found. 
Due to the significant distribution shift between training and test data, the network thus also infers confident, yet highly erroneous predictions in homogeneous regions on real data. 
In this section, we address this problem by carefully designing an architecture that greatly limits the risk of the aforementioned issues. Our architecture is visualized in Figure~\ref{fig:arch}.

Current state-of-the-art dense matching architectures rely on feature correlation layers. Features $f$ are extracted at resolution $h \times w$ from a pair of input images, and densely correlated either globally or within a local neighborhood of size $d$. In the latter case, the output correlation volume is best thought of as a 4D tensor $C \in \reals^{h \times w \times d \times d}$. Computed as dense scalar products $C_{ijkl} = (f_{ij}^r)\tp f_{i+k,j+l}^q$, it encodes the deep feature similarity between a location $(i,j)$ in the reference frame $I^r$ and a displaced location $(i+k,j+l)$ in the query $I^q$. Standard flow architectures~\cite{GLUNet, Melekhov2019, Hui2018, Sun2018} process the correlation volume with a flow decoder, by first vectorizing the last two dimensions, before applying a sequence of convolutional layers over the \emph{reference coordinates $(i,j)$}  to predict the final flow.  

\parsection{Correlation uncertainty module}
The straightforward strategy for predicting the distribution parameters $\Phi(X;\theta)$ is to simply increase the number of output channels of the flow decoder to include all parameters of the predictive distribution. 
However, this allows the network to rely primarily on the local neighborhood when estimating the flow and confidence at location $(i,j)$. It thus ignores the actual reliability of the match and appearance information at the specific location. It results in over-smoothed and overly confident predictions, unable to identify ambiguous and unreliable matching regions, such as the sky. This is visualized in Fig.~\ref{fig:arch-visual-common}.

We instead design an architecture that assesses the uncertainty at a specific location $(i,j)$, without relying on neighborhood information. We note that the 2D slice $C_{ij\cdot\cdot} \in \reals^{d\times d}$ of the correlation volume encapsulates rich information about the matching ability of location $(i,j)$, in the form of a confidence map. 
In particular, it encodes the distinctness, uniqueness, and existence of the correspondence.
We therefore create a \emph{correlation uncertainty module} $U_{\theta}$ that independently reasons about each correlation slice as $U_\theta(C_{ij\cdot\cdot})$. In contrast to standard decoders, the convolutions are therefore applied over \emph{the displacement dimensions $(k,l)$}. Efficient parallel implementation is ensured by moving the first two dimensions of $C$ to the batch dimension using a simple tensor reshape. Our strided convolutional layers then gradually decrease the size $d \times d$ of the displacement dimensions $(k,l)$ until a single vector $u_{ij} = U_\theta(C_{ij\cdot\cdot}) \in \reals^n$ is achieved for each spatial coordinate $(i,j)$ (see Fig.~\ref{fig:arch}). 

\parsection{Uncertainty predictor} 
The cost volume does not capture uncertainty arising at motion boundaries, crucial for real data with independently moving objects.
We thus additionally integrate predicted flow information in the estimation of its uncertainty. In practice, we concatenate the estimated mean flow $\mu$ with the output of the correlation uncertainty module $U_{\theta}$, and process it with multiple convolution layers. The uncertainty predictor outputs all parameters of the mixture \eqref{eq:mixture}, except for the mean flow $\mu$ (see Fig.~\ref{fig:arch}). 
As shown in Fig.~\ref{fig:arch-visual-sep}, our uncertainty decoder, comprised of the correlation uncertainty module and the uncertainty predictor, successfully masks out most of the inaccurate and unreliable matching regions. 

\parsection{Uncertainty propagation across scales} 
In a multi-scale architecture, the uncertainty prediction is further propagated from one level to the next. Specifically, the flow decoder and the uncertainty predictor both take as input the parameters $\Phi$ of the distribution predicted at the previous level, in addition to their original inputs.

\begin{figure*}[t]
\centering%
\includegraphics[width=0.99\textwidth]{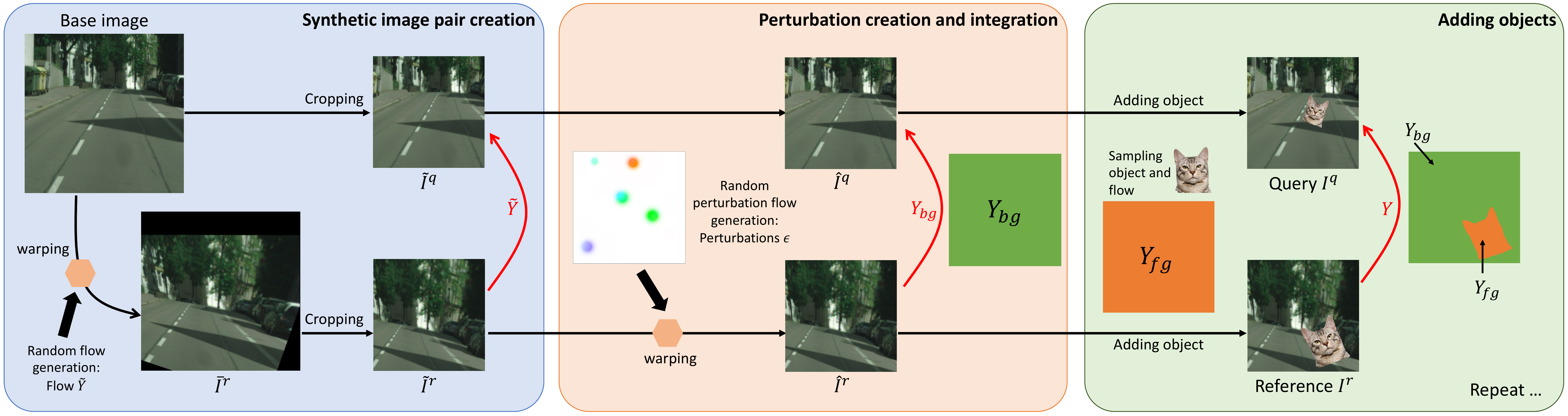}
\vspace{-2mm}
\caption{The synthetic image pair generation pipeline for our self-supervised training. Our pipeline is divided in three parts. First, we randomly generate a synthetic flow field $\tilde{Y}$, used to create an image pair from a base image. Next, perturbations are added to the synthetic flow field, leading to a new pair of query and reference images related by the background flow field $Y_{bg}$. Finally, each object is iteratively added to the image pair, using a randomly sampled object flow $Y_{fg}$. The final ground-truth flow field $Y$ relating the image pair is also updated accordingly.}
\vspace{-4mm}
\label{fig:dataset}
\end{figure*}

\subsection{Data for Self-supervised Uncertainty}
\label{subsec:perturbed-data}

While designing a suitable architecture greatly alleviates the uncertainty generalization issue, the network still tends to rely on global smoothness assumptions and interpolation, especially around object boundaries (see Fig.~\ref{fig:arch-visual-sep}). While this learned strategy indeed minimizes the Negative Log Likelihood loss \eqref{eq:nll} on self-supervised training samples, it does not generalize to real image pairs. 
In this section, we further tackle this problem from the data perspective in the context of self-supervised learning. 

We aim at generating less predictable synthetic motion patterns than simple homography transformations, to prevent the network from primarily relying on interpolation. This forces the network to focus on the appearance of the image region in order to predict its motion and uncertainty.
Given a base flow $\tilde{Y} \in \mathbb{R}^{H \times W}$ relating $\tilde{I}^r \in \mathbb{R}^{H \times W}$ to $\tilde{I}^q \in \mathbb{R}^{H \times W}$ by a simple transformation, such as a homography~\cite{Melekhov2019, Rocco2017a, GOCor, GLUNet}, we create a residual flow $\epsilon = \sum_i \varepsilon_i \in \mathbb{R}^{H \times W}$ by adding small local perturbations $\varepsilon_i$. 
The query image $I^q = \tilde{I}^q$ is left unchanged  while the reference $I^r$ is generated by warping $\tilde{I}^r$ according to the residual flow $\epsilon$. 
The final perturbed flow map $Y$ between $I^r$ and $I^q$ is achieved by composing the base flow $\tilde{Y}$ with the residual flow $\epsilon$. 

The main purpose our perturbations $\varepsilon$ is to teach the network to be uncertain in regions where they cannot easily be identified. Specifically, in homogeneous regions such as the sky, the perturbations do not change the appearance of the reference image ($I^r \approx \tilde{I}^r$) and are therefore unnoticed by the network. However, since the perturbations break the global smoothness of the synthetic flow, the flow errors of those pixels are higher. In order to decrease the loss \eqref{eq:nll}, the network thus needs to estimate a larger uncertainty for the perturbed regions. We show the impact of introducing the flow perturbations in Fig.~\ref{fig:arch-visual-sep-pertur}.

\subsection{Self-supervised dataset creation pipeline}
\label{sec:training-strategy}

We train our final model using a combination of self-supervision, where the image pairs and corresponding ground-truth flows are generated by artificial warping, and real sparse ground-truth. In this section, we focus on our self-supervised data generation pipeline. It is illustrated in Fig~\ref{fig:dataset} and detailed below. 

\parsection{Synthetic image pair creation} First, an image pair $(\tilde{I}^r, \tilde{I}^q)$ of dimension $H \times W$ is generated by warping a base image B with a randomly generated geometric transformation, such as a homography or Thin-plate Spline (TPS) transformations. 
The base image $B$ is first resized to a fixed size $\overline{H} \times \overline{W}$, larger than the desired training image size $H \times W$. We then sample a random dense flow field $\tilde{Y}$ of the same dimension $\overline{H} \times \overline{W}$, and generate $\overline{I}^r$ by warping image $B$ with $\tilde{Y}$. 
The query image $\tilde{I}^q$ is created by centrally cropping the base image $B$ to the fixed training image size $H \times W$, while the reference $\tilde{I}^r$ results from the cropping of $\overline{I}^r$. 
The new image pair $(\tilde{I}^r, \tilde{I}^q)$ is related by the ground-truth flow $\tilde{Y}$, which is also cropped and adjusted accordingly. 
The overall procedure is similar to~\cite{Rocco2017a, GLUNet, GOCor}.

\parsection{Perturbation creation and integration} Next, perturbations are added to the reference, as detailed in Sec.~\ref{subsec:perturbed-data}, resulting in a new pair ($\hat{I}^q, \hat{I}^r$) that is related by background flow field $Y_{bg}$. 

\parsection{Adding objects} To better simulate real scenarios, the synthetic image pair is augmented with random independently moving objects. This is performed by sampling objects from the COCO dataset~\cite{coco}, using their provided ground-truth segmentation masks. To generate motion, we randomly sample an affine flow field $Y_{fg}$ for each object, referred to as the foreground flow. 
The objects are inserted into the images ($\hat{I}^q, \hat{I}^r$) using their corresponding segmentation masks, giving the final training image pair $(I^r, I^q)$.
The final synthetic flow $Y$ relating the reference to the query is composed of the object motion flow field $Y_{fg}$ at the location of the moving object in the reference image, and the background flow field $Y_{bg}$ otherwise. The same procedure is repeated iteratively, by considering previously added objects as part of the background.

\subsection{Injective mask} 
\label{sec:injective-mask}

Occlusions are pervasive in real scenes. They are generally caused by moving objects or 3D motion of the underlying scene. When a reference frame pixel is occluded or outside the view of the query image, its estimated flow needs to be interpolated or extrapolated. Occluded regions are therefore one of the most important sources of uncertainty in practical applications. Moreover, matching occluded and other non-visible pixels is often virtually impossible or even undefined. 

In our self-supervised pipeline, occlusions are simulated by inserting independently moving objects. Since the background image is also transformed by a random warp, the flow is known even in occluded areas. As we have the ground-truth flow vector for every pixel in the reference image, the most straightforward alternative is to train the network by applying the Negative Log Likelihood loss \eqref{eq:nll} over \emph{all} pixels of the reference image. 
However, this choice makes it difficult for the network to generalize to real scenes. 
The background flow is easier to learn compared to the object flows, since the regular motion of the former largely dominates within the image. 
As a result, the network tends to ignore the independently moving objects, while predicting background flow for all pixels in the image.

To alleviate this problem, another alternative is to mask out all occluded regions from the loss computation~\cite{Melekhov2019}. However, by removing all supervisory signals in occluded regions, the network is unable to learn the interpolation and extrapolation of predictions, which is important under mild occlusions in real scenes.

We therefore propose to mask out pixels from the objective based on another condition, namely the \emph{injectivity} of the ground-truth correspondence mapping. Specifically, we mask out the minimal region that ensures a one-to-one ground-truth mapping between the frames. If two or more pixels in the reference frame map to the same pixel location in the query, we only preserve the \emph{visible} pixel and mask out the occluded ones. This allows the network to focus on flow estimation of visible moving objects as opposed to occluded background regions. However, occlusions that do not violate the injectivity condition are preserved in the loss. The network thus learns vital interpolation and extrapolation capabilities. Importantly, the injectivity condition implies an unambiguous and well-defined reverse (i.e.\ inverse) flow. By only allowing such one-to-one matches during training, the network can learn to exploit the uniqueness of a correspondence as a powerful cue when assessing its uncertainty and existence. 

\begin{figure}[t]
\centering%
\includegraphics[width=0.40\textwidth]{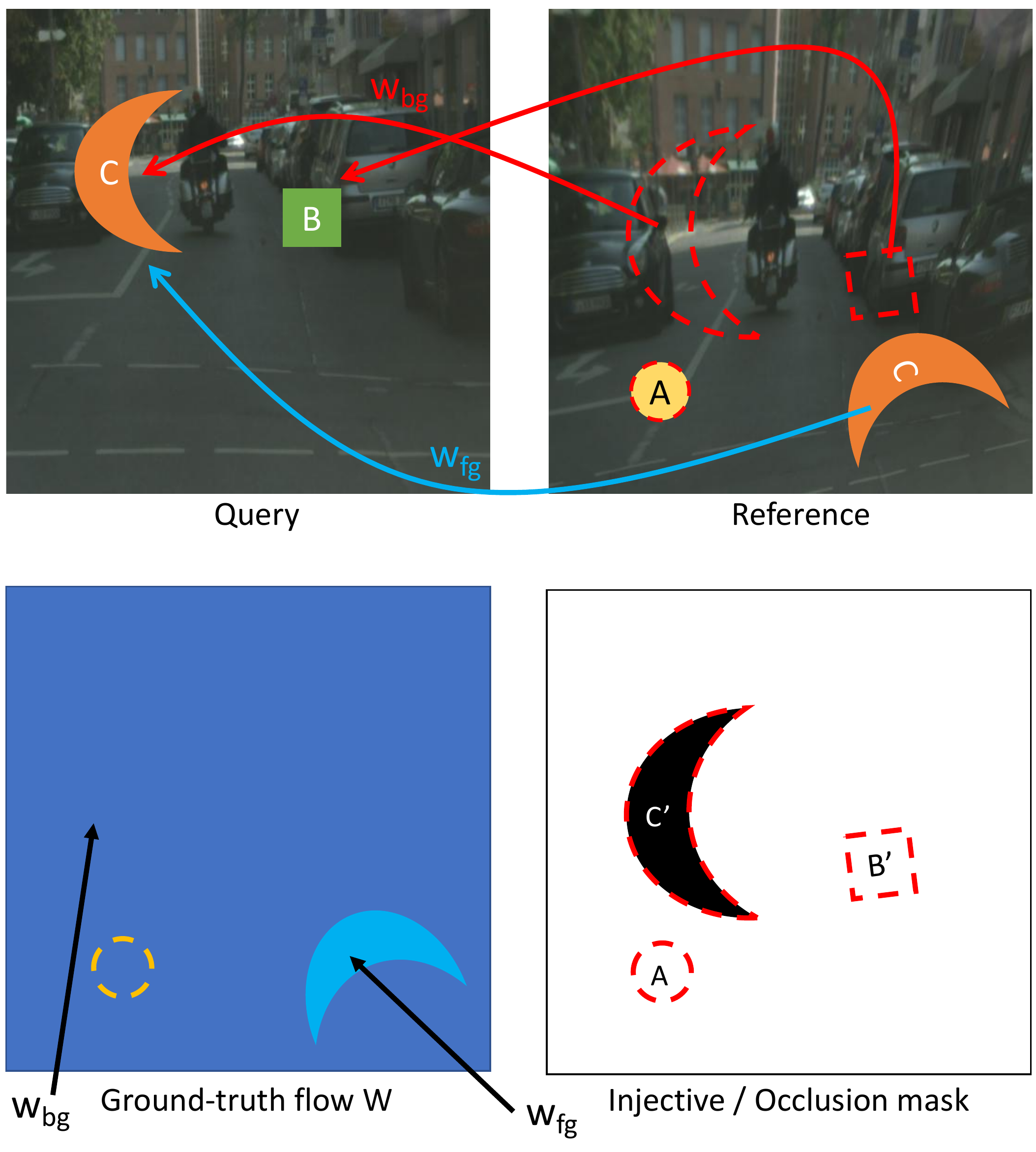}
\vspace{-2mm}
\caption{Pair of query and reference images with independently moving objects $A$, $B$ and $C$. Objects $A$ and $B$ are solely visible in the reference and query images respectively. The regions in the reference image that are occluded by the objects are represented with dashed red contours. They correspond to areas $A$, $B'$ and $C'$ in the occlusion mask. Our injective mask corresponds to $C'$, \ie the black area in the lower right frame. 
}
\vspace{-4mm}
\label{fig:mask}
\end{figure}

We illustrate and further discuss our injective masking approach for self-supervised training using the minimal example visualized in Fig.~\ref{fig:mask}. The query image has two objects, $B$ and $C$. Object $C$ is also visible in the reference image while object B is not present there. The reference image also includes an object $A$, which is solely visible in the reference. Next, we discuss how to handle the cases $A$, $B$, and $C$ when defining the ground-truth flow and our injective mask. Note that the ground-truth flow and our injective mask are both defined in the coordinate system of the reference frame. 

\noindent\textbf{Object $A$} is only visible in the reference frame. The network therefore cannot deduce its flow from the available information. However, the image region covered by object $A$ in the reference has a one-to-one background flow. We therefore adopt this as ground-truth in region $A$. It allows the network to learn to interpolate the flow for pixels not visible in the reference frame (i.e.\ the region behind object $A$).

\noindent\textbf{Object $B$} is solely visible in the query. The background region covered by $B$ in the query corresponds to $B'$ in the reference. Since no other pixels in the reference are mapped to $B$, we can use the background flow as ground-truth in region $B'$ without violating the injectivity condition. 
Through the supervisory signal in $B'$, the network learns to interpolate the flow for pixels not visible in the query frame (i.e.\ the region behind object $B$).

\noindent\textbf{Object $C$} is visible in both frames. The region $C'$ in the reference is occluded by object $C$ in the query frame. 
Both the background flow in region $C'$ and the object flow in region $C$ of the reference map to the same region $C$ in the query frame. Including both regions in the ground-truth leads to a non-injective mapping. We therefore mask out the occluded region $C'$ from the loss, while preserving the visible region $C$. This forces the network to focus on learning the flow for the moving object $C$, which is more challenging. When including both regions in the loss \eqref{eq:nll} instead, the network tends to ignore the object $C$ in favor of the occluded background flow $C'$, by predicting a high uncertainty for the former.

\parsection{Injective mask} 
The final injective mask for the example given in Fig.~\ref{fig:mask} thus corresponds to the region $\Omega_\text{inj}=C'$, visualized in black. In comparison, the full occlusion mask is given by the larger region $\Omega_\text{occ}=C' \cup B' \cup A$, which is outlined by the red dashed region in Fig.~\ref{fig:mask}. Hence, our approach does not mask out occluded regions that preserve the injectivity of the ground-truth flow.

The example described in Fig.~\ref{fig:mask} covers all important cases in the construction of our ground-truth flow and the injective mask. In order to achieve a general approach that is applicable to any number and configuration of objects, we follow an iterative procedure. We first construct the background image pair and its corresponding flow as described in Sec.~\ref{sec:training-strategy}. In each iteration, we then add one new object as discussed in Sec.~\ref{sec:training-strategy}. The added object belongs to one of the cases $A$, $B$, and $C$ above. We update the ground-truth flow and mask as previously described, while considering all previously added objects as part of the background. Note that we preserve the flow for regions in the reference frame that are out-of-view in the query frame, since they comply with the one-to-one property of the ground-truth mapping.

\parsection{Masked objective}
Our final objective is obtained by masking the Negative Log-Likelihood \eqref{eq:nll} with our injective mask.
For the input image pair $X = \left(I^q, I^r \right)$ and ground-truth flow $Y$, we simply sum over all pixels that are not in the masked-out region $\Omega_\text{inj}$,
\begin{equation}
\label{eq:nll-mask}
    L(\theta; X, Y) = - \sum_{(i,j) \notin \Omega_\text{inj}}  \log p\big(y_{ij} | \varphi_{ij}(X;\theta)\big) \,.
\end{equation}
In Appendix~B.2, we present visualization of our injective mask for multiple example training image pairs. 

\subsection{Geometric Matching Inference}
\label{sec:geometric-inference}

Our PDC-Net+ generates a predictive distribution, from which we extract the mean flow field and its corresponding confidence map. This provides substantial versatility, allowing our approach to be deployed in multiple scenarios.
In this section, we present multiple inference strategies, which utilize our confidence estimation to also improve the accuracy of the final flow prediction. 
The simplest application of PDC-Net+ predicts the flow and its confidence in a single forward pass. Our approach further offers the opportunity to perform \emph{multi-stage} and \emph{multi-scale} flow estimation on challenging image pairs, without any additional components, by leveraging our internal confidence estimation.
Finally, our dense flow regression can also be used to find robust matches between two sparse sets of detected keypoints.

\parsection{Confidence value} From the predictive distribution $p(y|\varphi(X;\theta))$, we aim at extracting a single confidence value, encoding the reliability of the corresponding predicted mean flow vector $\mu$. 
Previous probabilistic regression methods mostly rely on the variance as a confidence measure~\cite{Gast018, IlgCGKMHB18,  KendallG17, Walz2020}. However, we observe that the variance can be sensitive to outliers. Instead, we compute the probability $P_R$ of the true flow $y$ being within a radius $R$ of the estimated mean flow $\mu$. It can be computed as,
\begin{equation}
\label{eq:pr}
P_R  = P(\|y - \mu\|_\infty < R) = \sum_{m} \alpha_m \left[1-\exp (-\sqrt{2}\frac{R}{\sigma_m}) \right]^2 
\end{equation}
Compared to the variance, the probability value $P_R$ provides a more interpretable measure of the uncertainty. This confidence measure is then used to identify the accurate from the inaccurate matches by thesholding $P_R$. The accurate matches may then be further utilized in down-stream applications.

\begin{figure}[t]
\centering%
\includegraphics[width=0.49\textwidth]{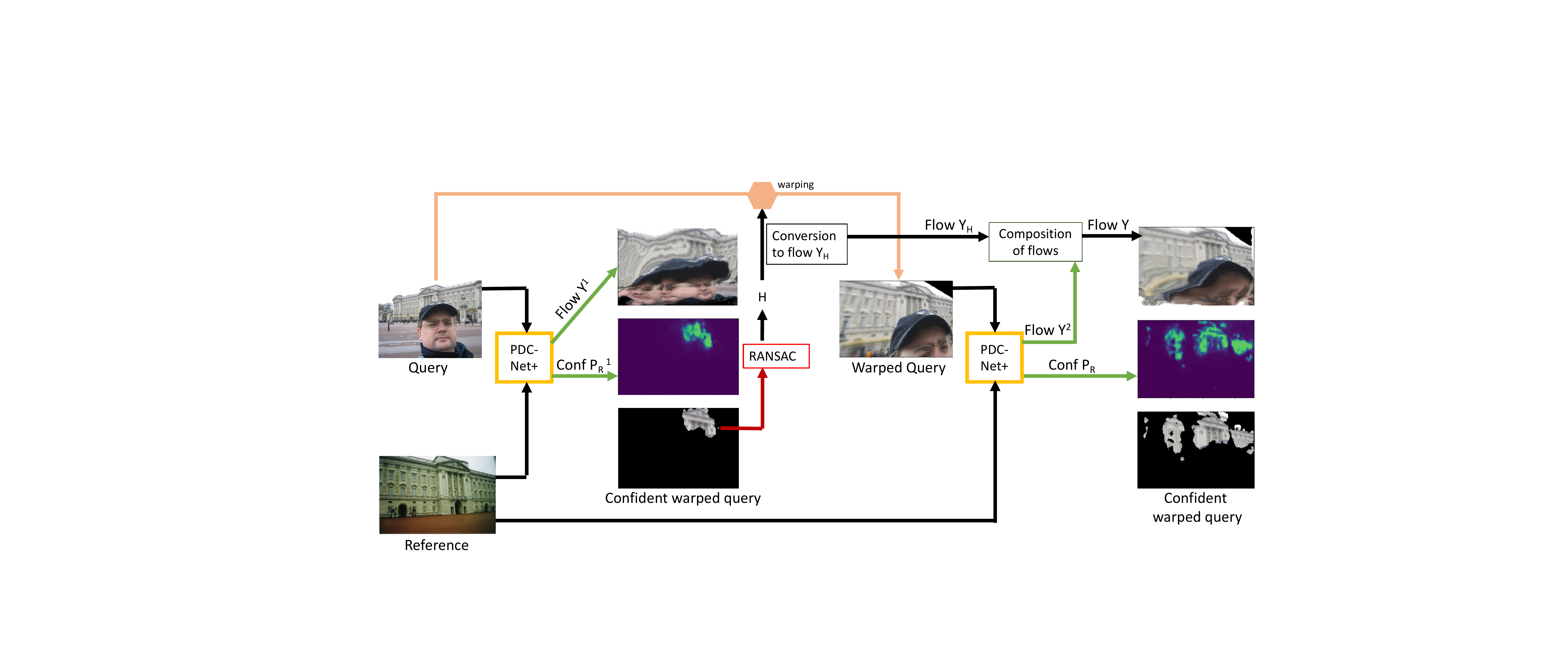}
\caption{Illustration of our multi-stage refinement strategy (H) to predict the flow $Y$ and confidence map $P_R$ relating the reference to the query. 
}
\vspace{-4mm}
\label{fig:multi-stage}
\end{figure}

\parsection{Multi-stage refinement strategy (H)}  
For extreme viewpoint changes with large scale or perspective variations, it is particularly difficult to infer the correct motion field in a single network pass. While this is partially alleviated by multi-scale architectures, it remains a major challenge in geometric matching. 
Our approach allows to split the flow estimation process into two parts, the first estimating a simple transformation, which is then used as initialization to infer the final, more complex transformation. 

One of the major benefits of our confidence estimation is the ability to \emph{identify} a set of accurate matches from the densely estimated flow field. This is performed by thresholding the confidence map $P_R$ as $P_{R=1} > \gamma$. Following the first forward network pass, these accurate correspondences can be used to estimate a coarse transformation relating the image pair, such as a homography transformation. After warping the query according to the homography, a second forward pass can then be applied to the coarsely aligned image pair. The final flow field is constructed as a composition of the fine flow and the homography transform flow. This process is illustrated in Fig.~\ref{fig:multi-stage}. In our experiments Sec.~\ref{sec:exp}, we indicate this version of our approach with (H). 
While previous works also use multi-stage refinement~\cite{Rocco2017a, RANSAC-flow}, our approach is much simpler, applying the \emph{same} network in both stages and benefiting from the internal confidence estimation. 

\parsection{Multi-scale refinement strategy (MS)} For datasets with very extreme geometric transformations, we further suggest a multi-scale strategy that utilizes our confidence estimation. In particular, we extend our two-stage refinement approach  by resizing the reference image to different resolutions. Specifically, following~\cite{RANSAC-flow}, we use seven scales: 0.5, 0.88, 1, 1.33, 1.66 and 2.0. As for the implementation, to avoid obtaining very large input images (for scaling ratio 2.0 for example), we use the following scheme: we resize the reference image for scaling ratios below 1, keeping the aspect ratio fixed and the query image unchanged. For ratios above 1, we instead resize the query image by one divided by the ratio, while keeping the reference image unchanged. This ensures that the resized images are never larger than the original image dimensions. 
The resulting resized image pairs are then passed through the network and we fit a homography for each pair, using our predicted flow and uncertainty map. 
From all image pairs with their corresponding scaling ratios, we then select the homography with the highest percentage of inliers, and scale it to the images original resolutions. The original image pair is then coarsely aligned using this homography. Lastly, we follow the same procedure used in our two-stage refinement strategy to predict the final flow. We refer to this process as Multi-Scale (MS).

\parsection{Using our dense flow for sparse matching} Dense correspondences are useful in many applications but are not mandatory in geometric transformation estimation tasks. In such tasks, a detect-then-describe strategy can also be used. In the first step, locally salient and repeatable points are detected in both images. Secondly, descriptors are extracted at the keypoint locations and matched to establish correspondences across the images. Our dense approach can be used to replace the second step, i.e.\ description and matching. 
Instead of solely relying on descriptor similarities~\cite{SIFT, SURF, Brief, ORB, Belousov2017, superpoint, Dusmanu2019CVPR}, our dense probabilistic correspondence network additionally learns to utilize, e.g., local motion patterns and global consistency across the views.

Specifically, we also employ our predicted dense flow and confidence map to find correspondences given sets of sparse keypoints, detected in a pair of images. 
We denote the set of $N$ and $L$ sparse feature points detected in the query $I^q$ and reference image $I^r$ respectively as $\mathcal{X}^q = \{x^q_i\}_{i=1}^N \subset  \mathbb{R}^{2}$ and $\mathcal{X}^r = \{x^r_i\}_{i=1}^L \subset  \mathbb{R}^{2}$. PDC-Net+ predicts the dense flow field $Y^{r \rightarrow q}$ and confidence map $P_R^{r \rightarrow q}$ relating the reference to the query image. 
We first disregard all reference feature points $x^r_i \in K^r$ for which $P_R^{r \rightarrow q}[x^r_i] < \gamma$, where $\gamma \in [0, 1)$ is a threshold. 
Given a reference feature point $x^r_i \in \mathcal{X}^r$ such that $P_R^{r \rightarrow q}[x^r_i] \geq \gamma$, we compute its predicted matching point in the query $\widehat{x}^q_i = Y^{r \rightarrow q}[x^r_i] + x^r_i$ according to the predicted flow $Y^{r \rightarrow q}$. 
We subsequently find the query point $x^q_j \in \mathcal{X}^r$ closest to $\widehat{x}^q_i$, that is $j = \text{argmin}_k \left\| \widehat{x}^q_i - x^q_k \right\|$. 
Finally, we retain the match between $x^r_i$ and $x^q_j$ if $\left\| \widehat{x}^q_i - x^q_j \right\| < d$, where $d$ is set as a pixel distance threshold.
As a result, we can potentially identify a corresponding feature point $x^q_j \in \mathcal{X}^q$ for each $x^r_i \in \mathcal{X}^r$. The set of such correspondences is denoted is $C_{r \rightarrow q}$.

We can use the same strategy to establish correspondences $C_{q \rightarrow r}$ in the reverse direction from $I^q$ to $I^r$. Given the two sets of correspondences from the two matching directions, we optionally only keep the correspondences for which the cyclic consistency error is below a certain threshold.

\section{Experimental Results}
\label{sec:exp}

We integrate our probabilistic approach into a generic pyramidal correspondence network and perform comprehensive experiments on multiple geometric matching and optical flow datasets. We further evaluate our probabilistic dense correspondence network, PDC-Net+, for several downstream tasks, including pose estimation, image-based localization, and image retrieval. For all tasks and datasets, we employ the \emph{same network}, PDC-Net+, with the \emph{same weights} without any task or dataset-specific fine-tuning.   
Further results, analysis, visualizations and implementation details are provided in the Appendix. 

\subsection{Implementation Details}
\label{subsec:arch}

We train a single network, termed PDC-Net+, and use the same network and weights for all experiments.

\parsection{Architecture} 
We adopt the recent GLU-Net-GOCor~\cite{GOCor, GLUNet} as our base architecture. It consists of two sub-modules operating at two image resolutions, each built as a two-level pyramidal network. The feature extraction backbone consists of a VGG-16 network~\cite{Chatfield14} pre-trained on ImageNet. At each level, we add our uncertainty decoder and propagate the uncertainty prediction to the next level as detailed in Sec.~\ref{sec:uncertainty-arch}. 

We model the probability distribution $p\left(y | \varphi \right)$ with the constrained mixture presented in Sec.~\ref{sec:constained-mixture}, using $M=2$ Laplace components. The first is fixed to $\sigma^2_1 = \beta_1^- = \beta_1^+ = 1$ in order to represent the very accurate predictions. The second component models larger errors and outliers, obtained by setting the constraints as $ 2 = \beta_2^- \leq \sigma^2_2 \leq \beta_2^+ = HW$, where $\beta_2^+$ is set to the size $H \times W$ of the training images. We ablate these design choices in Sec.~\ref{subsec:ablation-study}

For inference, we refer to using a single forward-pass of the network as (D), our multi-stage approach (Sec.~\ref{sec:geometric-inference}) as (H) and our multi-scale approach (Sec.~\ref{sec:geometric-inference}) as (MS). 

\parsection{Training datasets} 
We train our network in two stages.
First, we follow the self-supervised training procedure introduced in Sec.~\ref{sec:training-strategy}. In particular, we augment the data with four random independently moving objects from the COCO~\cite{coco} dataset, with probability 0.8. 
In the second training stage, we extend the self-supervised data with real image pairs with sparse ground-truth correspondences from the MegaDepth dataset~\cite{megadepth}. We additionally fine-tune the backbone feature extractor in this stage.

\parsection{Training details} We train on images pairs of size $520 \times 520$. The first training stage involves 350k iterations, with a batch size of 15. The learning rate is initially set to $10^{-4}$, and halved after 133k and 240k iterations. 
For the second training stage, the batch size is reduced to 10 due to the added memory consumption when fine-tuning the backbone feature extractor. We train for 225k iterations in this stage. The initial learning rate is set to $5 \cdot 10^{-5}$ and halved after 210k iterations. The total training takes about 10 days on two NVIDIA TITAN RTX with 24GB of memory. 
For the GOCor modules~\cite{GOCor}, we train with 3 local and global optimization iterations. 

\parsection{Differences to PDC-Net} PDC-Net+ and PDC-Net use the same architecture and probabilistic formulation. Their training strategies differ as follows. In the first stage of training, only a single independently moving object is added to the training image pairs for PDC-Net as opposed to four for PDC-Net+. Moreover, PDC-Net is trained without any mask for the self-supervised part, while the injective mask (Sec.~\ref{sec:injective-mask}) is used for PDC-Net+. 
Finally, during the second stage of training, the backbone weights are finetuned with the same learning rate as the rest of the network weights in PDC-Net+, while it was divided by five in PDC-Net. 
PDC-Net+ is also trained for a total of 575k iterations against 330k for PDC-Net. This is mostly due to the introduction of multiple objects in the first training stage, which substantially increases the difficulty of the data, thus requiring longer training.

\subsection{Geometric Correspondences and Flow}
\label{subsec:correspondence-est}

We first evaluate our PDC-Net+ in terms of the quality of the predicted flow field.

\subsubsection{Datasets} 
We evaluate our approach on three standard datasets with sparse ground-truth, namely the \textbf{RobotCar}~\cite{RobotCar, RobotCarDatasetIJRR}, \textbf{MegaDepth}~\cite{megadepth} and \textbf{ETH3D}~\cite{ETH3d} datasets, as well as the homography estimation dataset \textbf{HPatches}. 

\parsection{MegaDepth} MegaDepth is a large-scale dataset, containing image pairs with extreme viewpoint and appearance variations. We follow the same procedure and 1600 test images as~\cite{RANSAC-flow}. It results in approximately 367K correspondences. Following~\cite{RANSAC-flow}, all the images and ground-truths are resized such that the smallest dimension has 480 pixels.  As metrics, we employ the Percentage of Correct Keypoints at a given pixel threshold $T$ (PCK-$T$).

\parsection{RobotCar} The RobotCar dataset depicts outdoor road scenes, taken under different weather and lighting conditions. Images are particularly challenging due to their numerous textureless regions. 
Following the protocol used in~\cite{RANSAC-flow}, images and ground-truths are resized such that the smallest dimension has the length $480$. The correspondences originally introduced by~\cite{RobotCarDatasetIJRR} are used as ground-truth, consisting of approximately 340M matches. We employ the Percentage of Correct Keypoints at a given pixel threshold $T$ (PCK-$T$) as metrics.

\parsection{ETH3D}  Finally, ETH3D represents indoor and outdoor scenes captured from a moving hand-held camera. It contains 10 image sequences at $480 \times 752$ or $514 \times 955$ resolution, depicting indoor and outdoor scenes.
We follow the protocol of~\cite{GLUNet}, sampling image pairs at different intervals to analyze varying magnitude of geometric transformations, resulting in about 500 image pairs and 600k to 1100k matches per interval. As evaluation metrics, we use the Average End-Point Error (AEPE) and Percentage of Correct Keypoints (PCK). 

\parsection{HPatches} The HPatches dataset~\cite{Lenc} depicts planar scenes divided in sequences, with transformations restricted to homographies.  Each image sequence contains a query image and 5 reference images taken under increasingly larger viewpoints changes. 
In line with~\cite{Melekhov2019, RANSAC-flow}, we exclude the sequences labelled \verb|i_X|, which only have illumination changes and employ only the 59 sequences labelled with \verb|v_X|, which have viewpoint changes.  It results in a total of 295 image pairs with dense ground-truth flow fields. Following~\cite{Melekhov2019, RANSAC-flow}, we evaluate on images and ground-truths resized to $240 \times 240$ and employ the AEPE and PCK as the evaluation metrics.

\begin{figure}[b]
\centering%
\vspace{-3mm}
\includegraphics[width=0.99\columnwidth, trim=135 0 0 0]{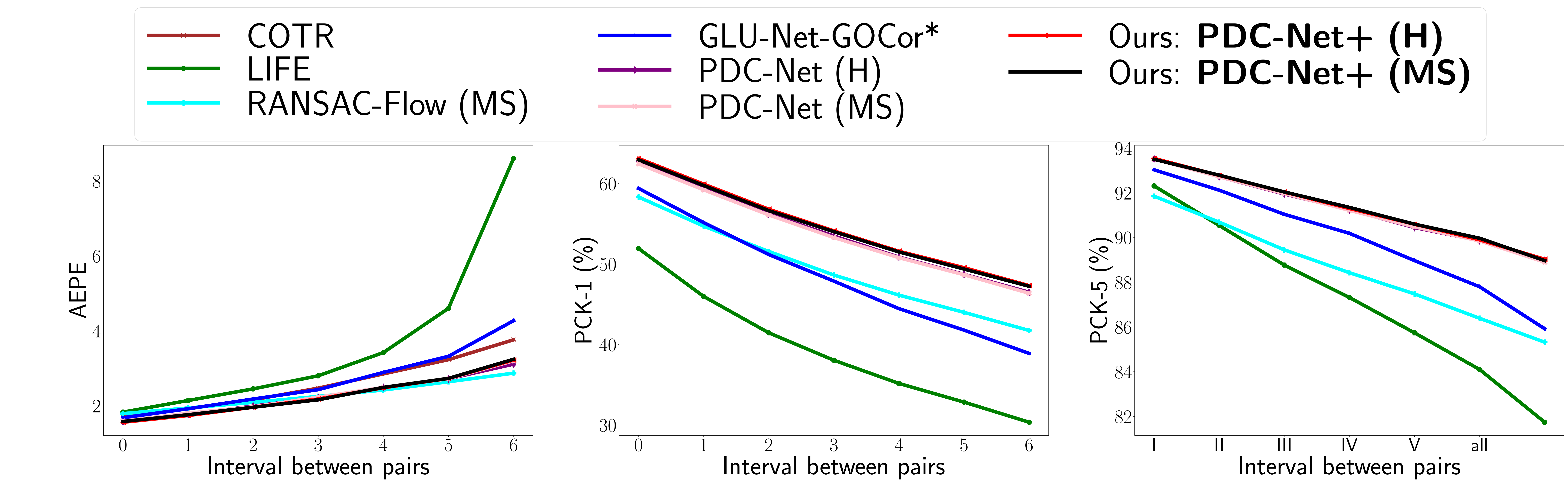}
\vspace{-3mm}
\caption{Results on ETH3D~\cite{ETH3d}.  AEPE (left), PCK-1 (center) and PCK-5 (right) are plotted w.r.t.\ the inter-frame interval length.}
\label{fig:ETH3D}
\end{figure}

\subsubsection{Compared methods} 
We compare to dense matching methods, trained for the geometric matching task. SIFT-Flow~\cite{LiuYT11}, NC-Net~\cite{Rocco2018b}, DGC-Net~\cite{Melekhov2019}, GLU-Net~\cite{GLUNet} and GLU-Net-GOCor~\cite{GLUNet, GOCor} are all trained on self-supervised data from different image sources than MegaDepth. The second set of compared methods are trained on MegaDepth images, namely RANSAC-Flow~\cite{RANSAC-flow}, LIFE~\cite{LIFE}, COTR~\cite{COTR} and PDC-Net~\cite{pdcnet}. Note that COTR cannot be evaluated on the MegaDepth test split because its training scenes overlap with the test scenes.
We also compare with the non-probabilistic baseline GLU-Net-GOCor*, which was introduced alongside the initial PDC-Net~\cite{pdcnet}. It is trained using the same settings and data as PDC-Net, but without the probabilistic formulation.

\subsubsection{Results} 
In Tab.~\ref{tab:megadepth} we report the results on MegaDepth and RobotCar. Our method PDC-Net+ outperforms all previous works by a large margin at all PCK thresholds.
Remarkably, our PDC-Net+ with a single forward pass (D) is \emph{40 times faster} than the recent RANSAC-Flow while being significantly better.
Our uncertainty-aware probabilistic approach PDC-Net+ also outperforms the baseline GLU-Net-GOCor* in flow accuracy. 
This clearly demonstrates the advantages of casting the flow estimation as a probabilistic regression problem, advantages which are not limited to uncertainty estimation. It also substantially benefits the accuracy of the flow itself through a more flexible loss formulation.  A visual comparison between PDC-Net+ and baseline GLU-Net-GOCor* on a MegaDepth image pair is shown in Fig.~\ref{fig:intro}, top. 

In Fig.~\ref{fig:ETH3D}, we plot the AEPE and PCKs on ETH3D. Our approach significantly outperforms previous approaches for all intervals in terms of PCK. As for AEPE, PDC-Net+ improves upon all methods, including the recent Transformer-based architecture COTR~\cite{COTR}, for intervals up to 11. It nevertheless obtains slightly worse AEPE than RANSAC-Flow for intervals of 13 and 15. However, RANSAC-Flow uses an extensive multi-scale scheme relying on off-the-shelf features extracted at different resolutions. In contrast, our multi-scale approach (MS) is much simpler and effective.

\begin{table}[t]
\centering
\caption{PCK (\%) results on sparse correspondences of the MegaDepth~\cite{megadepth} and RobotCar~\cite{RobotCar, RobotCarDatasetIJRR} datasets. In the top part, methods are trained on different data than MegaDepth while approaches in the bottom part are trained on the MegaDepth training set. We additionally compare the run-time of all methods on $480 \times 480$ images on a NVIDIA GeForce RTX 2080 Ti GPU.
}
\vspace{-3mm}
\resizebox{\columnwidth}{!}{%
\begin{tabular}{lccc|ccc|c} \toprule
& \multicolumn{3}{c}{\textbf{MegaDepth}} & \multicolumn{3}{c}{\textbf{RobotCar}} & Run-\\
& PCK-1 $\uparrow$  & PCK-3 $\uparrow$  & PCK-5 $\uparrow$ & PCK-1 $\uparrow$ & PCK-3 $\uparrow$ & PCK-5 $\uparrow$ & time (ms) $\downarrow$\\ \midrule
SIFT-Flow~\cite{LiuYT11} & 8.70 & 12.19 & 13.30 & 1.12 & 8.13 & 16.45 & -\\
NCNet~\cite{Rocco2018b} & 1.98 & 14.47 & 32.80 & 0.81 & 7.13 & 16.93 & - \\
DGC-Net~\cite{Melekhov2019} & 3.55 & 20.33 & 32.28 & 1.19 & 9.35 & 20.17 & \textbf{25} \\
GLU-Net$_{static}$~\cite{GLUNet, GOCor} & 29.27 & 50.46 & 55.93  & 2.21 & 17.06 & 33.69 & 36 \\
GLU-Net$_{dyn}$~\cite{GLUNet} & 21.58 & 52.18 & 61.78 & 2.30 & 17.15 & 33.87 & 36 \\
GLU-Net-GOCor$_{dyn}$~\cite{GOCor} &  37.28 &  61.18  & 68.08 & 2.31 & 17.62 & 35.18 & 70 \\ 
\midrule
RANSAC-Flow (MS)~\cite{RANSAC-flow} & 53.47 & 83.45 & 86.81 & 2.10 & 16.07 & 31.66 & 3596 \\
LIFE~\cite{LIFE} & 39.98 & 76.14 & 83.14 & 2.30 & 17.40 & 34.30 & 78 \\
GLU-Net-GOCor*~\cite{pdcnet} & 57.77 & 78.61 & 82.24 & 2.33 & 17.21 & 33.67 & \textbf{71} \\
PDC-Net (D)~\cite{pdcnet} & 68.95 & 84.07 & 85.72 & 25.39 & 18.96 & 36.36 & 88 \\
PDC-Net (H)~\cite{pdcnet} & 70.75 & 86.51 & 88.00 & 2.54 & 18.97 & 36.37 & 284 \\
PDC-Net (MS)~\cite{pdcnet} & 71.81 & 89.36 & 91.18 & 2.58 & 18.87 & 36.19 & 1017 \\
\textbf{PDC-Net+ (D)} & 72.41 & 86.70 & 88.12 & 2.57 & \textbf{19.12} & \textbf{36.71} & 88 \\
\textbf{PDC-Net+ (H)} & 73.92 & 89.21 & 90.48 & 2.56 & 19.00 & 36.56 & 284 \\
\textbf{PDC-Net+ (MS)} & \textbf{74.51} & \textbf{90.69} & \textbf{92.10} & \textbf{2.63} & 19.01 & 36.57 & 1017 \\
\bottomrule
\end{tabular}%
}\vspace{1mm}\vspace{-4mm}
\label{tab:megadepth}
\end{table}

\begin{table}[b]
\centering
\caption{Dense correspondence estimation results on HPatches~\cite{Lenc}. Here, the images and ground-truth flow fields are resized to $240 \times 240$, following~\cite{Melekhov2019}.}\label{tab:hp-240}
\vspace{-3mm}
\resizebox{0.34\textwidth}{!}{%
\begin{tabular}{lccc}
\toprule
& AEPE $\downarrow$ & PCK-1 (\%)  $\uparrow$ &  PCK-5 (\%) $\uparrow$ \\ \midrule
DGC-Net & 9.07 & 50.01 & 77.4\\
GLU-Net & 7.4 & 59.92 & 83.47\\
GLU-Net-GOCor & 6.62 & 58.45 & 85.89 \\
LIFE &  4.3  & 61.36 & 91.94\\
RANSAC-Flow (MS) & 3.79 & 78.42 &96.06\\
GLU-Net-GOCor* & 5.06 & 64.8  &90.24\\
PDC-Net (H) & 4.32 & 85.97  &94.59\\
PDC-Net (MS)& \textbf{3.56} & 87.4  &96.18 \\
\textbf{PDC-Net+ (H)} & 4.29 & 86.56  &94.97\\
\textbf{PDC-Net+ (MS)} & 3.59 & \textbf{87.83} &\textbf{96.36}\\ \midrule
\end{tabular}%
}
\end{table}

We also present results on HPatches in Tab.~\ref{tab:hp-240}. Our PDC-Net+ (H) outperforms all direct approaches in accuracy (PCK) and in robustness (AEPE). Moreover, PDC-Net+ with our multi-scale strategy (MS) obtains better PCK and AEPE results than RANSAC-Flow (MS), whereas it adopts an additional pre-processing step to identify rotated images by choosing the homography with the higher percentage of inliers between pairs of rotated images. 

\subsubsection{Generalization to optical flow} 

We additionally show that our approach generalizes well to the accurate estimation of optical flow, even though it is trained for the very different task of geometric matching.
We use the established \textbf{KITTI} dataset~\cite{Geiger2013}, and evaluate according to the standard metrics, namely AEPE and Fl. 
Since we do not fine-tune on KITTI, we show results on the training splits. In Tab.~\ref{tab:optical-flow}, we compare to methods specifically designed for optical flow~\cite{Sun2018, Sun2018, GOCor, Hui2018, YinDY19, Hui2019, VCN, RAFT} and trained using task-specific datasets~\cite{Dosovitskiy2015,Ilg2017a}. We also compare with more generic dense networks trained on other datasets.
Our approach outperforms all previous generic matching methods (upper part in Tab.~\ref{tab:optical-flow}) by a large margin in terms of both Fl and AEPE. Compared to PDC-Net, the training strategy introduced in PDC-Net+, that leverages multiple objects (Sec.~\ref{sec:training-strategy}) as well as the injective mask (Sec.~\ref{sec:injective-mask}), is highly beneficial for the optical flow task, where independently moving objects are common. 
Our PDC-Net+ also significantly outperforms the specialized optical flow methods (bottom part in Tab.~\ref{tab:optical-flow}), even though it is not trained on any optical flow datasets.

\begin{table}[t]
\centering
\caption{Optical flow results on the training splits of KITTI~\cite{Geiger2013}. The upper part contains generic matching networks, while the bottom part lists specialized optical flow methods, not trained on KITTI. }
\vspace{-3mm}\resizebox{0.35\textwidth}{!}{%
\begin{tabular}{lcc|cc}
\toprule
             & \multicolumn{2}{c}{\textbf{KITTI-2012}} & \multicolumn{2}{c}{\textbf{KITTI-2015}} \\ 
  & AEPE  $\downarrow$            & Fl   (\%)   $\downarrow$      & AEPE  $\downarrow$              & Fl  (\%)  $\downarrow$ \\ \midrule
DGC-Net~\cite{Melekhov2019}  &  8.50  &   32.28 &  14.97   &      50.98 \\
GLU-Net~\cite{GLUNet} & 3.14 & 19.76 & 7.49 & 33.83 \\
GLU-Net-GOCor$_{\textit{dyn}}$~\cite{GOCor} & 2.68 & 15.43 & 6.68 & 27.57 \\ 
RANSAC-Flow~\cite{RANSAC-flow} & - & - &  12.48 & - \\
GLU-Net-GOCor* & 2.26 & 9.89 & 5.53 & 18.27 \\
LIFE &  2.59 & 12.94  & 8.30 & 26.05 \\
COTR & 2.26 & 10.50 & 6.12 & 16.90 \\
PDC-Net (D) & 2.08 & 7.98 & 5.22 & 15.13 \\
\textbf{PDC-Net+ (D)} & \textbf{1.76} & \textbf{6.60} & \textbf{4.53} & \textbf{12.62} \\
\midrule 
PWC-Net~\cite{Sun2018} & 4.14 & 21.38 & 10.35 & 33.7  \\
PWC-Net-GOCor~\cite{Sun2018, GOCor} & 4.12 & 19.31 & 10.33 & 30.53 \\
LiteFlowNet~\cite{Hui2018} & 4.00 & - &  10.39 & 28.5  \\
HD$^3$F~\cite{YinDY19} & 4.65 & - & 13.17 & 24.0 \\
LiteFlowNet2~\cite{Hui2019} & 3.42 & - & 8.97 & 25.9 \\
VCN~\cite{VCN} & - & - & 8.36 & 25.1 \\
RAFT~\cite{RAFT} & - & - &  5.04 & 17.4  \\
\bottomrule
\end{tabular}%
}
\label{tab:optical-flow}
\end{table} 
\vspace{-3mm}

\begin{figure*}[t]
\centering
\newcommand{\wid}{0.9\textwidth}
\includegraphics[width=\wid]{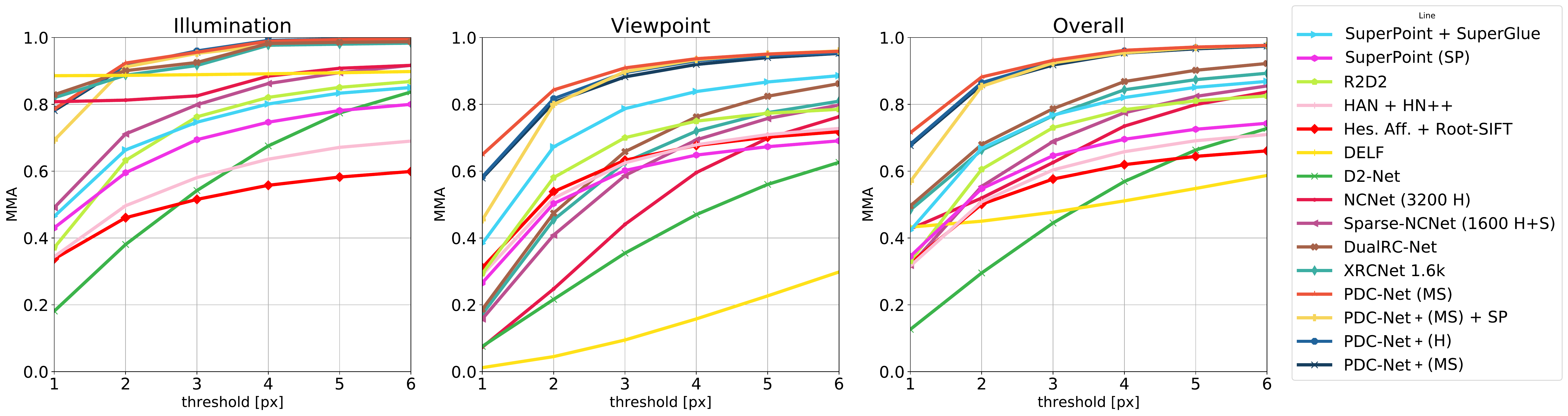}
\vspace{-3mm}
\caption{Sparse correspondence identification on HPatches~\cite{Lenc}, evaluating the 2000 most confident matches for each method.}
\label{fig:sparse-hp}
\end{figure*}

\subsection{Uncertainty Estimation} 
\label{subsec:uncertainty-est}

Next, we evaluate the quality of the uncertainty estimates provided by our approach. 

\parsection{Sparsification and error curves} 
To assess the quality of the uncertainty estimates, we rely on sparsification plots, in line with~\cite{Aodha2013LearningAC,Ilg2017a, ProbFlow}. 
The pixels having the highest uncertainty are progressively removed and the AEPE or PCK of the remaining pixels is plotted in the sparsification curve. These plots reveal how well the estimated uncertainty relates to the true errors. Ideally, larger uncertainty should correspond to larger errors. Gradually removing the predictions with the highest uncertainties should therefore monotonically improve the accuracy of the remaining correspondences. The sparsification plot is compared with the best possible ranking of the predictions, according to their actual errors computed with respect to the ground-truth flow. We refer to this curve as the oracle plot.

Note that, for each model the oracle is different. Hence, an evaluation using a single sparsification plot is not possible. To this end, we use the Sparsification Error, constructed by directly comparing each sparsification plot to its corresponding oracle plot by taking their difference. Since this measure is independent of the oracle, a fair comparison between different methods is possible. As evaluation metric, we use the Area Under the Sparsification Error curve (AUSE). 

The sparsification and error plots provide an insightful and directly relevant assessment of the uncertainty. The AUSE directly evaluates the ability to filter out inaccurate and incorrect correspondences, which is the main purpose of the uncertainty estimate in \eg pose estimation and 3D reconstruction. Furthermore, sparsification and error plots are also applicable to non-probabilistic confidence and uncertainty estimation techniques, allowing for a direct comparison with previous work~\cite{RANSAC-flow,Melekhov2019}.

\begin{figure}
\centering
\newcommand{\wid}{0.23\textwidth}
\subfloat{\includegraphics[width=\wid]{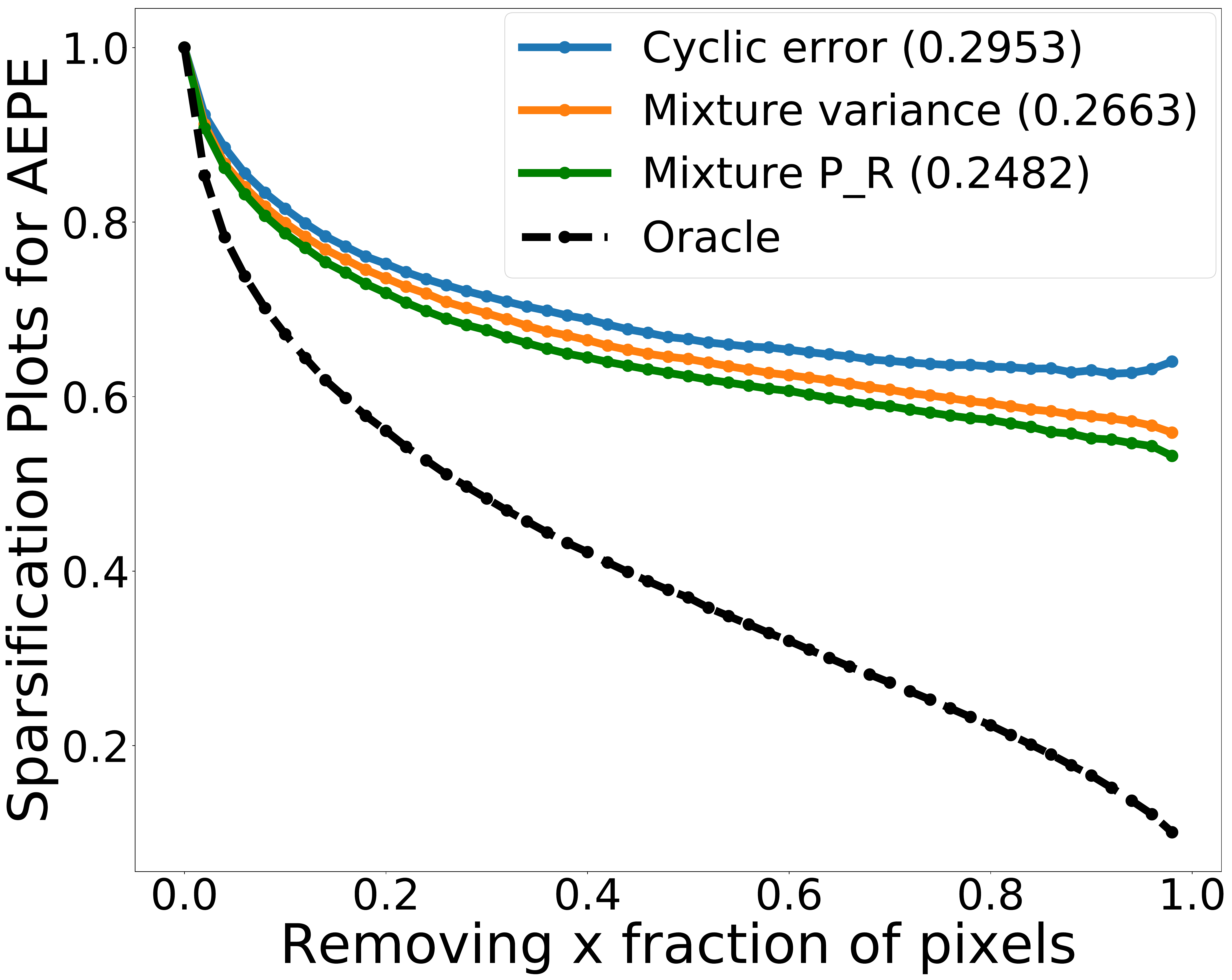}}~%
\subfloat{\includegraphics[width=\wid]{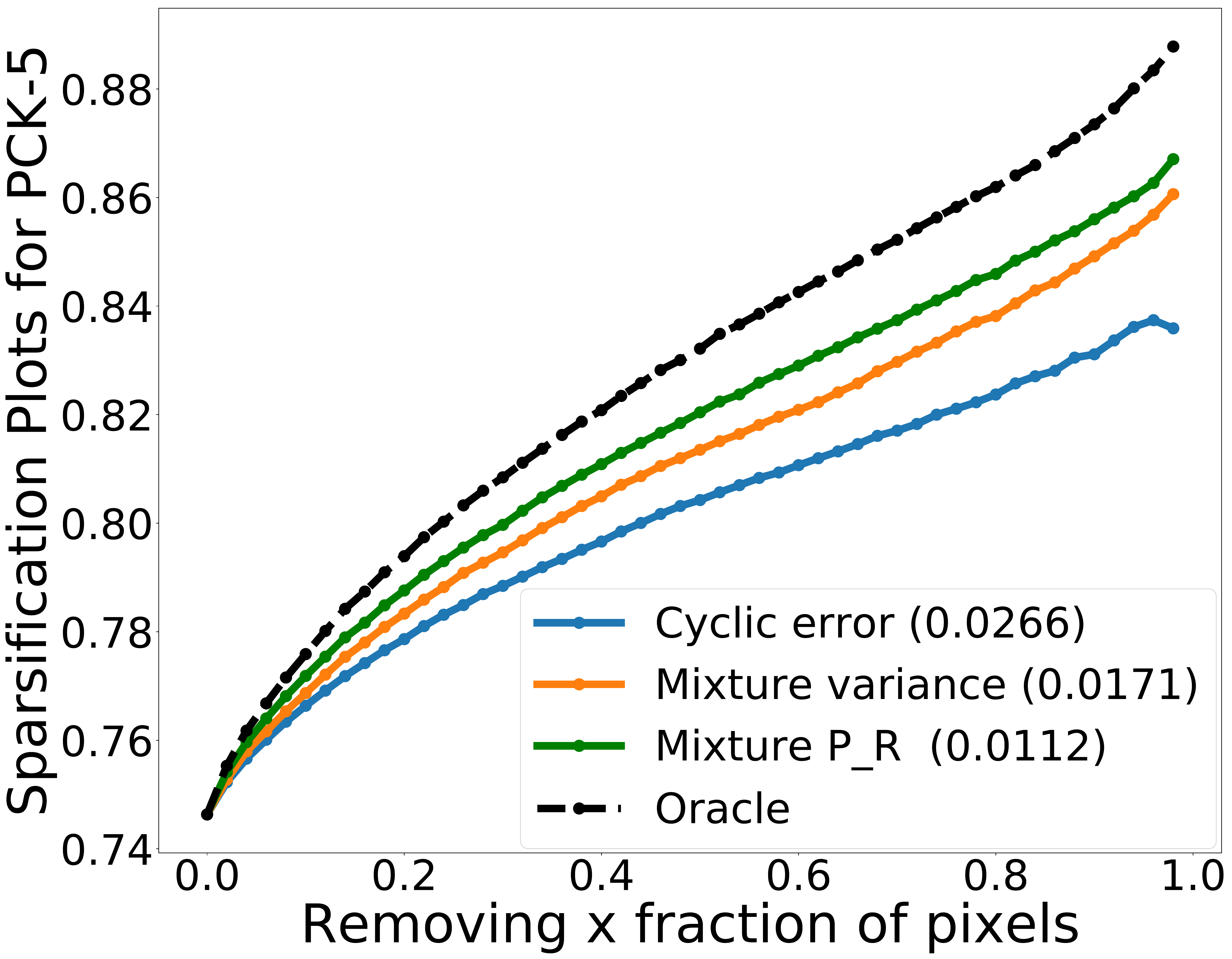}} \\
\subfloat{\includegraphics[width=\wid]{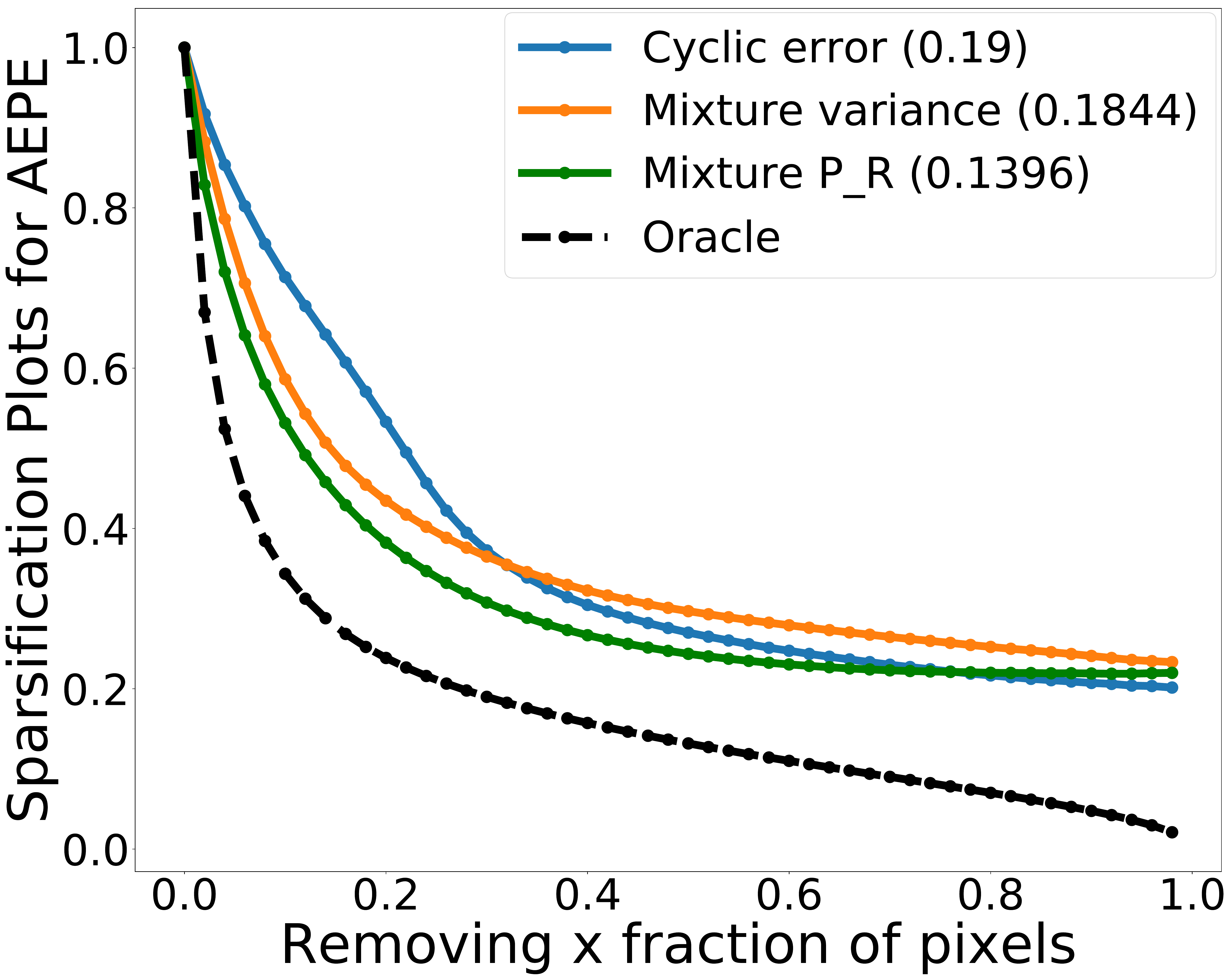}}~%
\subfloat{\includegraphics[width=\wid]{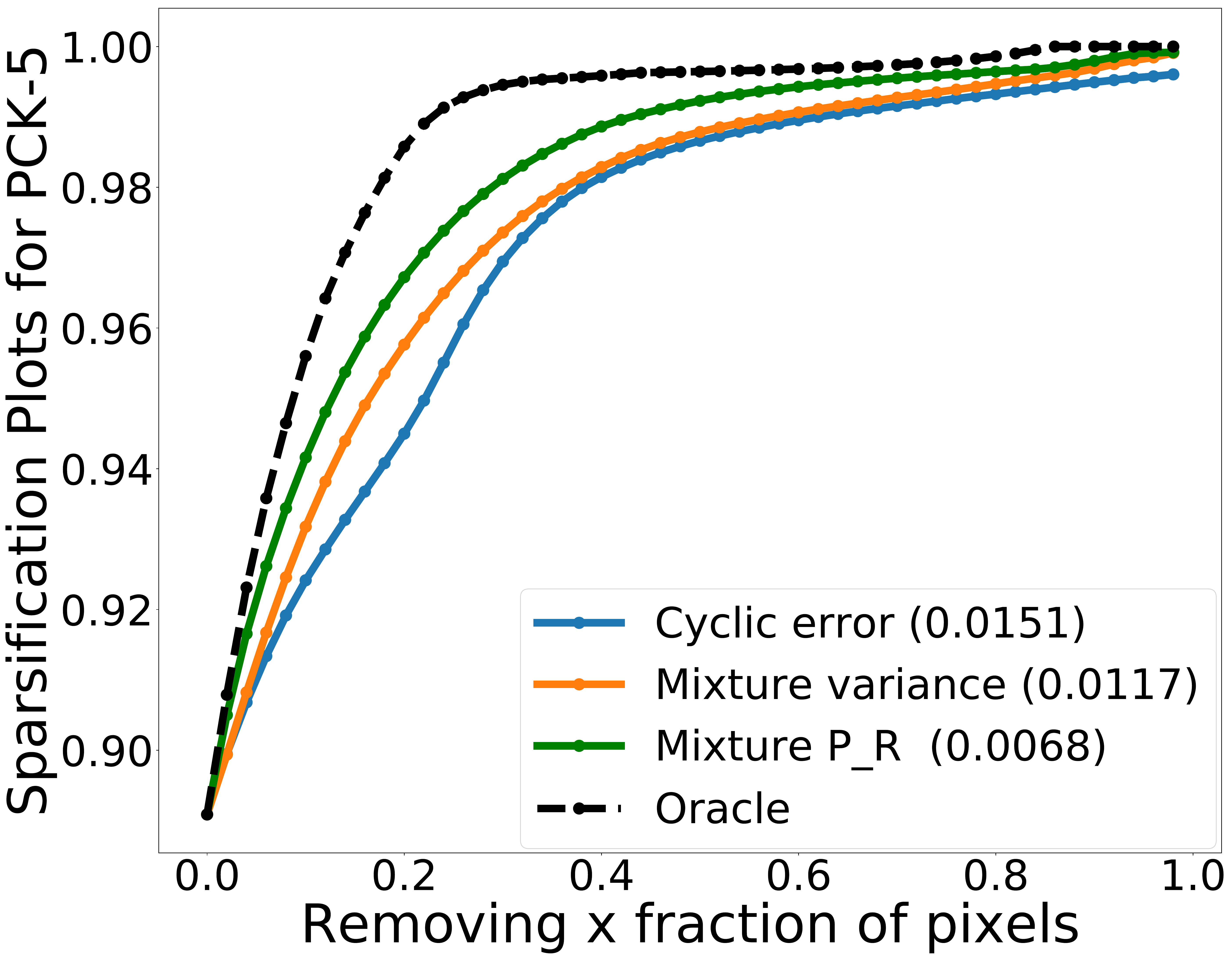}}~%
\vspace{-2mm}
\caption{Sparsification plots for AEPE (left) and PCK-5 (right) on MegaDepth (top) and KITTI-2015 (bottom), comparing multiple uncertainty measures when using PDC-Net+ (D).  Smaller AUSE (in parenthesis) is better.}
\label{fig:sparsification-unc}
\vspace{-4mm}
\end{figure}

\parsection{Comparison of uncertainty measures} In Fig.~\ref{fig:sparsification-unc}, we report the sparsification plots and AUSE comparing different uncertainty measures applied to the probabilistic formulation of our PDC-Net+. In particular, we compare our confidence estimate $P_R$ \eqref{eq:pr} with computing the variance of the mixture model $V = \sum_{m=1}^M \alpha_m \sigma^2_m$ and the forward-backward consistency error~\cite{Meister2017}. The latter is a common approach to rank and filter matches.  Since the same flow regression model is used for all uncertainty measures, the sparsification plots and the Oracle are directly comparable. Using $P_R$ leads to sparsification plots closer to the Oracle (and smaller AUSE) on both MegaDepth and KITTI-2015. When using the PCK-5 metric on MegaDepth, our $P_R$ confidence reduces the AUSE by $58\%$ compared to forward-backward consistency and $35\%$ compared to using the mixture variance.
Moreover, as observed in Fig.~\ref{fig:sparsification-unc}, our approach reduces the AEPE by 30 and 70 \% respectively on MegaDepth and KITTI-2015 by removing only 30 \% of the most uncertain matches.

\begin{figure}[b]
\centering
\vspace{-6mm}\newcommand{\wid}{0.23\textwidth}
\subfloat{\includegraphics[width=\wid]{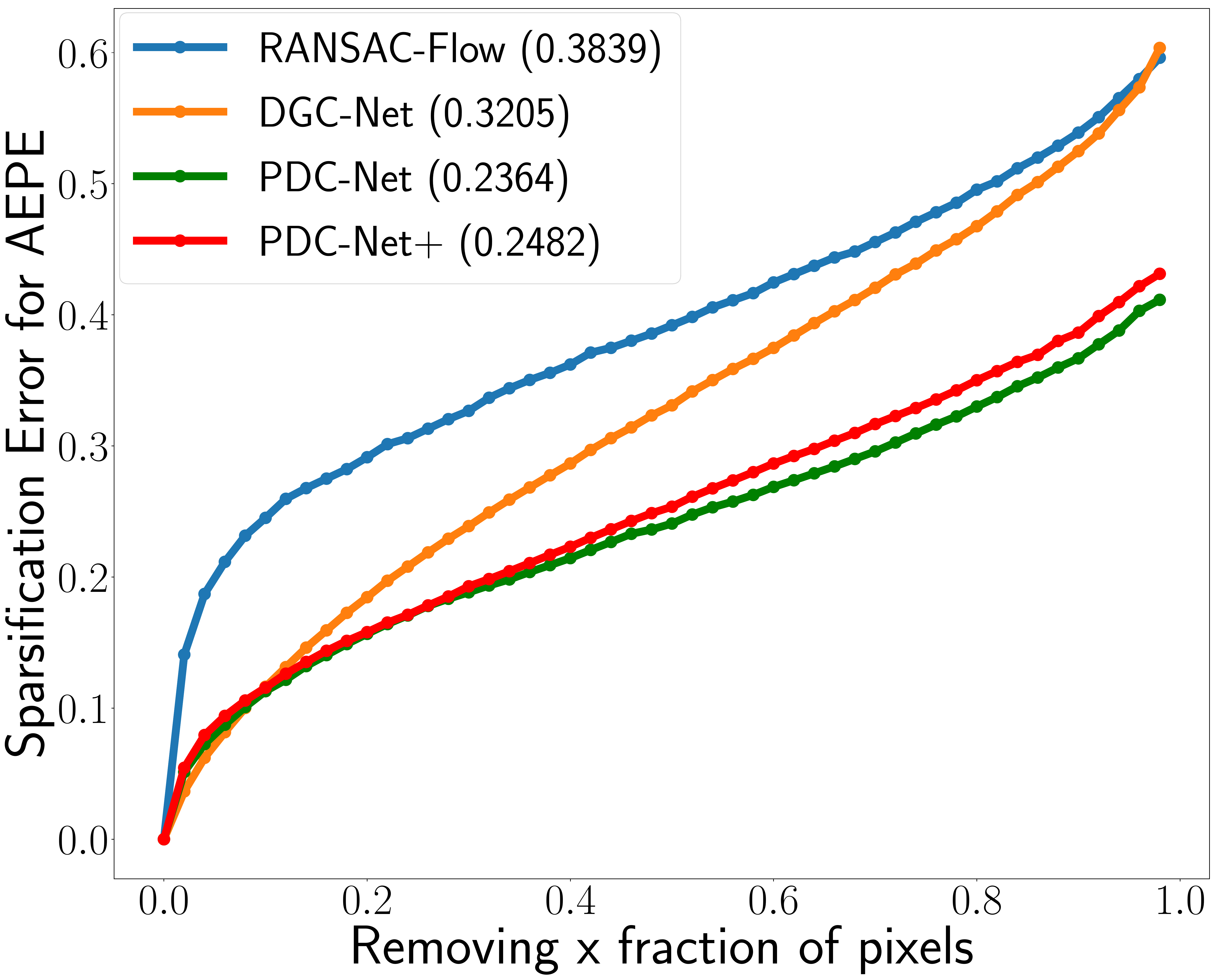}}~%
\subfloat{\includegraphics[width=\wid]{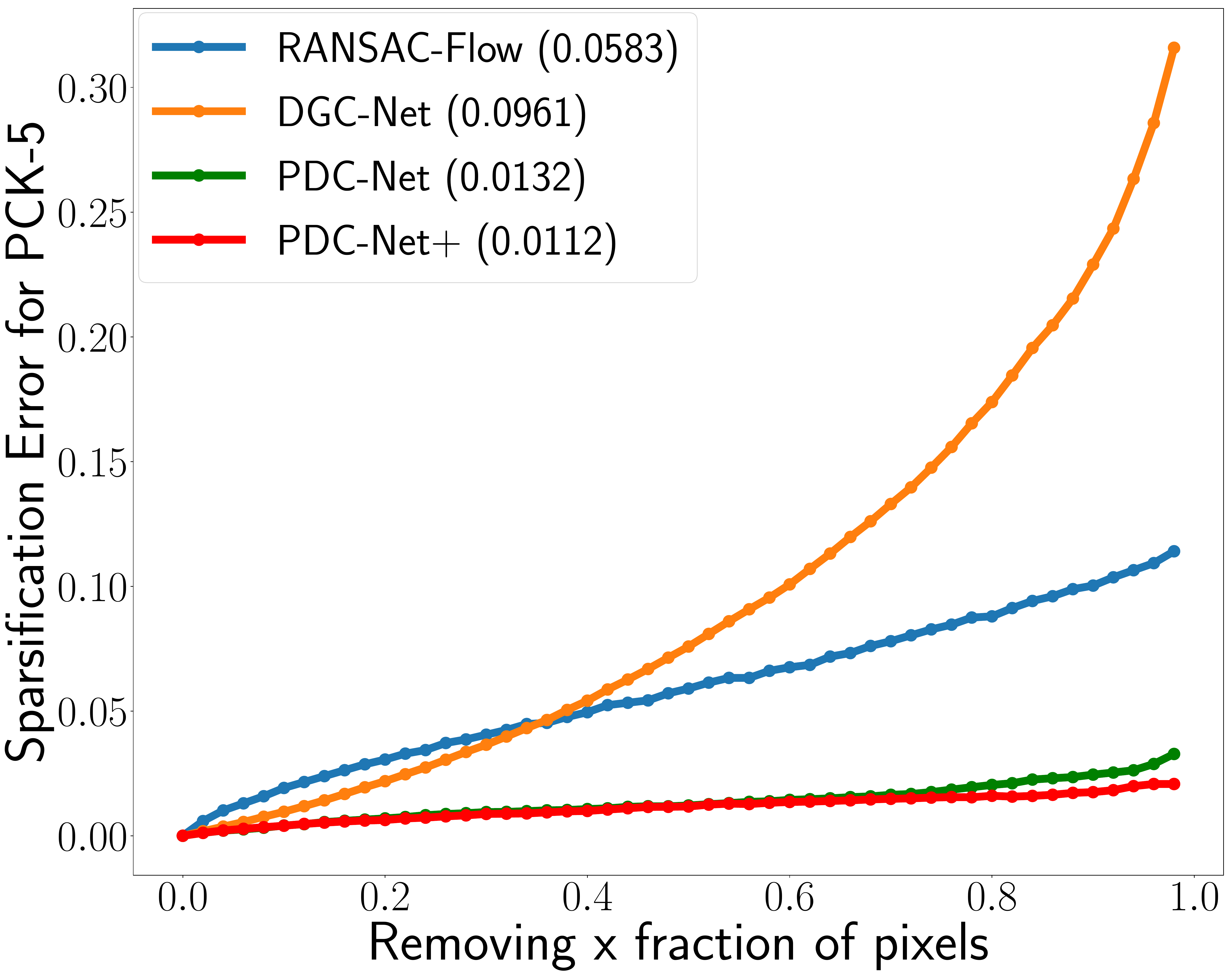}}~%
\vspace{-2mm}
\caption{Sparsification Error plots for AEPE (left) and PCK-5 (right) on MegaDepth, comparing different methods predicting the uncertainty of the match.  Smaller AUSE (in parenthesis) is better. }

\label{fig:sparsification}
\end{figure}

\parsection{Comparison to State-of-the-art} Fig.~\ref{fig:sparsification} shows the Sparsification Error plots on MegaDepth of our PDC-Net+ and PDC-Net~\cite{pdcnet}, compared to previous dense methods that provide a confidence estimation, namely DGC-Net~\cite{Melekhov2019} and RANSAC-Flow~\cite{RANSAC-flow}. PDC-Net+ and PDC-Net outperform other methods in AUSE, verifying the robustness of the predicted uncertainty estimates. In addition to the large improvements in accuracy, PDC-Net+ largely preserves the quality of the uncertainty estimates achieved by the original PDC-Net.

\parsection{Sparse correspondence evaluation on HPatches}
Finally, we evaluate the effectiveness of our PDC-Net+ for sparse correspondence identification using the HPatches dataset~\cite{Lenc}. We follow the standard protocol introduced by D2-Net~\cite{Dusmanu2019CVPR}, where 108 of the 116 image sequences are evaluated, with 56 and 52 sequences depicting viewpoint and illumination changes, respectively. 
The first image of each sequence is matched against the other five, giving 540 pairs. As metric, we use the Mean Matching Accuracy (MMA), which estimates the average number of correct matches for different pixel thresholds. For each method, the top 2000 matches are selected for evaluation, following~\cite{Rocco20, Xreo, DualRCNet}. 

For PDC-Net+, we first resize the images so that the size along the smallest dimension is 600 and then predict the dense flow relating the pair. We rank the estimated correspondences according to their predicted confidence map $P_{R=1}$ \eqref{eq:pr} and select the 2000 most confident for evaluation. 
We also evaluate the performance of PDC-Net+ in combination with the SuperPoint detector~\cite{superpoint}, referred to as `PDC-Net+ + SP'. To obtain matches relating the sets of sparse keypoints, we follow the procedure introduced in Sec.~\ref{sec:geometric-inference} with a maximum distance to keypoints of $d=4$. We use the default parameters for SuperPoint, with a Non-Max Suppression (NMS) window of 4. The matches are also ranked according to their predicted confidence map $P_{R=1}$ \eqref{eq:pr}. 

We compare our approach to recent detect-then-describe baselines: SuperPoint~\cite{superpoint}, SuperPoint + SuperGlue~\cite{SarlinDMR20}, R2D2~\cite{R2D2descriptor}, D2-Net~\cite{Dusmanu2019CVPR} and DELF~\cite{DELF}. We also compare with dense-to-sparse approaches: Sparse-NCNet~\cite{Rocco20}, DualRC-Net~\cite{DualRCNet} and XRCNet~\cite{Xreo}, along with the purely dense method NC-Net~\cite{Rocco2018b}. 
As shown in Fig.~\ref{fig:sparse-hp}, our initial approach PDC-Net sets a new state-of-the-art at all thresholds overall, closely followed by PDC-Net+. Remarkably, PDC-Net+ outperforms the next best method by approximately 20\% at a threshold of 1 pixel. As opposed to other methods, our approach is robust to both illumination and view-point changes. 
Moreover, PDC-Net+ in association with SuperPoint (PDC-Net+ + SP) also drastically outperforms SuperPoint and SuperPoint + SuperGlue. It demonstrates that our dense matching approach is also a strong alternative for establishing sparse matches between two images.

\subsection{Pose Estimation}
\label{subsec:down-stream-tasks}

To show the joint performance of our flow and uncertainty prediction, we evaluate our approach for the task of pose estimation. Given a pair of images showing different viewpoints of the same scene, two-view geometry estimation aims at recovering their relative pose. Although it has long been dominated by sparse matching methods, we here evaluate our dense approach for this task.

\begin{table}[b]
\centering
\vspace{-1mm} 
\caption{Two-view geometry estimation on the outdoor dataset YFCC100M~\cite{YFCC}. The top section compares sparse methods while the bottom section contains dense methods.}
\vspace{-2mm}
\resizebox{\columnwidth}{!}{%
\begin{tabular}{lccc|ccc}
\toprule
             & \multicolumn{3}{c}{AUC $\uparrow$} & \multicolumn{3}{c}{mAP $\uparrow$} \\ 
             & @5\textdegree & @10\textdegree & @20\textdegree &   @5\textdegree &   @10\textdegree & @20\textdegree   \\ \midrule
SIFT~\cite{SIFT} + ratio test & 24.09 & 40.71 & 58.14 & 45.12 & 55.81 & 67.20 \\
SIFT~\cite{SIFT} + OANet~\cite{OANet} & 29.15 & 48.12 & 65.08 & 55.06 & 64.97 & 74.83 \\
SIFT~\cite{SIFT} + SuperGlue~\cite{SarlinDMR20} & 30.49 & 51.29 & 69.72 & 59.25 & 70.38 & 80.44 \\
SuperPoint~\cite{superpoint} (SP) & - & - & - &  30.50 & 50.83 & 67.85  \\
SP~\cite{superpoint} + OANet~\cite{OANet} & 26.82 & 45.04 & 62.17 & 50.94 & 61.41 & 71.77 \\
SP~\cite{superpoint} + SuperGlue~\cite{SarlinDMR20} & \textbf{38.72} & \textbf{59.13} & \textbf{75.81} & \textbf{67.75} & \textbf{77.41} & \textbf{85.70} \\
\midrule          
D2D~\cite{D2D} & - & - & - & 55.58 & 66.79 & - \\
RANSAC-Flow (MS+SegNet)~\cite{RANSAC-flow} & - & - & - & 64.88 & 73.31 & 81.56  \\
RANSAC-Flow (MS)~\cite{RANSAC-flow} & - & - & - & 31.25 & 38.76 & 47.36  \\
PDC-Net (D) & 32.21 & 52.61 & 70.13 & 60.52 & 70.91 & 80.30 \\
PDC-Net (H) & 34.88  & 55.17 & 71.72 & 63.90 & 73.00 & 81.22  \\
PDC-Net (MS) & 35.71 & 55.81 & 72.26 & 65.18 & 74.21 & 82.42 \\ 
\textbf{PDC-Net+ (D)} & 34.76 & 55.37 & 72.55 & 63.93 & 73.81 & 82.74 \\
\textbf{PDC-Net+ (H)} & \textbf{37.51} & \textbf{58.08} & \textbf{74.50} & \textbf{67.35} & \textbf{76.56} & \textbf{84.56} \\
\bottomrule
\end{tabular}%
}\vspace{1mm}
\label{tab:YCCM}
\end{table}

\parsection{Datasets and metrics} We evaluate on both an outdoor and an indoor dataset, namely \textbf{YFCC100M}~\cite{YFCC} and \textbf{ScanNet}~\cite{scannet} respectively. 
The YFCC100M dataset contains images of touristic landmarks. The provided ground-truth poses were created by generating 3D reconstructions from a subset of the collections~\cite{heinly2015_reconstructing_the_world}.  We follow the standard set-up of~\cite{OANet} and evaluate on 4 scenes of the YFCC100M dataset, each comprising 1000 image pairs. 
ScanNet is a large-scale indoor dataset composed of monocular sequences with ground truth poses and depth images. We follow the set-up of~\cite{SarlinDMR20} and evaluate on 1500 image pairs.

\begin{table}[t]
\centering
\caption{Two-view geometry estimation on the indoor dataset ScanNet~\cite{scannet}. The top section compares sparse methods, the middle section present dense-to-sparse approaches, and the bottom one contains dense methods. The symbol $\dagger$~denotes that the network was not trained on ScanNet. }
\vspace{-2mm}
\resizebox{\columnwidth}{!}{%
\begin{tabular}{lccc|ccc}
\toprule
             & \multicolumn{3}{c}{AUC $\uparrow$} & \multicolumn{3}{c}{mAP $\uparrow$} \\ 
             & @5\textdegree & @10\textdegree & @20\textdegree &   @5\textdegree &   @10\textdegree & @20\textdegree   \\ \midrule
ORB~\cite{ORB} + GMS~\cite{GMS} & 5.21 & 13.65 & 25.36 & - & - & - \\
D2-Net~\cite{Dusmanu2019CVPR} + NN & 5.25 & 14.53 & 27.96 & - & - & - \\
ContextDesc~\cite{contextdesc} + Ratio Test & 6.64 & 15.01 & 25.75 & - & - & - \\
SuperPoint~\cite{superpoint} + OANet~\cite{OANet} & 11.76 & 26.90 & 43.85 & - & - & - \\
SuperPoint~\cite{superpoint} + SuperGlue~\cite{SarlinDMR20} & \textbf{16.16} & \textbf{33.81} & \textbf{51.84} & - & - & - \\
\midrule
DRC-Net \textdagger ~\cite{DualRCNet} & 7.69 & 17.93 & 30.49 & - & - & - \\
LoFTR-OT \textdagger~\cite{LOFTR} & 16.88 & 33.62 & 50.62 & - & - & - \\
LoFTR-OT~\cite{LOFTR}& 21.51 & 40.39 & 57.96 & - & - & -  \\
LoFTR-DT~\cite{LOFTR} & \textbf{22.02} & \textbf{40.8} & \textbf{57.62} & - & - & - \\ \midrule

PDC-Net \textdagger~\cite{pdcnet} (D) & 17.70 & 35.02 & 51.75 & 39.93 & 50.17 & 60.87 \\
PDC-Net \textdagger~\cite{pdcnet} (H) & 18.70 & 36.97 & 53.98 & 42.87 & 53.07 & 63.25 \\
PDC-Net \textdagger~\cite{pdcnet} (MS) & 18.44 & 36.80 & 53.68 & 42.40 & 52.83 & 63.13 \\
\textbf{PDC-Net+ \textdagger~(D)} & 19.02 & 36.90 & 54.25 & 42.93 & 53.13 & 63.95 \\
\textbf{PDC-Net+ \textdagger~(H)} & \textbf{20.25} & \textbf{39.37} & \textbf{57.13} & \textbf{45.66} & \textbf{56.67} & \textbf{67.07} \\ 
\bottomrule
\end{tabular}%
}\vspace{1mm}
\vspace{-3mm}\label{tab:scannet}
\end{table}

We use the predicted matches to estimate the Essential matrix with RANSAC~\cite{ransac} and the 5-point Nister algorithm~\cite{TPAMI.2004.17} with an inlier threshold of 1 pixel divided by the focal length. The rotation matrix $\hat{R}$ and translation vector $\hat{T}$ are finally computed from the estimated Essential matrix. As evaluation metrics, in line with~\cite{YiTOLSF18,SarlinDMR20}, we report the area under the cumulative pose error curve (AUC) at different thresholds (5\textdegree, 10\textdegree, 20\textdegree), where the pose error is the maximum of the angular deviation between ground truth and predicted vectors for both rotation and translation. For completeness, we also report the mean Average Precision (mAP) of the pose error for the same thresholds, following~\cite{OANet, RANSAC-flow}. 
 
\parsection{Matches selection for PDC-Net+} The original images are resized to have a minimum dimension of 480, similar to~\cite{RANSAC-flow}, and the intrinsic camera parameters are modified accordingly. Our PDC-Net+ network outputs flow fields at a quarter of the image's input resolution, which are then normally bi-linearly up-sampled. For pose estimation, we instead directly select matches at the outputted resolution and further scale them to the input image resolution. From the estimated dense flow field, we identify the accurate correspondences by thresholding the predicted confidence map $P_R$ \eqref{eq:pr} as $P_{R=1} > 0.1$, and use them for pose estimation. 

\begin{figure}[b]
\centering%
\vspace{-4mm}
\subfloat{\includegraphics[width=0.23\textwidth]{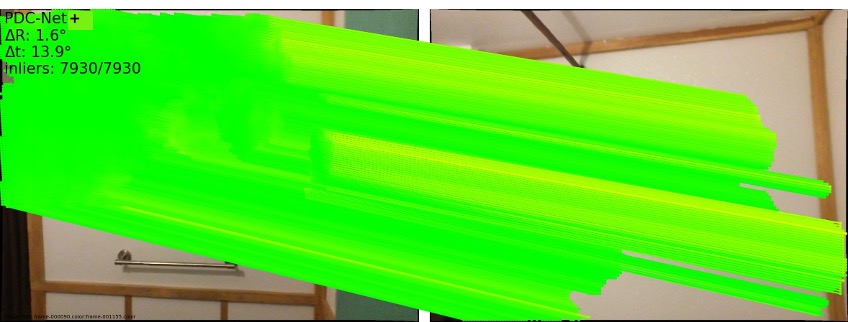} }~%
\subfloat{\includegraphics[width=0.23\textwidth]{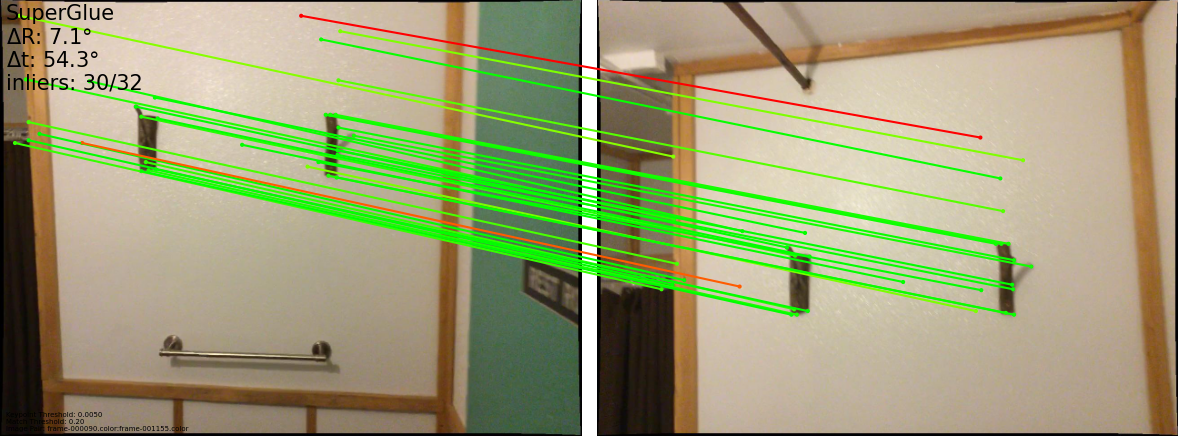} }~%
\vspace{-2mm}
\caption{Comparison of matches found by our PDC-Net+ (left) and Superpoint~\cite{superpoint} + SuperGlue~\cite{SarlinDMR20} (right) on an images pair of ScanNet~\cite{scannet}. Correct matches are green lines and mismatches are red lines.}
\label{fig:scannet-matches}
\end{figure}

\parsection{Results on outdoor} Results are presented in Tab.~\ref{tab:YCCM}. Our PDC-Net+ approach outperforms all dense methods by a large margin. We note that RANSAC-Flow relies on a semantic segmentation network to better filter unreliable correspondences, in \eg sky.  Without this segmentation, the performance is drastically reduced. In contrast, our approach can directly estimate highly robust and generalizable confidence maps, without the need for additional network components. The confidence masks of RANSAC-Flow and our approach are visually compared in Fig.~\ref{fig:arch-visual-sep-pertur}-\ref{fig:arch-visual-ransac}.
Interestingly, our dense PDC-Net+ approach also outperforms multiple standard sparse methods, such as SIFT~\cite{SIFT} and SuperPoint~\cite{superpoint}. Our approach is even competitive with the recent SuperGlue~\cite{SarlinDMR20}, which employs a graph-based network for sparse matching. 

\parsection{Results on indoor} Results on the ScanNet dataset are presented in Tab.~\ref{tab:scannet}. Our PDC-Net+ (H) approach scores second among all methods, slightly behind the very recent LoFTR~\cite{LOFTR}, which employs a Transformer-based architecture. However, this version of LoFTR is trained on ScanNet, while we only train on the outdoor MegaDepth dataset. Our approach thus demonstrates highly impressive generalization properties towards the very different indoor ScanNet dataset. When only considering approaches that are not trained on ScanNet, our PDC-Net+ (H) ranks first, with a significant improvement compared to LoFTR \textdagger. A visual example of our PDC-Net+ applied to a ScanNet image pair is presented in Fig.~\ref{fig:intro}, middle. 

We also note that our approach outperforms all the sparse methods, including SuperPoint + SuperGlue, which relies on a Transformer architecture and is trained specifically for ScanNet. 
The ScanNet dataset highlights the limitations of detector-based approaches. The images depict many homogeneous surfaces, such as white walls, on which detectors often fail. Dense approaches, such as ours, have an advantage in this context. A visual comparison between PDC-Net+ and SuperPoint + SuperGlue is shown in Fig.~\ref{fig:scannet-matches}.

\subsection{Image-based localization}

\begin{table}[b]
\centering
\vspace{-1mm} 
\caption{Visual Localization on the Aachen Day-Night dataset~\cite {SattlerWLK12, SattlerMTTHSSOP18}. We follow the fixed evaluation protocol of~\cite{SattlerMTTHSSOP18} and report the percentage of query images localized within $X$ meters and $Y^\circ$ of the ground-truth pose. }
\vspace{-2mm}
\resizebox{0.49\textwidth}{!}{%
\begin{tabular}{ll|ccc}
\toprule
   Method type &   Method        &  0.5m, 2\textdegree &  0.5m, 5\textdegree  & 5m, 10\textdegree  \\ \midrule
Sparse & D2-Net~\cite{Dusmanu2019CVPR} & 	74.5 & 86.7 & \textbf{100.0} \\
& R2D2~\cite{R2D2descriptor} & 69.4 & 86.7 & 94.9 \\ 
& R2D2~\cite{R2D2descriptor} (K=20k) & 76.5 & \textbf{90.8} & \textbf{100.0} \\
& SuperPoint~\cite{superpoint} & 73.5 & 79.6 &  88.8 \\
& SuperPoint~\cite{superpoint} + SuperGlue~\cite{SarlinDMR20} & \textbf{79.6} & \textbf{90.8} & \textbf{100.0} \\
& Patch2Pix & 78.6 & 88.8 & 99.0 \\ \midrule
Sparse-to-dense & Superpoint + S2DNet~\cite{GermainBL20} & 74.5 & 84.7 & 100.0 \\ \midrule

Dense-to-sparse  & Sparse-NCNet~\cite{Rocco20} & 76.5 & 84.7 & 98.0 \\
 & DualRC-Net~\cite{DualRCNet} & \textbf{79.6} & \textbf{88.8} & \textbf{100.0} \\
 & XRCNet-1600~\cite{Xreo} & 76.5 & 85.7 & \textbf{100.0} \\ \midrule
Dense  & RANSAC-flow (ImageNet)~\cite{RANSAC-flow} + Superpoint~\cite{superpoint} & 74.5 & 87.8 & \textbf{100.0} \\

(flow regression) & RANSAC-flow (MOCO)~\cite{RANSAC-flow} + Superpoint~\cite{superpoint} & 74.5 & 88.8 & \textbf{100.0} \\
& PDC-Net~\cite{pdcnet} & 76.5 & 85.7 & \textbf{100.0} \\
& PDC-Net~\cite{pdcnet} + SuperPoint~\cite{superpoint} & \textbf{80.6} & 87.8 & \textbf{100.0} \\
& \textbf{PDC-Net+} & \textbf{80.6} & 89.8 & \textbf{100.0} \\
& \textbf{PDC-Net+ + SuperPoint} & 79.6 & \textbf{90.8} & \textbf{100.0} \\
\bottomrule
\end{tabular}%
}
\label{tab:aachen}
\end{table}

Finally, we evaluate our approach for the task of image-based localization. 
Image-based localization aims at estimating the absolute 6 DoF pose of a query image with respect to a 3D model. It requires sets of accurate and localized matches as input.
We evaluate our approach on the \textbf{Aachen Benchmark}~\cite{SattlerMTTHSSOP18, SattlerWLK12}. It features 4,328 daytime reference images taken with a handheld smartphone, for which ground truth camera poses are provided. 

\parsection{Evaluation metric} Each estimated query pose is compared to the ground-truth pose, by computing the absolute orientation error $\left | R_{err}  \right |$ and the position error $T_{err}$. $R_{err}$ computes the angular deviation between the estimated and ground-truth rotation matrices. The position error $  T_{err} $ is measured as the Euclidean distance $ \left \|\hat{T} - T \right \|$ between the estimated $\hat{T}$ and the ground-truth position $T$.
Because the ground-truth poses are not publicly available, evaluation is done by submitting to the evaluation server of~\cite{SattlerMTTHSSOP18}. It then reports the percentage of query images localized within $X$ meters and $Y^\circ$ of the ground-truth pose, \ie $\left | R_{err}  \right | < Y \textrm{and } T_{err} < X $, for predefined $X$ and $Y$ thresholds.

\parsection{Image matching in fixed pipeline} We evaluate on the local feature challenge of the Visual Localization benchmark~\cite{SattlerMTTHSSOP18, SattlerWLK12}. We follow the evaluation protocol of~\cite{SattlerMTTHSSOP18} according to which up to 20 relevant day-time images with known camera poses are provided for each of the 98 night-time images. The aim is to localize the latter using only the 20 day-time images. We first compute matches between the day-time images, which are fed to COLMAP~\cite{SchonbergerF16} to build a 3D point cloud. We also provide matches between the queries and the provided short-list of retrieved images, which are used by COLMAP to compute a localisation estimate for the queries, through 2D-3D matches. 

For PDC-Net+ (and PDC-Net), we resize images so that their smallest dimension is 600. From the estimated dense flow fields, we select correspondences for which $P_{R=1} > 0.1$ at a quarter of the input image resolution and further scale and round them to the original image resolution. We compute matches from both directions and keep all the confident correspondences. 
Following~\cite{RANSAC-flow}, we also evaluate a version, referred to as PDC-Net+ + SuperPoint, for which we only use correspondences on a sparse set of keypoints using SuperPoint~\cite{superpoint}. We accept a match if the distance to the closest keypoint is below $d=4$ (see Sec.~\ref{sec:geometric-inference}) for an image size where the smallest dimension is 600. As additional requirement, matches must also have a corresponding confidence measure $P_{R=1} > 0.1$.  For SuperPoint, we use the default parameters with a Non-Max Suppression window of 4. We also compute matches from both directions and keep all the confident correspondences.

We present results in Table~\ref{tab:aachen}. We compare PDC-Net+ to multiple dense methods, namely RANSAC-Flow~\cite{RANSAC-flow} and PDC-Net~\cite{pdcnet}. For reference, we also report results of several recent sparse methods, as well as sparse-to-dense and dense-to-sparse approaches.
Using dense matches, our PDC-Net+ outperforms previous dense methods at all thresholds. In particular, we achieve a substantial performance gain of $6 \%$ compared to RANSAC-Flow on the most accurate threshold. 
Moreover, PDC-Net+ is also on par or better than all methods from the other categories. 

Interestingly, PDC-Net+ + SuperPoint also substantially improves upon SuperPoint as in Sec.~\ref{subsec:uncertainty-est}. It demonstrates that our dense flow for matching sparse sets of keypoints is a valid and better alternative than direct matching of descriptors.

\begin{table}[t]
\centering
\caption{Retrieval-based Localization on the Aachen Day-Night dataset~\cite{SattlerWLK12} for a pose error threshold of 5m, 10\textdegree \, in \%. Results for the retrieval methods are extracted from the official leaderboard~\cite{SattlerMTTHSSOP18}.}
\vspace{-3mm}
\resizebox{0.25\textwidth}{!}{%
\begin{tabular}{l|c|c}
\toprule
   Method     & \textbf{Day} & \textbf{Night} \\\midrule
NetVLAD~\cite{NetVLAD} &   18.9	&  14.3 \\
LLN~\cite{LLN}	&   20.8 &  17.3 \\
DenseVLAD~\cite{DenseVLAD} &  \textbf{22.8} &   \textbf{19.4}	\\
PDC-Net~\cite{pdcnet} retrieval &   20.1	&   \textbf{19.4} \\
\textbf{PDC-Net+ retrieval} & 17.1 & \textbf{19.4} \\

\bottomrule
\end{tabular}%
}\vspace{-3mm}
\label{tab:aachen-retrieval}
\end{table}

\begin{figure}[b]
\centering%
\vspace{-3mm}
\includegraphics[width=0.49\textwidth]{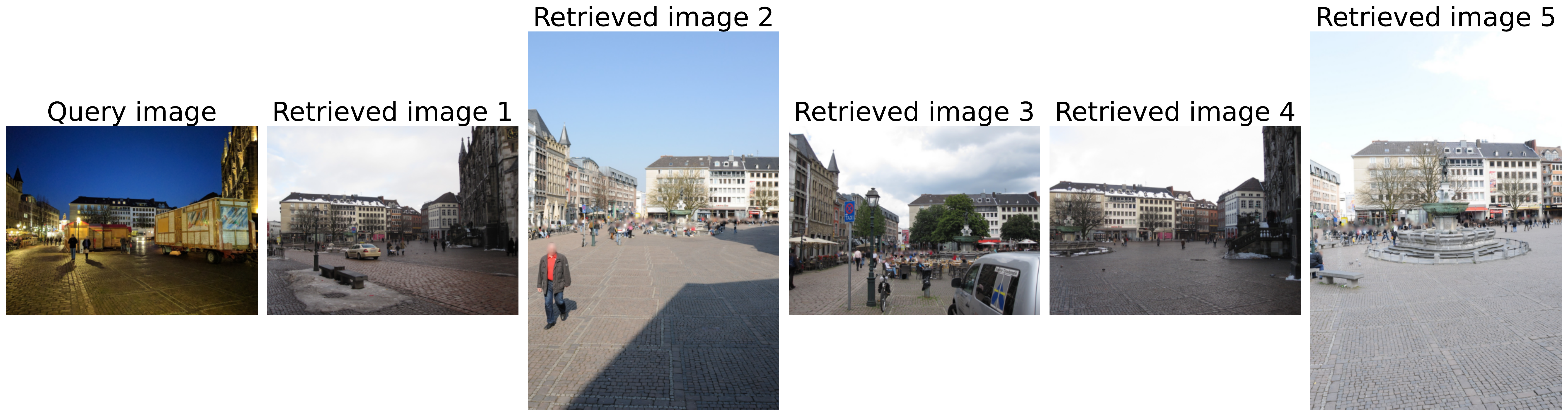}
\vspace{-8mm}
\caption{Example of retrieved images for a night query of the Aachen benchmark~\cite{SattlerWLK12, SattlerMTTHSSOP18}, using PDC-Net+.}
\label{fig:retrieval}
\end{figure}

\parsection{Retrieval} Another application enabled by our uncertainty prediction is image retrieval. For a specific query image, we compute the flow fields and confidence maps $P_R$ \eqref{eq:pr} relating the query to  all images of the database. We then rank the database images based on the number of confident matches for which $P_R > \gamma$, where $\gamma$ is a scalar threshold in $[0, 1)$. 
For computational efficiency, we compute the flow and confidence map by resizing the images to a relatively low resolution, \eg so that the smallest dimension is 400. We use a single forward pass of the network for prediction. 

We evaluate on the Aachen Benchmark~\cite{SattlerMTTHSSOP18, SattlerWLK12} with the 824 daytime and 98 nighttime query images. As previously, the evaluation is done through the server of~\cite{SattlerMTTHSSOP18}, which only reports metrics for pre-defined error thresholds. 
However, for many query images, the pose error corresponding to the nearest retrieved database image exceeds the pre-defined largest threshold of 5m, 10\textdegree. 
Nevertheless, we present pose accuracy results for this largest threshold in Tab.~\ref{tab:aachen-retrieval}.

We compare our approach based on PDC-Net+ to standard retrieval methods: NetVLAD~\cite{NetVLAD}, Landmark Localization Network (LLN)~\cite{LLN} and DenseVLAD~\cite{DenseVLAD}. 
Both PDC-Net+ and PDC-Net achieve competitive results compared to approaches that are specifically designed for the image retrieval task. Our single network can thus be applied for both retrieval and matching steps in the localization pipeline. 
In Fig.~\ref{fig:retrieval}, we present an example of the 5 closest images retrieved by PDC-Net+ for a night query.

\begin{figure}[t]
\centering%
\includegraphics*[width=\columnwidth]{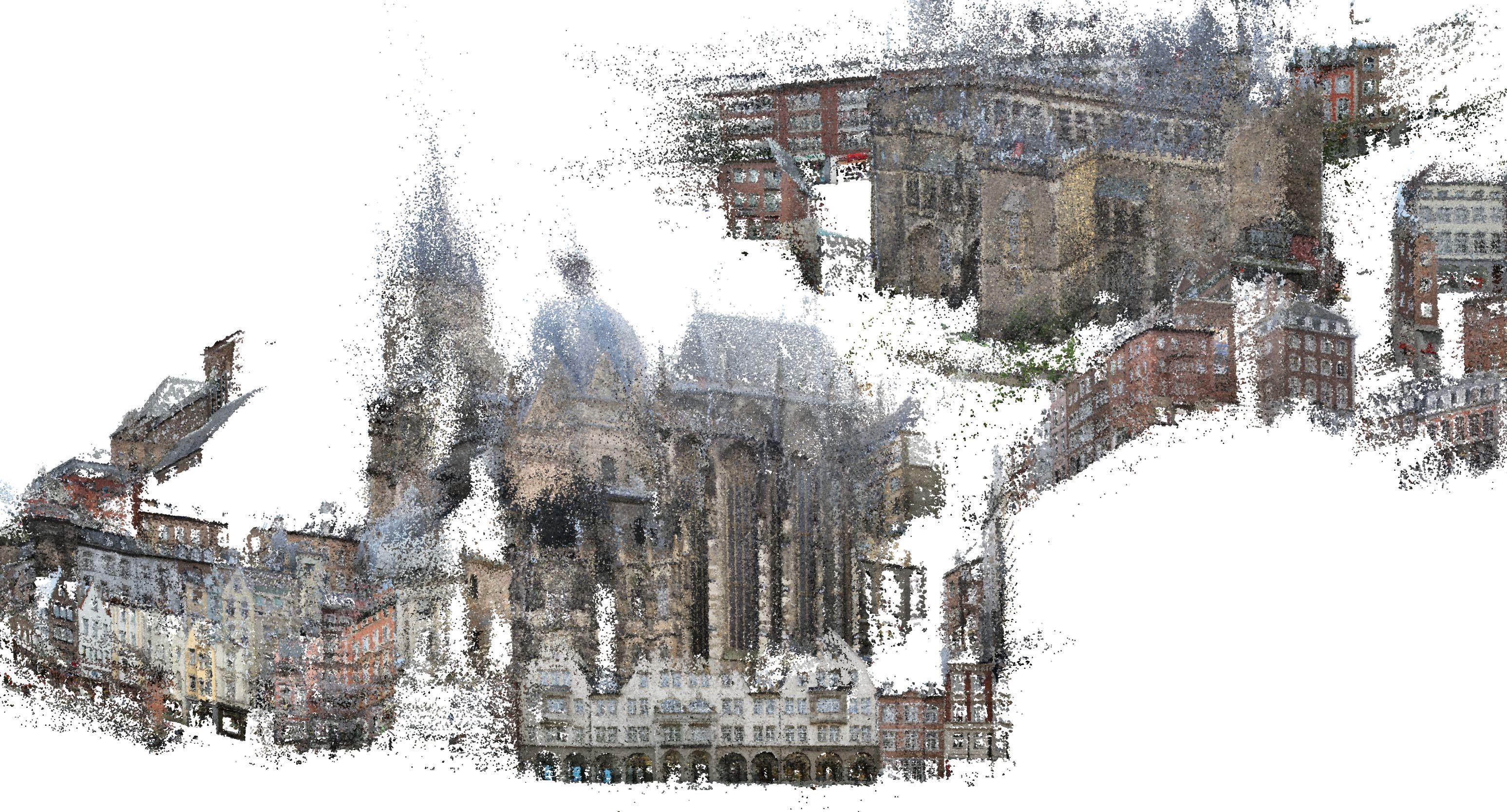}
\vspace{-5mm}
\caption{3D reconstruction of Aachen~\cite{SattlerWLK12} using the dense correspondences and uncertainties predicted by PDC-Net+.}\vspace{-3mm} 
\label{fig:aachen}
\end{figure}

\parsection{3D reconstruction} We also qualitatively show the usability of our approach for dense 3D reconstruction. In Fig.~\ref{fig:aachen}, we visualize the 3D point cloud produced by PDC-Net+ and COLMAP~\cite{SchonbergerF16}, corresponding to the fixed pipeline of the Visual Localization benchmark~\cite{SattlerMTTHSSOP18, SattlerWLK12}.

\subsection{Additional ablation study}
\label{subsec:ablation-study}

Here, we perform a detailed analysis of our approach. 

\subsubsection{Ablation of PDC-Net}

We first ablate components common to PDC-Net and PDC-Net+ in Tab.~\ref{tab:ablation}. As baseline, we use a simplified version of GLU-Net, called BaseNet~\cite{GLUNet}. It is a three-level pyramidal network predicting the flow between an image pair. 
Our probabilistic approach integrated in this smaller architecture is termed PDC-Net-s. 
All methods are trained using only the first-stage training of the initial PDC-Net, described in Sec.~\ref{subsec:arch}.

\parsection{Probabilistic model (Tab.~\ref{tab:ablation},~top section)} 
We first compare BaseNet to our approach PDC-Net-s, which models the flow distribution with a constrained mixture of Laplacians (Sec.~\ref{sec:constained-mixture}). On both KITTI-2015 and MegaDepth, our approach brings a significant improvement in terms of flow accuracy. For pose estimation on YFCC100M, using all correspondences of BaseNet performs very poorly. This demonstrates the crucial importance of robust uncertainty estimation. 
While an unconstrained mixture of Laplacians already drastically improves upon the single Laplace component, the permutation invariance of the former leads to poor uncertainty estimates, shown by the high AUSE. Constraining the mixture instead results in better metrics for both the flow and the uncertainty. 

\parsection{Uncertainty architecture (Tab.~\ref{tab:ablation}, 2$^{nd}$ section)} Although the compared uncertainty decoder architectures achieve similar quality in flow prediction (AEPE and Fl), they provide notable differences in uncertainty estimation (AUSE). 
Only using the correlation uncertainty module leads to the best results on YFCC100M, since the module enables to efficiently discard unreliable matching regions, in particular compared to the common decoder approach.
However, this module alone does not take into account motion boundaries. This leads to poor AUSE on KITTI-2015, which contains independently moving objects. Our final architecture (Fig.~\ref{fig:arch}), additionally integrating the mean flow into the uncertainty estimation, offers the best compromise.

\parsection{Perturbation data (Tab.~\ref{tab:ablation},~3$^{rd}$ section)} 
While introducing the perturbations does not help the flow prediction for BaseNet, it provides  significant improvements in uncertainty \emph{and} in flow performance for our PDC-Net-s. This emphasizes that the improvement of the uncertainty estimates originating from introducing the perturbations also leads to improved and more generalizable flow predictions. 

\begin{table}[t]
\centering
\caption{Ablation study. In the top part, different probabilistic models are compared (Sec.~\ref{subsec:proba-model}-~\ref{sec:constained-mixture}). In the second part, a constrained Mixture is used, and different architectures for uncertainty estimation are compared (Sec.~\ref{sec:uncertainty-arch}). In the third part, we analyze the impact of our training data with perturbations (Sec.~\ref{subsec:perturbed-data}). For all other sections, we model the flow as a constrained mixture of Laplace distributions, and we use our uncertainty prediction architecture. In the fourth part, we show the impact of propagating the uncertainty estimates in a multi-scale architecture (Sec.~\ref{sec:uncertainty-arch}). 
In the bottom part, we compare different parametrization of the constrained mixture (Sec.~\ref{sec:constained-mixture}).}
\vspace{-2mm}
\resizebox{0.48\textwidth}{!}{%
\begin{tabular}{l@{~}c@{~~}c@{~~}c@{~~}|@{~~}c@{~~}c@{~~}c@{~~}|@{~~}c@{~~}c} 
\toprule
& \multicolumn{3}{c}{\textbf{KITTI-2015}} & \multicolumn{3}{c}{\textbf{MegaDepth}} & \multicolumn{2}{c}{\textbf{YFCC100M}}\\
  & AEPE  & Fl (\%)  & AUSE & PCK-1 (\%)  & PCK-5 (\%) & AUSE & mAP @5\textdegree & mAP @10\textdegree  \\ \midrule
BaseNet (L1-loss) & 7.51 & 37.19 & - & 20.00 & 60.00 & - &  15.58 & 24.00 \\
Single Laplace & 6.86 & 34.27 & 0.220 & 27.45 & 62.24 & 0.210 & 26.95 & 37.10 \\
Unconstrained Mixture & 6.60 & 32.54 & 0.670 & 30.18 & 66.24 & 0.433 & 31.18 & 42.55 \\
Constrained Mixture (PDC-Net-s) & \textbf{6.66} & \textbf{32.32} & \textbf{0.205} & \textbf{32.51} & \textbf{66.50} & \textbf{0.210} & \textbf{33.77} & \textbf{45.17}  \\
\midrule

Commun Dec.  & 6.41 & 32.03 & \textbf{0.171} & 31.93 & \textbf{67.34} & 0.213 & 31.13  & 42.21 \\
Corr unc. module & \textbf{6.32} & \textbf{31.12} & 0.418 & 31.97 & 66.80 & 0.278 & \textbf{33.95} & \textbf{45.44} \\
Unc. Dec. (Fig~\ref{fig:arch}) (PDC-Net-s) & 6.66 & 32.32 & 0.205 & \textbf{32.51} & 66.50 & \textbf{0.210} & 33.77 & 45.17  \\ \midrule 

BaseNet \textbf{w/o} Perturbations & 7.21 & 37.35 & - &  20.74 & 59.35 &  - & 15.15 & 23.88 \\
BaseNet \textbf{w} Perturbations  & 7.51 & 37.19 & - & 20.00 & 60.00 & - &  15.58 & 24.00 \\
PDC-Net-s \textbf{w/o} Perturbations  & 7.15 & 35.28 & 0.256 &  31.53 & 65.03 & 0.219 & 32.50 & 43.17 \\
PDC-Net-s \textbf{w} Perturbations & \textbf{6.66} & \textbf{32.32} & \textbf{0.205} & \textbf{32.51} & \textbf{66.50} & \textbf{0.210} & \textbf{33.77} & \textbf{45.17} \\
\midrule
Uncert. not propagated  & 6.76 & \textbf{31.84} & 0.212 & 29.9 & 65.13 & 0.213 & 31.50 & 42.19  \\
PDC-Net-s  &  \textbf{6.66} &  32.32   &  \textbf{0.205}   &  \textbf{32.51}   &  \textbf{66.50}   &  \textbf{0.197}   &  \textbf{33.77}   &  \textbf{45.17}    \\
\midrule
$0 \leq \sigma^2_1 \leq 1$, $2 \leq \sigma^2_2 \leq \infty$ & 6.69 & 32.58 & \textbf{0.181} & 32.47 &  65.45 & 0.205 & 30.50 & 40.75 \\

$\sigma^2_1=1$, $2 \leq \sigma^2_2 \leq \infty$ 
& 6.66 &  32.32   &  0.205   &  \textbf{32.51}   &  66.50   &  \textbf{0.197}   &  \textbf{33.77}   &  \textbf{45.17} \\

$\sigma^2_1=1$, $2 \leq \sigma^2_2 \leq \beta_2^+=HW$ &  \textbf{6.61} & \textbf{31.67} & 0.208 & 31.83 & \textbf{66.52} & 0.204 & 33.05 & 44.48 \\
\bottomrule
\end{tabular}%
}\label{tab:ablation}
\vspace{-3mm}
\end{table}

\parsection{Propagation of uncertainty components (Tab.~\ref{tab:ablation},~4$^{th}$ section)} 
For all presented datasets and metrics, propagating the uncertainty predictions (Sec.~\ref{sec:uncertainty-arch}) boosts the performance of the final network. Only the Fl metric on KITTI-2015 is slightly worse. 

\parsection{Constrained mixture parametrization (Tab.~\ref{tab:ablation}, bottom section)} 
We compare fixing the first component $\sigma^2_1=\beta_1^- =\beta_1^+ = 1.0$ with a version where the first component is instead bounded as $\beta_1^- = 0.0 \leq \sigma^2_1 \leq \beta_1^+=1.0$. Both networks obtain similar flow performance on KITTI-2015 and MegaDepth. Only on optical flow dataset KITTI-2015, the second alternative of $\beta_1^- = 0.0 \leq \sigma^2_1 \leq \beta_1^+=1.0$ obtains a better AUSE. This is because KITTI-2015 shows ground-truth displacements with a much smaller magnitude than in the geometric matching datasets. When estimating the flow on KITTI, it thus results in a larger proportion of very small flow errors (lower EPE and higher PCK than on geometric matching data). As a result, on this dataset, being able to model a very small error (with $\sigma^2_1 \leq 1$) is beneficial.  However, fixing $\sigma^2_1=1.0$  instead produces better AUSE on MegaDepth and it gives significantly better results for pose estimation on YFCC100M. 

We then compare leaving the second component's higher bound unconstrained as $2 \leq \sigma^2_2 \leq \infty$ (PDC-Net-s) with a bounded version $2 \leq \sigma^2_2 \leq \beta_2^+ = HW$.  All results are very similar, the upper bounded network obtains slightly better flow results but slightly worse uncertainty performance (AUSE). However, we found that constraining $\beta_2^+$ leads to a more stable training in practice. We therefore adopt this setting, using the constraints $\sigma^2_1 = \beta_1^- = \beta_1^+=1$, and $2.0 = \beta_2^- \leq \sigma^2_2 \leq \beta_2^+ = HW$ for our final network (see Sec.~\ref{subsec:arch}).

\subsubsection{Ablation of PDC-Net+ versus PDC-Net}

We analyze aspects specific to our enhanced PDC-Net+ approach in Tab.~\ref{tab:ablation-pdcnetv2}. For all experiments, we use PDC-Net+ with only the first stage training, described in Sec.~\ref{subsec:arch}.

\parsection{Number of moving objects (Tab.~\ref{tab:ablation-pdcnetv2}, top)} Introducing multiple independently moving objects during training leads to better flow results on all datasets. Optical flow data particularly benefits from this training, with a substantial improvement of 3\% in Fl metric. It however results in a slight decrease in AUSE, due to the difficulty to model true errors in the presence of multiple objects. 

\parsection{Training mask (Tab.~\ref{tab:ablation-pdcnetv2}, bottom)} Training with the injective mask (Sec.~\ref{sec:injective-mask}) leads to a substantial improvement on KITTI-2015 compared to no mask. Indeed, the injective training mask ensures that the ground-truth flow is injective, while still supervising the flow in occluded areas. It thus allows the network to learn two crucial properties, namely to match independently moving objects and to extrapolate the flow in occluded regions. 
On the other hand, using the occlusion mask leads to significantly worse results. This is because the network does not learn to extrapolate the predicted flow in occluded areas, which is essential for the optical flow task.
As a result, our injective mask is the best alternative for training a single network that obtains strong results on all datasets.

\begin{table}[t]
\centering
\caption{Ablation study of PDC-Net+. In the top part, we compare training with a single or multiple independently moving objects (Sec.~\ref{sec:training-strategy}).  In the bottom part, different masking strategies during training are compared (Sec.~\ref{sec:injective-mask}). All metrics are computed with a single forward-pass (D).}
\vspace{-2mm}
\resizebox{0.48\textwidth}{!}{%
\begin{tabular}{lccc|ccc|cc} 
\toprule
& \multicolumn{3}{c}{\textbf{KITTI-2015}} & \multicolumn{3}{c}{\textbf{MegaDepth}} & \multicolumn{2}{c}{\textbf{YFCC100M}}\\
  & AEPE  & Fl (\%)  & AUSE & PCK-1 (\%)  & PCK-5 (\%) & AUSE & mAP @5\textdegree & mAP @10\textdegree  \\ \midrule
1 object & 6.70 & 19.61 & \textbf{0.123} & 52.62 & 73.78 & \textbf{0.215} & 44.38 & 54.04 \\
Multiple objects & \textbf{5.55} & \textbf{16.75} & 0.136 & \textbf{54.89} & \textbf{75.92} & 0.238 & \textbf{44.45} & \textbf{54.70}  \\
\midrule
No mask &  6.01 & 18.07 & 0.139 & 53.66 & 75.05 & 0.247 & 39.92 & 50.94 \\
Injective mask & \textbf{5.55} & \textbf{16.75} & 0.136 & 54.89 & 75.92 & \textbf{0.238} & \textbf{44.45} & 54.70 \\
Occlusion mask & 9.64 & 19.64 & \textbf{0.103} & \textbf{56.41} & \textbf{77.37} & 0.247 & 44.40 & \textbf{55.68} \\
\bottomrule
\end{tabular}%
}
\vspace{-3mm}
\label{tab:ablation-pdcnetv2}
\end{table}
\section{Conclusion}
\label{sec:conclusion}

In this paper, we propose a probabilistic deep network for estimating the dense image-to-image correspondences and an associated confidence estimate. Our network is trained to predict the parameters of the conditional probability density of the flow. To this end, we introduce a constrained mixture of Laplace distributions. To achieve robust and generalizable uncertainty prediction, we propose an architecture and improved self-supervised data generation pipeline.
We further introduced an enhanced self-supervised strategy that better captures independently moving objects and occlusions. We develop an injective training mask, which benefits learning in the presence of occlusions. 

We comprehensively evaluate our final PDC-Net+ approach for multiple different applications and settings. Our approach sets a new state-of-the-art on geometric matching and optical flow datasets, while achieving more robust confidence estimates. Moreover, our uncertainty estimation opens the door to applications traditionally dominated by sparse methods, such as pose estimation and image-based localization. 

Future work includes developing even more realistic self-supervised training pipelines, capable of better capturing 3D scenes. An alternative direction is to explore unsupervised learning strategies or the incorporation of 3D constraints as supervision. Another direction involves applying our generic PDC-Net+ formulation to improved backbone and flow estimation architectures.

\ifCLASSOPTIONcompsoc
  \section*{Acknowledgments}
\else
  \section*{Acknowledgment}
\fi

This work was supported by the ETH Z\"urich Fund (OK), a Huawei Gift, Huawei Technologies Oy (Finland), Amazon AWS, and an Nvidia GPU grant.

\ifCLASSOPTIONcaptionsoff
  \newpage
\fi



%
\bibliographystyle{IEEEtran}
\bibliography{biblio}

%

\newpage
\begin{IEEEbiography}[{\includegraphics[width=1in,height=1.25in,clip,keepaspectratio]{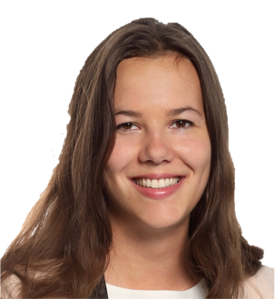}}]{Prune Truong} is a PhD student at the Computer Vision Lab of ETH Zürich, under the supervision of Prof. Luc Van Gool. She obtained her MSc in Mechanical Engineering at ETH Zürich in 2019, with honors. 
\end{IEEEbiography}

\vspace{-1.3cm}
\begin{IEEEbiography}[{\includegraphics[width=1in,height=1.25in,clip,keepaspectratio]{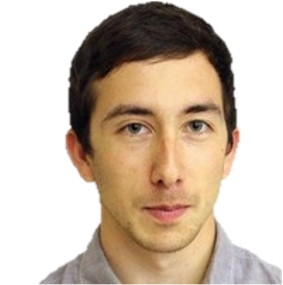}}]{Martin Danelljan} 
is a group leader and lecturer at ETH Z\"urich, Switzerland. He received his Ph.D.\ degree from Link\"oping University, Sweden in 2018. His Ph.D.\ thesis was awarded the biennial Best Nordic Thesis Prize at SCIA 2019. His main research interests are meta and online learning, deep probabilistic models, and conditional generative models. His research includes applications to visual tracking, video object segmentation, dense correspondence estimation, and super-resolution. His research in the field of visual tracking, in particular, has attracted much attention, achieving first rank in the 2014, 2016, and 2017 editions of the Visual Object Tracking (VOT) Challenge and the OpenCV State-of-the-Art Vision Challenge. He received the best paper award at ICPR 2016, the best student paper at BMVC 2019, and an outstanding reviewer award at ECCV 2020. He is also a co-organizer of the VOT, NTIRE, and AIM workshops. He serves as an Area Chair for CVPR 2022 and Senior PC for AAAI 2022.
\end{IEEEbiography}


\vspace{-1.3cm}
\begin{IEEEbiography}[{\includegraphics[width=1in,height=1.25in,clip,keepaspectratio]{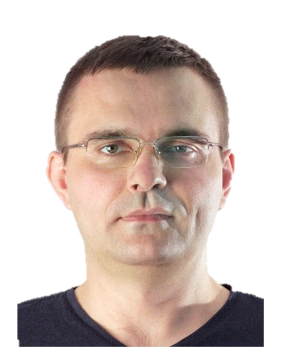}}]
{Radu Timofte} 
is lecturer and research group leader in the Computer Vision Laboratory, at ETH Zurich, Switzerland. Radu Timofte is also Chair of Computer Science at University of W\"urzburg, Germany, and a 2022 winner of the Humboldt Professorship for Artificial Intelligence Award. He obtained a PhD degree in Electrical Engineering at the KU Leuven, Belgium in 2013, the MSc at the Univ. of Eastern Finland in 2007, and the Dipl. Eng. at the Technical Univ. of Iasi, Romania in 2006. He is associate editor for top journals. He serves(d) as area chair for ACCV 2018, ICCV 2019, ECCV 2020, ACCV 2020, CVPR 2021, IJCAI 2021, CVPR 2022 and as Senior PC member for IJCAI 2019 and 2020. He received a NIPS 2017 best reviewer award. His work received the best student paper award at BMVC 2019, a best scientific paper award at ICPR 2012, the best paper award at CVVT workshop (ECCV 2012), the best paper award at ChaLearn LAP workshop (ICCV 2015), the best scientific poster award at EOS 2017, the honorable mention award at FG 2017, and his team won a number of challenges. 
He is co-founder of Merantix, co-organizer of NTIRE, CLIC, AIM, MAI and PIRM events, and member of IEEE, CVF, and ELLIS. 
\end{IEEEbiography}

\vspace{-1.3cm}
\begin{IEEEbiography}[{\includegraphics[width=1in,height=1.25in,clip,keepaspectratio]{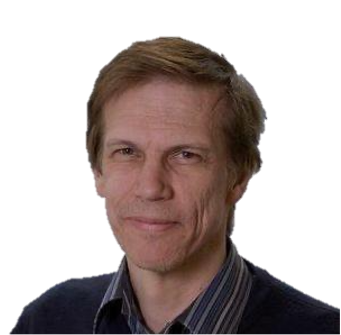}}]
{Luc Van Gool} is both a full professor at KU Leuven (Belgium) and at ETH Zürich (Switzerland). His main area of expertise is computer vision.  
He received the Koenderink Prize at European Conference on Computer Vision in 2016, the David Marr prize (best paper award) at the International Conference on Computer Vision in 1998 and the U.V. Helava Award, one of the most prestigious ISPRS awards, in 2012. He was also awarded an ERC Advanced Grant in 2011 for his project VarCity (Variation \& the City), was nominated `Distinguished Researcher’ by the IEEE Society of Computer Vision in 2017, and received the 5-yearly excellence prize by the Flemish Fund for Scientific Research in 2016. He received several other best paper awards as well. Luc Van Gool is co-founder of several spin-offs. He has been involved in the organization of several, major conferences and as an associate editor for multiple, first-tier scientific journals. 
\end{IEEEbiography}




\clearpage
\newpage
\appendices

In this appendix, we first give a detailed derivation of our probabilistic model as a constrained mixture of Laplace distributions in Sec.~\ref{sec-sup:proba}. In Sec.~\ref{sec-sup:training}, we then derive our probabilistic training loss and explain our training procedure in more depth.  We subsequently follow by providing additional information about the architecture of our proposed networks as well as implementation details in Sec.~\ref{sec-sub:arch-details}. 
In Sec.~\ref{sec-sup:details-evaluation}, we extensively explain the evaluation datasets and set-up.  Then, we present more detailed quantitative and qualitative results in Sec.~\ref{sec-sup:results}.
Finally, we perform additional ablative experiments in Sec.~\ref{sec-sup:ablation}.

\section{Detailed derivation of probabilistic model}
\label{sec-sup:proba}

Here we provide the details of the derivation of our uncertainty estimate. 

\parsection{Probabilistic formulation} We model the flow estimation as a probabilistic regression with a constrained mixture density of Laplacian distributions (Sec.~\ref{sec:constained-mixture} of the main paper). Our mixture model, corresponding to equation \eqref{eq:mixture} of the main paper, is expressed as,
\begin{equation}
\label{eq:mixture-sup}
p\left(y | \varphi \right) = \sum_{m=1}^{M} \alpha_{m} \mathcal{L}(y| \mu, \sigma^2_m)
\end{equation}
where, for each component $m$, the bi-variate Laplace distribution $\mathcal{L}(y| \mu, \sigma^2_m)$ is computed as the product of two independent uni-variate Laplace distributions, such as, 
\begin{subequations}
\label{eq:simple-laplace}
\begin{align}
\mathcal{L}(y| \mu, \sigma^2_m) &= \mathcal{L}(u,v| \mu_u, \mu_v, \sigma^2_u, \sigma^2_v) \\
&= \mathcal{L}(u| \mu_u, \sigma^2_u). \mathcal{L}(v| \mu_v, \sigma^2_v) \\
&= \frac{1}{\sqrt{2 \sigma_u^2}} e^{-\sqrt{\frac{2}{\sigma_u^2}}|u-\mu_u|} .  \nonumber \\
& \hspace{5mm} \frac{1}{\sqrt{2 \sigma_v^2}}e^{-\sqrt{\frac{2}{\sigma_v^2}}|v-\mu_v|}
\end{align}
\end{subequations}
where $\mu = [\mu_u, \mu_v]^T \in \mathbb{R}^2$ and  $\sigma^2_m = [\sigma^2_u, \sigma^2_v]^T \in \mathbb{R}^2$ are respectively the mean and the variance parameters of the distribution $\mathcal{L}(y| \mu, \sigma^2_m)$. In this work, we additionally define equal variances in both flow directions, such that $\sigma^2_m = \sigma^2_u = \sigma^2_v \in \mathbb{R}$. As a result, equation \eqref{eq:simple-laplace} simplifies, and when inserting into \eqref{eq:mixture-sup}, we obtain,
\begin{equation}
\label{eq:mixture-sup-final}
p\left(y | \varphi \right) =\sum_{m=1}^{M} \alpha_{m} \frac{1}{2 \sigma_m^2} e^{-\sqrt{\frac{2}{\sigma_m^2}}|y-\mu|_1}.
\end{equation}

\begin{figure}[t]
\centering
(A) MegaDepth \\
\includegraphics[width=0.47\textwidth]{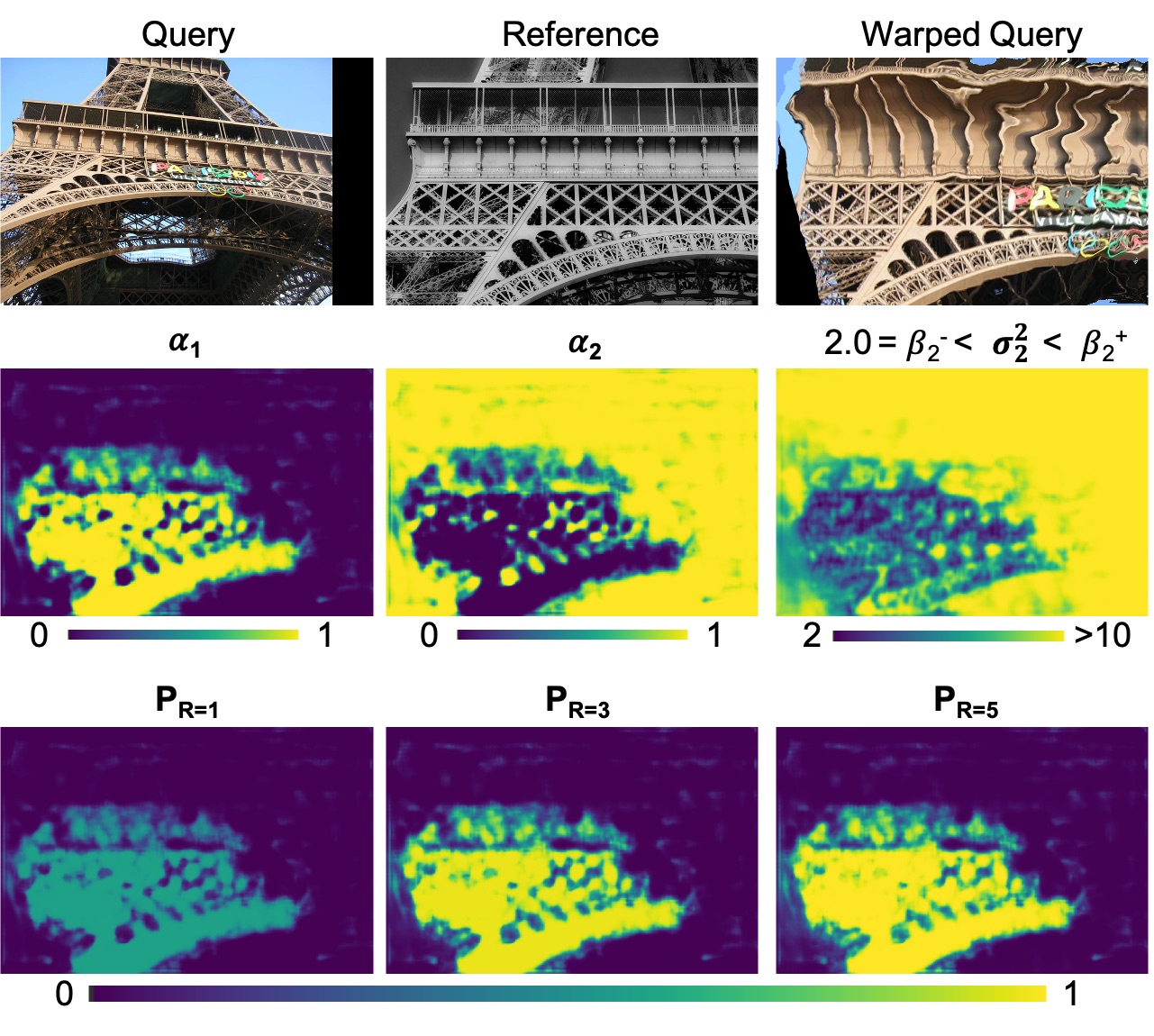} \\
(B) KITTI-2015 \\
\includegraphics[width=0.47\textwidth]{ 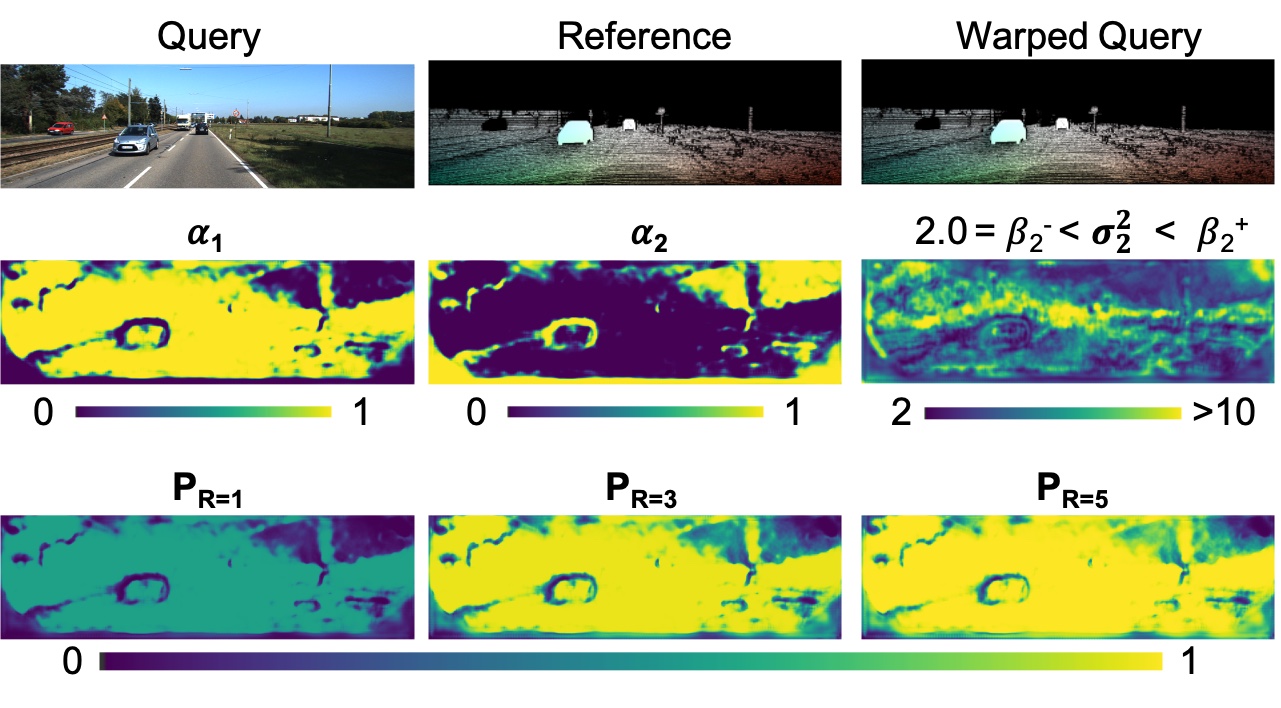} \\
\caption{Visualization of the mixture parameters $(\alpha_m )_{m=1}^M$ and $\sigma^2_2$ predicted by our network PDC-Net+, on multiple image pairs. PDC-Net+ has $M=2$ Laplace components and here, we do not represent the scale parameter $\sigma^2_1$, since it is fixed as $\sigma^2_1 = 1.0$. We also show the resulting confidence maps $P_R$ for multiple R. }\vspace{-4mm}
\label{fig:conf-map}
\end{figure}

\parsection{Confidence estimation} Our network $\Phi$ thus outputs, for each pixel location, the parameters of the predictive distribution, \ie the mean flow $\mu$ along with the variance $\sigma^2_m$ and weight $\alpha_m$ of each component, as $\big( \mu, (\alpha_m )_{m=1}^M, ( \sigma^2_m  )_{m=1}^M  \big) = \varphi(X; \theta)$. However, we aim at obtaining a \emph{single} confidence value to represent the relibiability of the estimated flow vector $\mu$.
As a final confidence measure,  we thus compute the probability $P_R$ of the true flow being within a radius $R$ of the estimated mean flow vector $\mu$. This is expressed as,
\begin{subequations}
\begin{align}
    P_R & = P(\|y - \mu\|_\infty < R) \\
    & = \int_{\{y\in \reals^2: \|y - \mu\|_\infty < R\}}\! p(y|\varphi) dy \\
    &= \sum_{m} \alpha_m \int_{\mu_u-R}^{\mu_u+R}
    \frac{1}{\sqrt{2} \sigma_m} e^{-\sqrt{2}\frac{|u-\mu_u|}{\sigma_m}} du   \nonumber \\
      &    \hspace{15mm}   \int_{\mu_v-R}^{\mu_v+R}
    \frac{1}{\sqrt{2} \sigma_m} e^{-\sqrt{2}\frac{|v-\mu_v|}{\sigma_m}}dv \\
    &= \sum_{m} \alpha_m \left[1-\exp (-\sqrt{2}\frac{R}{\sigma_m}) \right]^2
\end{align}
\end{subequations}
This confidence measure is used to identify the accurate matches by thesholding $P_R$. In Fig~\ref{fig:conf-map}, we visualize the estimated mixture parameters $(\alpha_m )_{m=1}^M, ( \sigma^2_m  )_{m=1}^M$, and the resulting confidence map $P_R$ for multiple image pair examples.

\section{Training details}
\label{sec-sup:training}

In this section, we derive the numerically stable Negative Log-Likelihood loss, used for training our network PDC-Net+. We also describe in details the employed training datasets. 

\subsection{Training loss}
\label{sec-sup:training-loss}

Similar to conventional approaches, probabilistic methods are generally trained using a set of \emph{iid}.\ image pairs $\mathcal{D} = \left \{X^{(n)}, Y^{(n)} \right \}_{n=1}^N$.
The negative log-likelihood provides a general framework for fitting a distribution to the training dataset as,
\begin{subequations}
\label{eq:nll-sup}
\begin{align}
L(\theta; \mathcal{D}) &=  - \frac{1}{N} \sum_{n=1}^{N} \log p\left(Y^{(n)} | \Phi(X^{(n)}; \theta) \right)  \\
&= - \frac{1}{N} \sum_{n=1}^{N} \sum_{ij} \log p\big(y^{(n)}_{ij} | \varphi_{ij}(X^{(n)};\theta)\big) 
\end{align}
\end{subequations}
Note that to avoid clutter, we do not include here our training injective mask (Sec. \ref{sec:injective-mask} of the main paper) in the loss formulation. 
Inserting \eqref{eq:mixture-sup-final} into \eqref{eq:nll-sup}, we obtain for the last term the following expression,
\begin{subequations}
\label{eq:p-sup}
\begin{align}
L_{ij} &= -\log p\big(y^{(n)}_{ij} | \varphi_{ij}(X^{(n)};\theta)\big) \\
&= -\log \left ( \sum_{m=1}^{M} \alpha_{m} \frac{1}{2 \sigma_m^2} e^{  -\sqrt{\frac{2}{\sigma_m^2}}|y-\mu|_1  }   \right) \\
&=  -\log \left(   \sum_{m=1}^{M} \frac{e^{\tilde{\alpha}_m}}{\sum_{m=1}^{M}e^{\tilde{\alpha}_m}} \frac{1}{2 \sigma_m^2} e^{   -\sqrt{\frac{2}{\sigma_m^2}}|y-\mu|_1  } \right) \\
&=\log  \left( \sum_{m=1}^{M}e^{\tilde{\alpha}_m} \right) \nonumber  \\ 
& \;\;\;\;\;  -\log \left(  \sum_{m=1}^{M} 
e^{\tilde{\alpha}_m}\frac{1}{2 \sigma_m^2} e^{  -\sqrt{\frac{2}{\sigma_m^2}}|y-\mu|_1  } \right) \\
&= \log   \left( \sum_{m=1}^{M}e^{\tilde{\alpha}_m}  \right) \nonumber \\ & 
\;\;\;\;\; -\log \left(\sum_{m=1}^{M} e^{\tilde{\alpha}_m - \log(2) - s_m -\sqrt{2}e^{-\frac{1}{2}s_m}.|y-\mu|_1}  \right)   
\end{align}
\end{subequations}
where $s_m = log(\sigma^2_m)$. Indeed, in practise, to avoid division by zero, we use $s_{m} = \log(\sigma^2_m)$ for all components $m \in \left \{0, .., M \right \}$ of the mixture density model. For the implementation of the loss, we use a numerically stable logsumexp function.

With a simple regression loss such as the L1 loss, the large errors represented by the heavy tail of the distribution in Fig.~\ref{fig:distribution} of the main paper have a disproportionately large impact on the loss, preventing the network from focusing on the more accurate predictions. On the contrary, the loss \eqref{eq:p-sup} enables to down-weight the contribution of these examples by predicting a high variance parameter for them. Modelling the flow estimation as a conditional predictive distribution thus improves the accuracy of the estimated flow itself. 

\subsection{Injective mask}

\begin{figure}[t]
\centering
\includegraphics[width=0.48\textwidth]{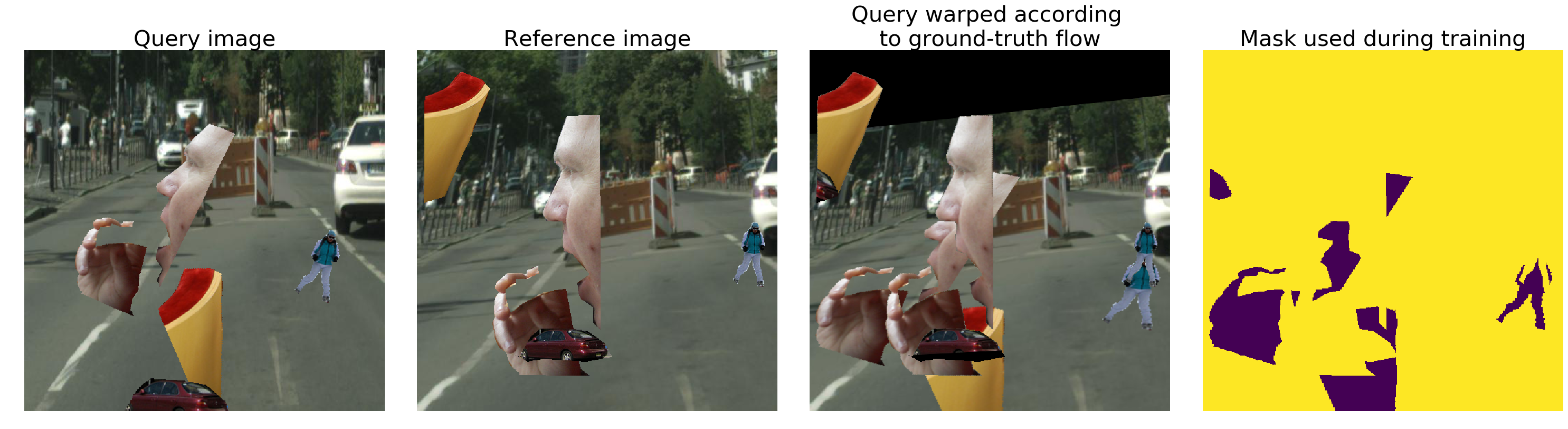} \\
\includegraphics[width=0.48\textwidth, trim=0 0 0 0]{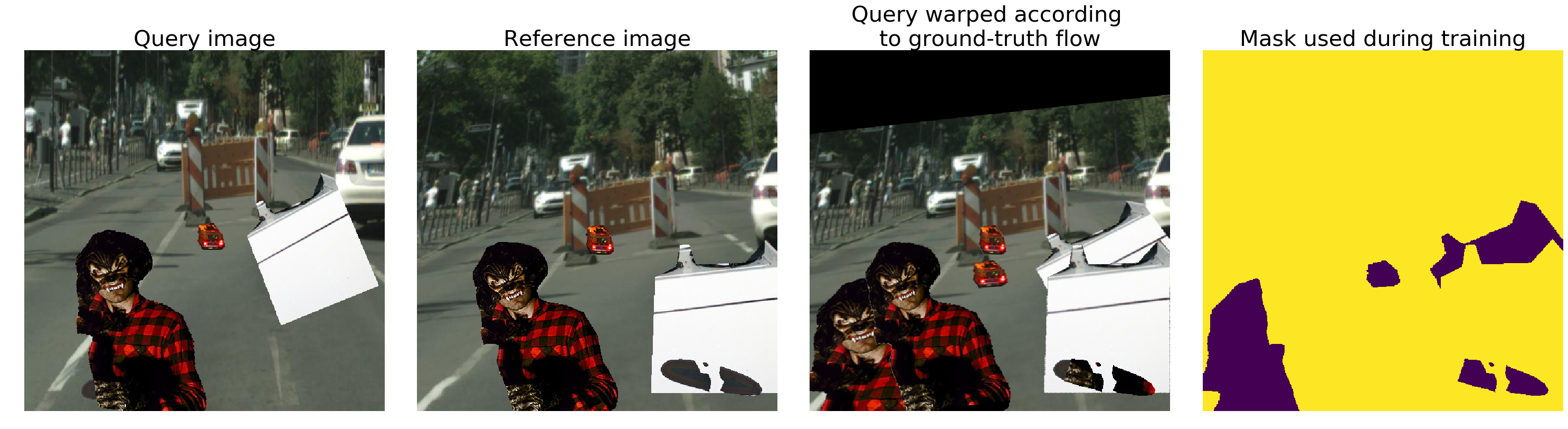} \\
\includegraphics[width=0.48\textwidth]{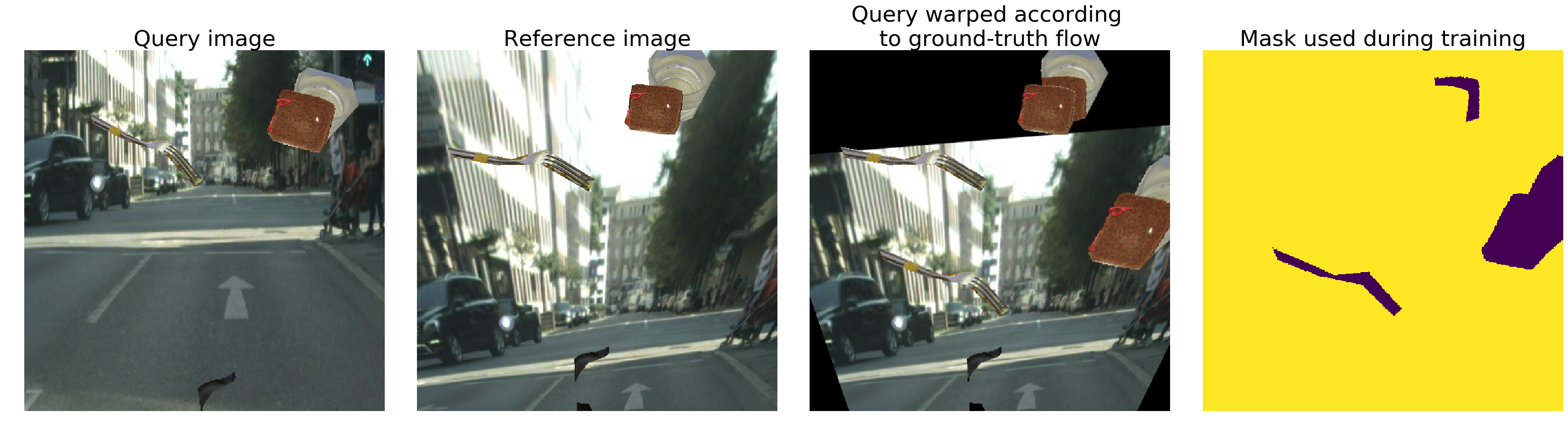} \\
\includegraphics[width=0.48\textwidth]{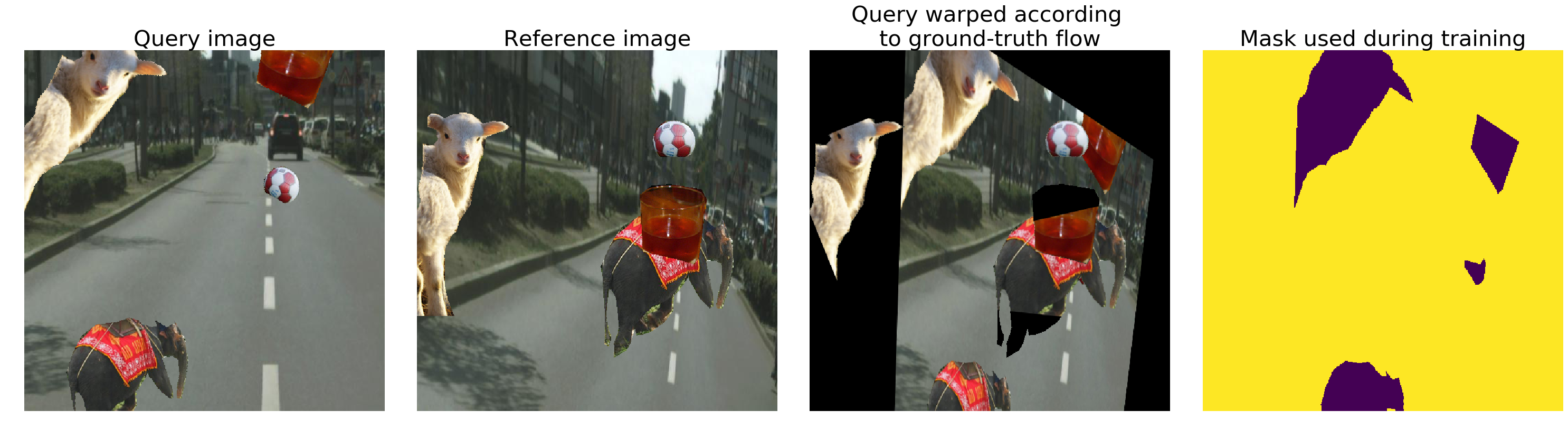} \\
\includegraphics[width=0.48\textwidth]{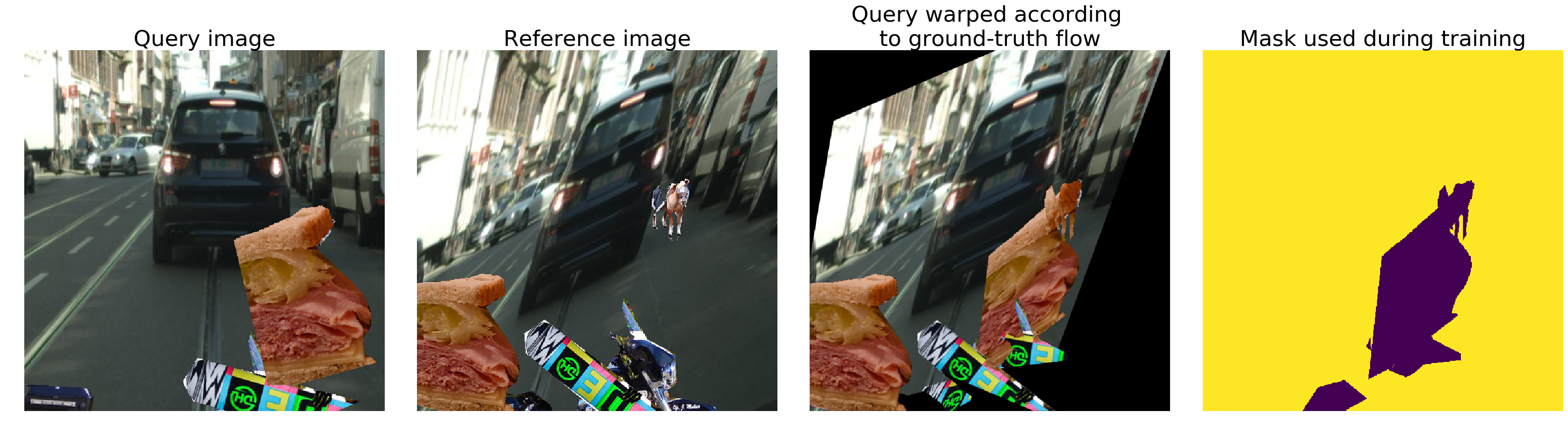} \\
\includegraphics[width=0.48\textwidth]{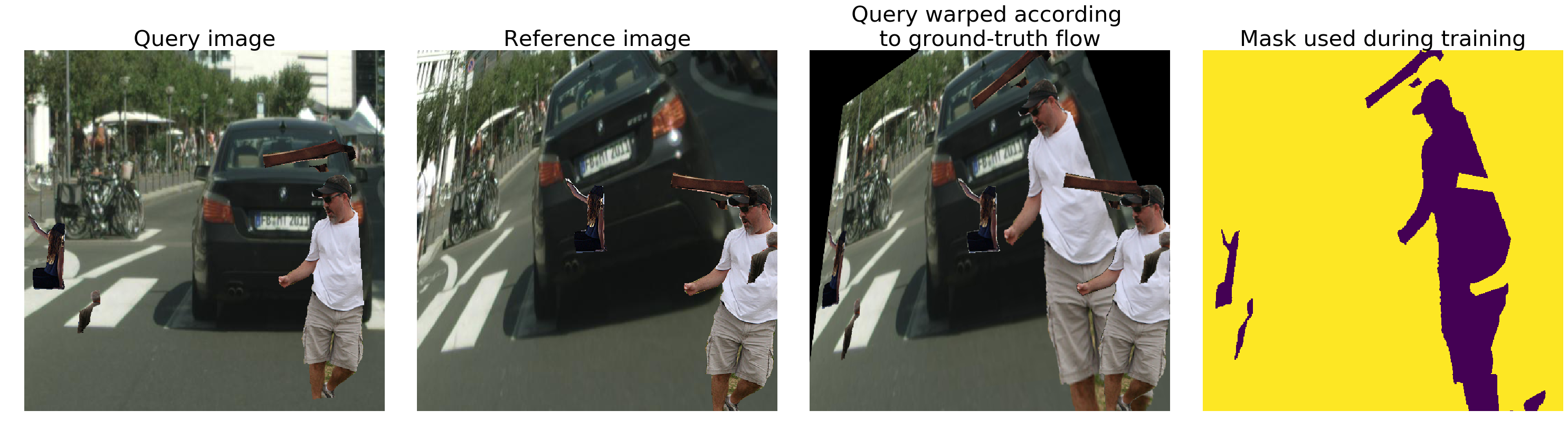} \\
\vspace{-2mm}
\caption{Examples of training image pairs used during the first training stage. On the right, we represent the injective mask, where yellow areas are equal to one while purples ones are zero.}
\label{fig:mask-examples}
\end{figure}

We show qualitative examples of the training image pairs used during the first training stage as well as the corresponding training mask (opposite of the injective mask) in Fig.~\ref{fig:mask-examples}. 

\subsection{Training datasets}

Due to the limited amount of available real correspondence data, most matching methods resort to self-supervised training, relying on synthetic image warps generated automatically. We here provide details on the synthetic dataset that we use for self-supervised training, as well as additional information on the implementation of the perturbation data (Sec.~\ref{subsec:perturbed-data} of the main paper). Finally, we also describe the generation of the sparse ground-truth correspondence data from the MegaDepth dataset~\cite{megadepth}. 

\parsection{Base synthetic dataset}  For our base synthetic dataset, we use the same data as in~\cite{GOCor}. Specifically, pairs of images are created by warping a collection of images from the DPED~\cite{Ignatov2017}, CityScapes~\cite{Cordts2016} and ADE-20K~\cite{Zhou2019} datasets, according to synthetic affine, TPS and homography transformations. The transformation parameters are the ones originally used in DGC-Net~\cite{Melekhov2019}. 

These image pairs are further augmented with additional random independently moving objects. To do so, the objects are sampled from the COCO dataset~\cite{coco}, and inserted on top of the images of the synthetic data using their segmentation masks. To generate motion, we randomly sample different affine transformation parameters for each foreground object, which are independent of the background transformations.
It results in 40K image pairs, cropped at resolution $520 \times 520$.

\begin{figure}[t]
\centering
\includegraphics[width=0.49\textwidth]{ 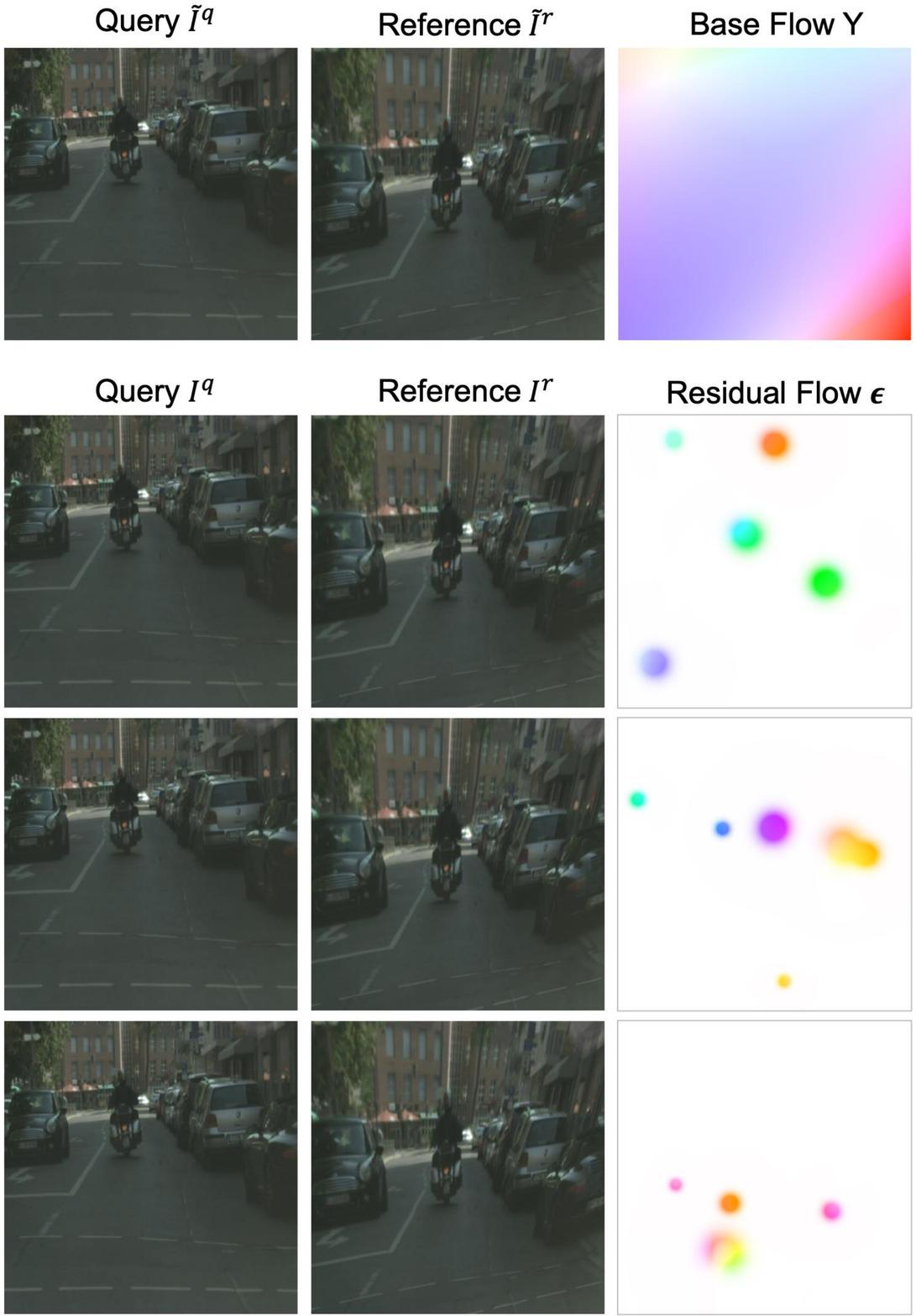}
\caption{Visualization of our perturbations applied to a pair of reference and query images (Sec.~\ref{subsec:perturbed-data} of the main paper). }\vspace{-2mm}
\label{fig:pertu}
\end{figure}

\parsection{Perturbation data for robust uncertainty estimation} Even with independently moving objects, the network still learns to primarily rely on interpolation when estimating the flow field and corresponding uncertainty map relating an image pair. We here describe in more details our data generation strategy for more robust uncertainty prediction. From a base flow field $Y \in \reals^{H \times W \times 2}$ relating a reference image $\tilde{I}^r \in \reals^{H \times W \times 3}$ to a query image $\tilde{I}^q \in \reals^{H \times W \times 3}$, we introduce a residual flow $\epsilon = \sum_i \varepsilon_i$, by adding small local perturbations $\varepsilon_i \in \reals^{H \times W \times 2}$.
More specifically, we create the residual flow by first generating an elastic deformation motion field $E$ on a dense grid of dimension $H \times W$, as described in~\cite{Simard2003}. Since we only want to include perturbations in multiple small regions, we generate binary masks $S_i \in \mathbb{R}^{H \times W \times 2}$, each delimiting the area on which to apply one local perturbation $\varepsilon_i$.  The final residual flow (perturbations) thus take the form of $\epsilon = \sum_i \varepsilon_i$, where $\varepsilon_i = E \cdot S_i$. 
Finally, the query image $I^q = \tilde{I}^q$ is left unchanged  while the reference $I^r$ is generated by warping $\tilde{I}^r$ according to the residual flow $\epsilon$, as $I^r(x) = \tilde{I^r}(x+\epsilon(x))$ for $x \in \mathbb{R}^2$. The final perturbed flow map $Y$ between $I^r$ and $I^q$ is achieved by composing the base flow $\tilde{Y}$ with the residual flow $\epsilon$, as $Y(x) = \tilde{Y}(x+\epsilon(x)) + \epsilon(x)$.

In practise, for the elastic deformation field $E$, we use the implementation of~\cite{info11020125}. The masks $S_i$ should be between 0 and 1 and offer a smooth transition between the two, so that the perturbations appear smoothly. To create each mask $S_i$, we thus generate a 2D Gaussian centered at a random location and with a random standard deviation (up to a certain value) on a dense grid of size $H \times W$. It is then scaled to 2.0 and clipped to 1.0, to obtain a smooth regions equal to 1.0 where the perturbation will be applied, and transition regions on all sides from 1.0 to 0.0. 

In Fig.~\ref{fig:pertu}, we show examples of generated residual flows and their corresponding perturbed reference $I^r$, for a particular base flow $Y$, and query $\tilde{I}^r$ and reference $\tilde{I}^q$ images. As one can see, for human eye, it is almost impossible to detect the presence of the perturbations on the perturbed reference $I^r$. This will enable to "fool" the network in homogeneous regions, such as the road in the figure example, thus forcing it to predict high uncertainty in regions where it cannot identify them.

\parsection{MegaDepth training} To generate the training pairs with sparse ground-truth, we adapt the generation protocol of D2-Net~\cite{Dusmanu2019CVPR}. Specifically, we use the MegaDepth dataset, consisting of 196 different scenes reconstructed from 1.070.468 internet photos using COLMAP~\cite{SchonbergerF16}. The camera intrinsics and extrinsics as well as depth maps from Multi-View Stereo are provided by the authors for 102.681 images.

For training, we use 150 scenes and sample up to 300 random images with an overlap ratio of at least 30\% in the sparse SfM point cloud at each epoch. For each pair, all points of the second image with depth information are projected into the first image. A depth-check with respect to the depth map of the first image is also run to remove occluded pixels. 
For the validation dataset, we sample up to 25 image pairs from 25 different scenes. We resize the images so that their largest dimension is 520.

During the second stage of training, we found it crucial to train on \emph{both} the synthetic dataset with perturbations and the sparse data from MegaDepth. Training solely on the sparse correspondences resulted in less reliable uncertainty estimates.

\section{Architecture details}
\label{sec-sub:arch-details}

In this section, we first describe the architecture of our proposed uncertainty decoder (Sec.~\ref{sec:uncertainty-arch} of the main paper). We then give additional details about our proposed final architecture PDC-Net+, as well as the initial PDC-Net and its corresponding baseline GLU-Net-GOCor*. We also describe the architecture of BaseNet and its probabilistic derivatives, employed for the ablation study. Finally, we share all training details and hyper-parameters.

\subsection{Architecture of the uncertainty decoder}
\label{subsec-sup:unc-dec}

\parsection{Correlation uncertainty module} We first describe the architecture of our Correlation Uncertainty Module $U_{\theta}$ (Sec.~\ref{sec:uncertainty-arch} of the main paper). The correlation uncertainty module processes each 2D slice $C_{ij\cdot \cdot}$ of the correspondence volume $C$ independently. 
More practically, from the correspondence volume tensor $C \in \mathbb{R}^ {b \times h \times w \times (d \times d)}$, where b indicates the batch dimension, we move the spatial dimensions $h \times w$ into the batch dimension and we apply multiple convolutions in the displacement dimensions $d \times d$, \ie on a tensor of shape $(b \times h \times w) \times d \times d \times 1$. By applying the strided convolutions, the spatial dimension is gradually decreased, resulting in an uncertainty representation $u \in \mathbb{R}^{(b \times h \times w) \times 1 \times 1 \times n}$, where $n$ denotes the number of channels. $u$ is subsequently rearranged, and after dropping the batch dimension, the outputted uncertainty tensor is $u \in \mathbb{R}^{h \times w \times n}$.

Note that while this is explained for a local correlation, the same applies for a global correlation except that the displacement dimensions correspond to $h \times w$. 
In Tab.~\ref{tab:arch-local-disp}-~\ref{tab:arch-global-disp}, we present the architecture of the convolution layers applied on the displacements dimensions, for a local correlation with search radius 4 and for a global correlation applied at dimension $ h \times w = 16 \times 16$, respectively.

\begin{table}[b]
\centering
\caption{Architecture of the correlation uncertainty module for a local correlation, with a displacement radius of 4. Implicit are the BatchNorm and ReLU operations that follow each convolution, except for the last one. K refers to kernel size, s is the used stride and p the padding. }
\resizebox{0.49\textwidth}{!}{%
\begin{tabular}{l|l|l}
\toprule
Inputs & Convolutions & Output size \\ \midrule
C; $(b \times h \times w) \times 9 \times 9 \times 1$ & $conv_0$, $K=(3 \times 3)$, s=1, p=0 &  $(b \times h \times w) \times 7 \times 7 \times 32$ \\ \midrule
$conv_0$; $(b \times h \times w) \times 7 \times 7 \times 32$ & $conv_1$, $K=(3 \times 3)$, s=1, p=0 &  $(b \times h \times w) \times 5 \times 5 \times 32$ \\ \midrule
$conv_1$; $(b \times h \times w) \times 5 \times 5 \times 32$ & $conv_2$, $K=(3 \times 3)$, s=1, p=0 &  $(b \times h \times w) \times 3 \times 3 \times 16$ \\ \midrule
$conv_2$; $(b \times h \times w) \times 3 \times 3 \times 16$ & $conv_3$, $K=(3 \times 3)$, s=1, p=0 &  $(b \times h \times w) \times 1 \times 1 \times n$ \\ 
\bottomrule
\end{tabular}%
}
\label{tab:arch-local-disp}
\end{table}

\begin{table}[b]
\centering
\caption{Architecture of the correlation uncertainty module for a global correlation, constructed at resolution $16 \times 16$. Implicit are the BatchNorm and ReLU operations that follow each convolution, except for the last one. K refers to kernel size, s is the used stride and p the padding. }
\resizebox{0.49\textwidth}{!}{%
\begin{tabular}{l|l|l}
\toprule
Inputs & Convolutions & Output size \\ \midrule
C $(b \times h \times w) \times 16 \times 16 \times 1$ & $conv_0$;  $K=(3 \times 3)$, s=1, p=0  &  $(b \times h \times w) \times 14 \times 14 \times 32$ \\ \midrule
$conv_0$; $(b \times h \times w) \times 14 \times 14 \times 32$ &  \begin{tabular}[c]{@{}l@{}}$3 \times 3$ max pool, s=2 \\ $conv_1$, $K=(3 \times 3)$, s=1, p=0\end{tabular}  & $(b \times h \times w) \times 5 \times 5 \times 32$ \\ \midrule

$conv_1$; $(b \times h \times w) \times 5 \times 5 \times 32$ & $conv_2$, $K=(3 \times 3)$, s=1, p=0 &  $(b \times h \times w) \times 3 \times 3 \times 16$ \\ \midrule
$conv_2$; $(b \times h \times w) \times 3 \times 3 \times 16$ & $conv_3$,  $K=(3 \times 3)$, s=1, p=0 &  $(b \times h \times w) \times 1 \times 1 \times n$ \\ 
\bottomrule
\end{tabular}%
}
\label{tab:arch-global-disp}
\end{table}

\parsection{Uncertainty predictor} We then give additional details of the Uncertainty Predictor, that we denote $Q_{\theta}$ (Sec.~\ref{sec:uncertainty-arch} of the main paper). 
The uncertainty predictor takes the flow field $Y \in \reals^{h \times w \times 2}$ outputted from the flow decoder, along with the output $u \in \reals^{h \times w \times n}$ of the correlation uncertainty module $U_{\theta}$. In a multi-scale architecture, it additionally takes as input the estimated flow field and predicted uncertainty components from the previous level. At level $l$, for each pixel location $(i,j)$, this is expressed as:
\begin{equation}
\big((\tilde{\alpha}_m )_{m=1}^M, ( h_m  )_{m=1}^M  \big)^l = Q_{\theta} \big( Y^l; u^l;  \Phi^{l-1} \big)_{ij}
\end{equation}
where $\tilde{\alpha}_m$ refers to the output of the uncertainty predictor, which is then passed through a SoftMax layer to obtain the final weights $\alpha_m$. $\sigma^2_m$ is obtained from $h_m$ according to constraint equation \eqref{eq:constraint} of the main paper. 

In practise, we have found that instead of feeding the flow field $Y \in \reals^{h \times w \times 2}$ outputted from the flow decoder to the uncertainty predictor, using the second last layer from the flow decoder leads to slightly better results. This is because the second last layer from the flow decoder has larger channel size, and therefore encodes more information about the estimated flow.

Architecture-wise, the uncertainty predictor $Q_{\theta}$ consists of 3 convolutional layers. The numbers of feature channels at each convolution layers are respectively 32, 16 and $2M$ and the spatial kernel of each convolution is $3 \times 3$ with stride of 1 and padding 1. The first two layers are followed by a batch-normalization layer with a leaky-Relu non linearity. The final output of the uncertainty predictor is the result of a linear 2D convolution, without any activation.

\subsection{Architecture of PDC-Net+}
\label{subsec-sup:pdcnet}

Our PDC-Net+ has the same architecture than the initial PDC-Net~\cite{pdcnet}. Specifically, we use GLU-Net-GOCor~\cite{GLUNet, GOCor} as our base architecture, predicting the dense flow field relating a pair of images. It is a 4 level pyramidal network, using a VGG feature backbone. It is composed of two sub-networks, L-Net and H-Net which act at two different resolutions. 
The L-Net takes as input rescaled images to $H_L \times W_L= 256 \times 256$ and process them with a global GOCor module followed by a local GOCor module. The resulting flow is then upsampled to the lowest resolution of the H-Net to serve as initial flow, by warping the query features according to the estimated flow. The H-Net takes input images at unconstrained resolution $H\times W$, and refines the estimated flow with two local GOCor modules. However, as opposed to original GLU-Net-GOCor~\cite{GLUNet, GOCor}, we here use residual connections layers instead of DenseNet connections~\cite{Huang2017} and feed-forward layers for the flow and mapping decoders respectively. 

From the baseline GLU-Net-GOCor, we create our probabilistic approach PDC-Net+ by inserting our uncertainty decoder at each pyramid level. As noted in~\ref{subsec-sup:unc-dec}, in practise, we feed the second last layer from the flow decoder to the uncertainty predictor of each pyramid level instead of the predicted flow field. It leads to slightly better results. 
The uncertainty prediction is additionally \emph{propagated from one level to the next}. More specifically, the flow decoder takes as input the uncertainty prediction (all parameters $\Phi$ of the predictive distribution except for the mean flow) of the previous level, in addition to its original inputs (which include the mean flow of the previous level). The uncertainty predictor also takes the uncertainty and the flow estimated at the previous level. 
As explained in Sec.~\ref{subsec:arch} of the main paper, we use a constrained mixture with $M=2$ Laplace components. 
The first component is set so that $\sigma^2_1 = 1$, while the second is learned as $ 2 = \beta_2^- \leq \sigma^2_2 \leq \beta_2^+$. Therefore, the uncertainty predictor only estimates $\sigma^2_2$ and $(\alpha_m )_{m=1}^{M=2}$ at each pixel location.
We found that fixing $\sigma^2_1 = \beta_1^- = \beta_1^+ = 1.0$ results in better performance than for example $\beta_1^- = 0.0 \leq \sigma^2_1 \leq \beta_1^+=1.0$. Indeed, in the later case, during training, the network focuses primarily on getting the very accurate, and confident, correspondences (corresponding to $\sigma^2_1$) since it can arbitrarily reduce the variance. Generating fewer, but accurate predictions then dominate during training to the expense of other regions. This is effectively alleviated by setting $\sigma^2_1=1.0$, which can be seen as introducing a strict prior on this parameter. 

\subsection{Inference multi-stage and multi-scale }

Here, we provide implementation details for our multi-stage and multi-scale inference strategy (Sec.~\ref{sec:geometric-inference} of the main paper). 
After the first network forward pass, we select matches with a corresponding confidence probability $P_{R=1}$ superior to 0.1, for $R=1$.  Since the network estimates the flow field at a quarter of the original image resolution, we use the filtered correspondences at the quarter resolution and scale them to original resolution to be used for homography estimation.  To estimate the homography, we use OpenCV’s findHomography with RANSAC and an inlier threshold of 1 pixel.

\subsection{Architecture of BaseNet}
\label{arch-basenet}

As baseline used in our ablation study, we utilize BaseNet, introduced in~\cite{GOCor} and inspired by GLU-Net~\cite{GLUNet}. It estimates the dense flow field relating an input image pair. 
The network is composed of three pyramid levels and it uses VGG-16~\cite{Chatfield14} as feature extractor backbone. The coarsest level is based on a global correlation layer, followed by a mapping decoder estimating the correspondence map at this resolution. The two next pyramid levels instead rely on local correlation layers. The dense flow field is then estimated with flow decoders, taking as input the correspondence volumes resulting from the local feature correlation layers. Moreover, BaseNet is restricted to a pre-determined input resolution $H_L \times W_L = 256 \times 256$ due to its global correlation at the coarsest pyramid level. It estimates a final flow-field at a quarter of the input resolution $H_L \times W_L$, which needs to be upsampled to original image resolution $H \times W$. The mapping and flow decoders have the same number of layers and parameters as those used for GLU-Net~\cite{GLUNet}. However, here, to reduce the number of weights, we use feed-forward layers instead of DenseNet connections~\cite{Huang2017} for the flow decoders.

We create the different probabilistic versions of BaseNet, presented in the ablation study Tab.~\ref{tab:ablation} of the main paper, by modifying the architecture minimally. 
For the network referred to as PDC-Net-s, which also employs our proposed uncertainty architecture (Sec.~\ref{sec:uncertainty-arch} of the main paper), we add our uncertainty decoder at each pyramid layer, in a similar fashion as for our final network PDC-Net+. The flow prediction is modelled with a constrained mixture employing $M=2$ Laplace components. The first component is set so that $\sigma^2_1 = 1$, while the second is learned as $2 = \beta_2^- \leq \sigma^2_2 \leq \beta_2^+ = \infty$.
We train all networks on the synthetic data with the perturbations, which corresponds to our first training stage (Sec.~\ref{subsec:arch}).

\subsection{Implementation details} 

Since we use pyramidal architectures with $K$ levels, we employ a multi-scale training loss, where the loss at different pyramid levels account for different weights. 
\begin{equation}
\label{eq:multiscale-loss}
    \mathcal{L}(\theta)=\sum_{l=1}^{K} \gamma_{l} L_l +\eta \left\| \theta \right\|\,,
\end{equation}
where  $\gamma_{l}$ are the weights applied to each pyramid level and $L_l$ is the corresponding loss computed at each level, which refers to the L1 loss for the non-probabilistic baselines and the negative log-likelihood loss \eqref{eq:nll-sup} for the probabilistic models, including our approach PDC-Net+. The second term of the loss \eqref{eq:multiscale-loss} regularizes the weights of the network. 

For training, we use similar training parameters as in~\cite{GLUNet}. Specifically, as a preprocessing step, the training images are mean-centered and normalized using mean and standard deviation of ImageNet dataset~\cite{Hinton2012}. For all local correlation layers, we employ a search radius $r=4$. 

\parsection{BaseNet} For the training of BaseNet and its probabilistic derivatives (including PDC-Net-s), which have a pre-determined fixed input image resolution of $(H_L \times W_L = 256 \times 256)$, we use a batch size of 32 and train for 106.250 iterations. We set the initial learning rate to $10^{-2}$ and gradually decrease it by 2 after 56.250, 75.000 and 93.750 iterations. 
The weights in the training loss \eqref{eq:multiscale-loss} are set to be $\gamma_{1}=0.32, \gamma_{2}=0.08, \gamma_{3}=0.02$ and to compute the loss, we down-sample the ground-truth to estimated flow resolution at each pyramid level. 

\parsection{GLU-Net based architectures (GLU-Net-GOCor*, PDC-Net, PDC-Net+)} For GLU-Net-GOCor*, PDC-Net and PDC-Net+, we down-sample and scale the ground truth from original resolution $H\times W$ to $H_L \times W_L$ in order to obtain the ground truth flow fields for L-Net.
During the first stage of training, \ie on purely synthetic images, we down-sample the ground truth from the base resolution to the different pyramid resolutions without further scaling, so as to obtain the supervision signals at the different levels. During this stage, the weights in the training loss \eqref{eq:multiscale-loss} are set to be $\gamma_{1}=0.32, \gamma_{2}=0.08, \gamma_{3}=0.02, \gamma_{4}=0.01$, which ensures that the loss computed at each pyramid level contributes equally to the final loss \eqref{eq:multiscale-loss}.  
During the second stage of training however, which includes MegaDepth, since the ground-truth is sparse, it is inconvenient to down-sample it to different resolutions. We thus instead up-sample the estimated flow field to the ground-truth resolution and compute the loss at this resolution. In practise, we found that both strategies lead to similar results during the self-supervised training. During the second training stage,  the weights in the training loss \eqref{eq:multiscale-loss} are instead set to $\gamma_{1}=0.08, \gamma_{2}=0.08, \gamma_{3}=0.02, \gamma_{4}=0.02$, which also ensures that the loss terms of all pyramid levels have the same magnitude. 

\parsection{PDC-Net+} We train on images pairs of size $520 \times 520$. The first training stage involves 350k iterations, with a batch size of 15. The learning rate is initially set to $10^{-4}$, and halved after 133k and 240k iterations. 
For the second training stage, the batch size is reduced to 10 due to the added memory consumption when fine-tuning the backbone feature extractor. We train for 225K iterations in this stage. The initial learning rate is set to $5 \cdot 10^{-5}$ and halved after 210K iterations. The total training takes about 10 days on two NVIDIA TITAN RTX. 
For the GOCor modules~\cite{GOCor}, we train with 3 local and global optimization iterations. Moreover, in the case of PDC-Net+, a different VGG backbone is used for each sub-module, which act at different image resolution while the same VGG is used in both sub-modules for PDC-Net.

\parsection{PDC-Net and GLU-Net-GOCor*} During the first training stage on uniquely the self-supervised data, we train for 135K iterations, with batch size of 15. The learning rate is initially equal to $10^{-4}$, and halved after 80K and 108K iterations. Note that during this first training stage, the feature back-bone is frozen, but further finetuned during the second training stage. 
While finetuning on the composition of MegaDepth and the synthetic dataset, the batch size is reduced to 10 and we further train for 195K iterations. The initial learning rate is fixed to $5.10^{-5}$ and halved after 120K and 180K iterations. The feature back-bone is also finetuned according to the same schedule, but with an initial learning rate of $10^{-5}$.
For the GOCor modules~\cite{GOCor}, we train with 3 local and global optimization iterations. 

Our system is implemented using Pytorch~\cite{pytorch} and our networks are trained using Adam optimizer~\cite{adam} with weight decay of $0.0004$.

\section{Experimental setup and datasets}
\label{sec-sup:details-evaluation}

In this section, we first provide details about the evaluation datasets and metrics. We then explain the experimental set-up in more depth. 

\subsection{Evaluation metrics}

\parsection{AEPE} AEPE is defined as the Euclidean distance between estimated and ground truth flow fields, averaged over all valid pixels of the reference image.

\parsection{PCK} The Percentage of Correct Keypoints (PCK) is computed as the percentage of correspondences $\mathbf{\tilde{x}}_{j}$ with an Euclidean distance error $\left \| \mathbf{\tilde{x}}_{j} - \mathbf{x}_{j}\right \| \leq  T$, w.r.t.\ to the ground truth $\mathbf{x}_{j}$, that is smaller than a threshold $T$.

\parsection{Fl} Fl designates the percentage of outliers averaged over all valid pixels of the dataset~\cite{Geiger2013}. They are defined as follows, where $Y$ indicates the ground-truth flow field and $\hat{Y}$ the estimated flow by the network.
\begin{equation}
    Fl = \frac{    \left \| Y-\hat{Y} \right \| > 3  \textrm{ and }   \frac{\left \| Y-\hat{Y}  \right \|}{\left \| Y \right \|} > 0.05 } {\textrm{\#valid pixels}}
\end{equation}

\parsection{Sparsification Errors} We compute the sparsification error curve on each image pair, and normalize it to $[0,1]$. The final error curve is the average over all image pairs of the dataset.

\parsection{mAP} For the task of pose estimation, we use mAP as the evaluation metric, following~\cite{OANet}. The absolute rotation error $\left | R_{err}  \right |$ is computed as the absolute value of the rotation angle needed to align ground-truth rotation matrix $R$ with estimated rotation matrix $\hat{R}$, such as
\begin{equation}
    R_{err} = cos^{-1}\frac{Tr(R^{-1}\hat{R}) -1}{2} \;,
\end{equation} 
where operator $Tr$ denotes the trace of a matrix. The translation error $T_{err}$ is computed similarly, as the angle to align the ground-truth translation vector $T$ with the estimated translation vector $\hat{T}$. 
\begin{equation}
    T_{err} = cos^{-1}\frac{T \cdot \hat{T}}{\left \| T \right \|\left \| \hat{T} \right \|} \;,
\end{equation}
where $\cdot$ denotes the dot-product. The accuracy Acc-$\kappa$ for a threshold $\kappa$ is computed as the percentage of image pairs for which the maximum of $T_{err}$ and $\left | R_{err}  \right |$ is below this threshold. mAP is defined according to original implementation~\cite{OANet}, \ie mAP @5\textdegree\, is equal to Acc-5, mAP @10\textdegree\,  is the average of Acc-5 and Acc-10, while mAP @20\textdegree\,  is the average over Acc-5, Acc-10, Acc-15 and Acc-20. 

\parsection{AUC} AUC denotes the Area Under the Cumulative error plot, where the error is the maximum of $T_{err}$ and $\left| R_{err} \right|$.

\subsection{Evaluation datasets and set-up}
\label{details-eval-data}

\parsection{MegaDepth} The MegaDepth dataset depicts real scenes with extreme viewpoint changes. No real ground-truth correspondences are available, so we use the result of SfM reconstructions to obtain sparse ground-truth correspondences. We follow the same procedure and test images than~\cite{RANSAC-flow}. More precisely, we randomly sample 1600 pairs of images that shared more than 30 points. The test pairs are from different scenes than the ones we used for training and validation. We use 3D points from SfM reconstructions and project them onto the pairs of matching images to obtain correspondences. It results in approximately 367K correspondences. During evaluation, following~\cite{RANSAC-flow}, all the images and ground-truth flow fields are resized to have minimum dimension 480 pixels. The PCKs are calculated per dataset.

\parsection{RobotCar} In RobotCar, we used the correspondences originally introduced by~\cite{RobotCarDatasetIJRR}. During evaluation, following~\cite{RANSAC-flow}, all the images and ground-truth flows are resized to have minimum dimension 480 pixels. The PCKs are calculated per dataset.

\parsection{ETH3D}  The Multi-view dataset ETH3D~\cite{ETH3d} contains 10 image sequences at $480 \times 752$ or $514 \times 955$ resolution, depicting indoor and outdoor scenes. They result from the movement of a camera completely unconstrained, used for benchmarking 3D reconstruction. 
The authors additionally provide a set of sparse geometrically consistent image correspondences (generated by~\cite{SchonbergerF16}) that have been optimized over the entire image sequence using the reprojection error. We sample image pairs from each sequence at different intervals to analyze varying magnitude of geometric transformations, and use the provided points as sparse ground truth correspondences. This results in about 500 image pairs in total for each selected interval, or 600K to 1000K correspondences.  Note that, in this work, following~\cite{GLUNet, GOCor}, we computed the PCK per image and then average per sequence.
This metric is different than the one used for ETH3D in PDC-Net~\cite{pdcnet}, where the PCKs were calculated instead over the whole dataset (image pairs of all the sequences) per interval.  

\parsection{KITTI} The KITTI dataset~\cite{Geiger2013} is composed of real road sequences captured by a car-mounted stereo camera rig. The KITTI benchmark is targeted for autonomous driving applications and its semi-dense ground truth is collected using LIDAR. The 2012 set only consists of static scenes while the 2015 set is extended to dynamic scenes via human annotations.  The later contains  large motion, severe illumination changes, and occlusions. 

\parsection{Pose estimation} 
We  use the selected matches to estimate an essential matrix with RANSAC~\cite{ransac} and 5-pt Nister algorithm~\cite{TPAMI.2004.17}, relying on OpenCV’s 'findEssentialMat' with an inlier threshold of 1 pixel divided by the focal length. Rotation matrix $\hat{R}$ and translation vector $\hat{T}$ are finally computed from the estimated essential matrix, using OpenCV’s 'recoverPose'. 

\parsection{3D reconstruction} We use the set-up of~\cite{SattlerMTTHSSOP18}, which provides a list of image pairs to match. We compute dense correspondences between each pair. We resize the images by keeping the same aspect ratio so that the minimum dimension is 600. We select matches for which the confidence probability $P_{R=1}$ is above 0.1, and feed them to COLMAP reconstruction pipeline~\cite{SchonbergerF16}. Again, we select matches at a quarter of the image resolution and scale the matches to original resolution.

On all datasets, for evaluation, we use 3 and 7 steepest descent iterations in the global and local GOCor modules~\cite{GOCor} respectively. 
We reported results of RANSAC-Flow~\cite{RANSAC-flow} using MOCO features, which gave the best results overall. %

\section{Detailed results}
\label{sec-sup:results}

In this section, we provide detailed results on uncertainty estimation and we present extensive qualitative results and comparisons. We also show additional ablative experiments.

\subsection{Additional results on uncertainty estimation}

\begin{figure}[b]
\centering
\vspace{-6mm}\newcommand{\wid}{0.23\textwidth}
\subfloat{\includegraphics[width=\wid]{ 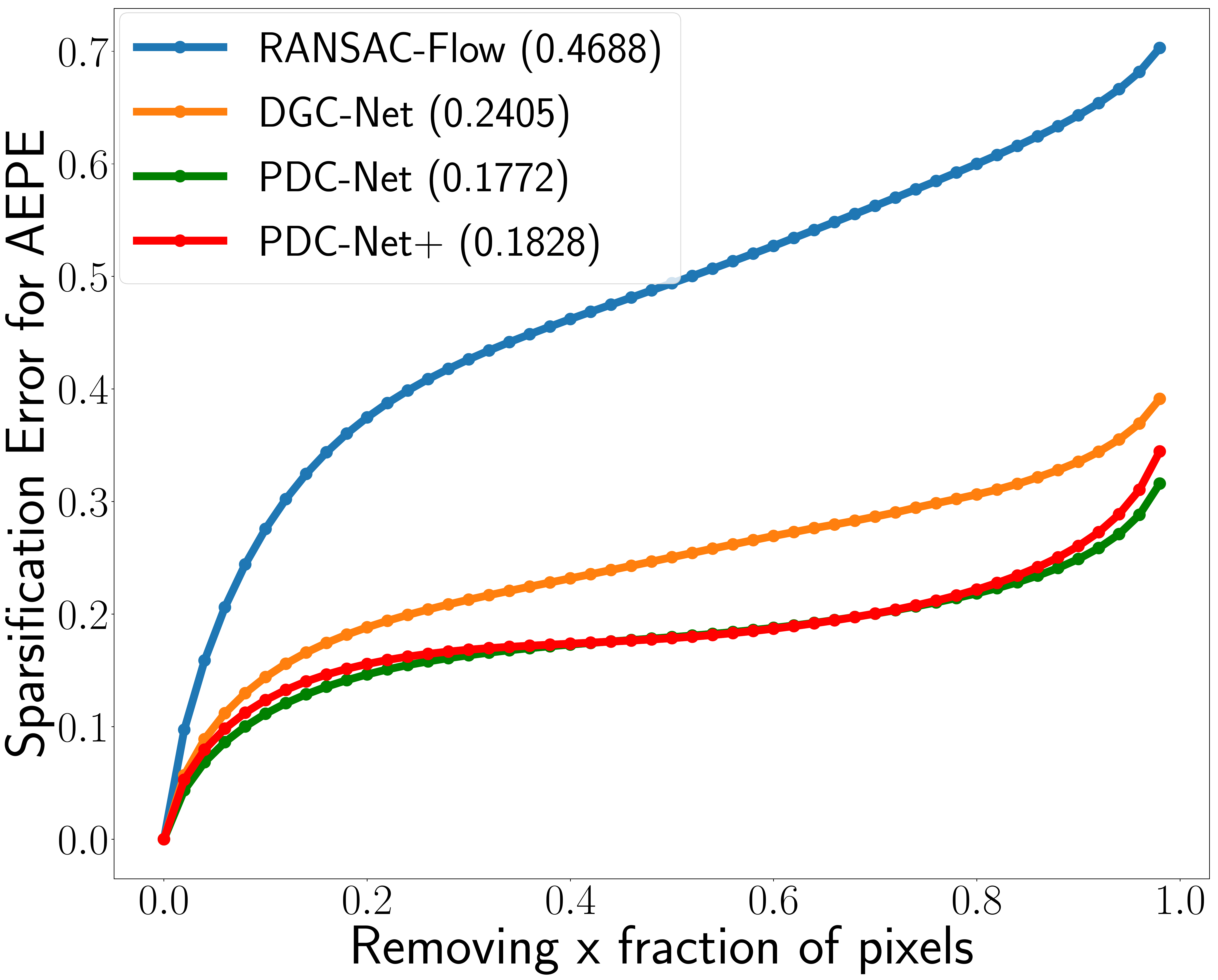}}~%
\subfloat{\includegraphics[width=\wid]{ 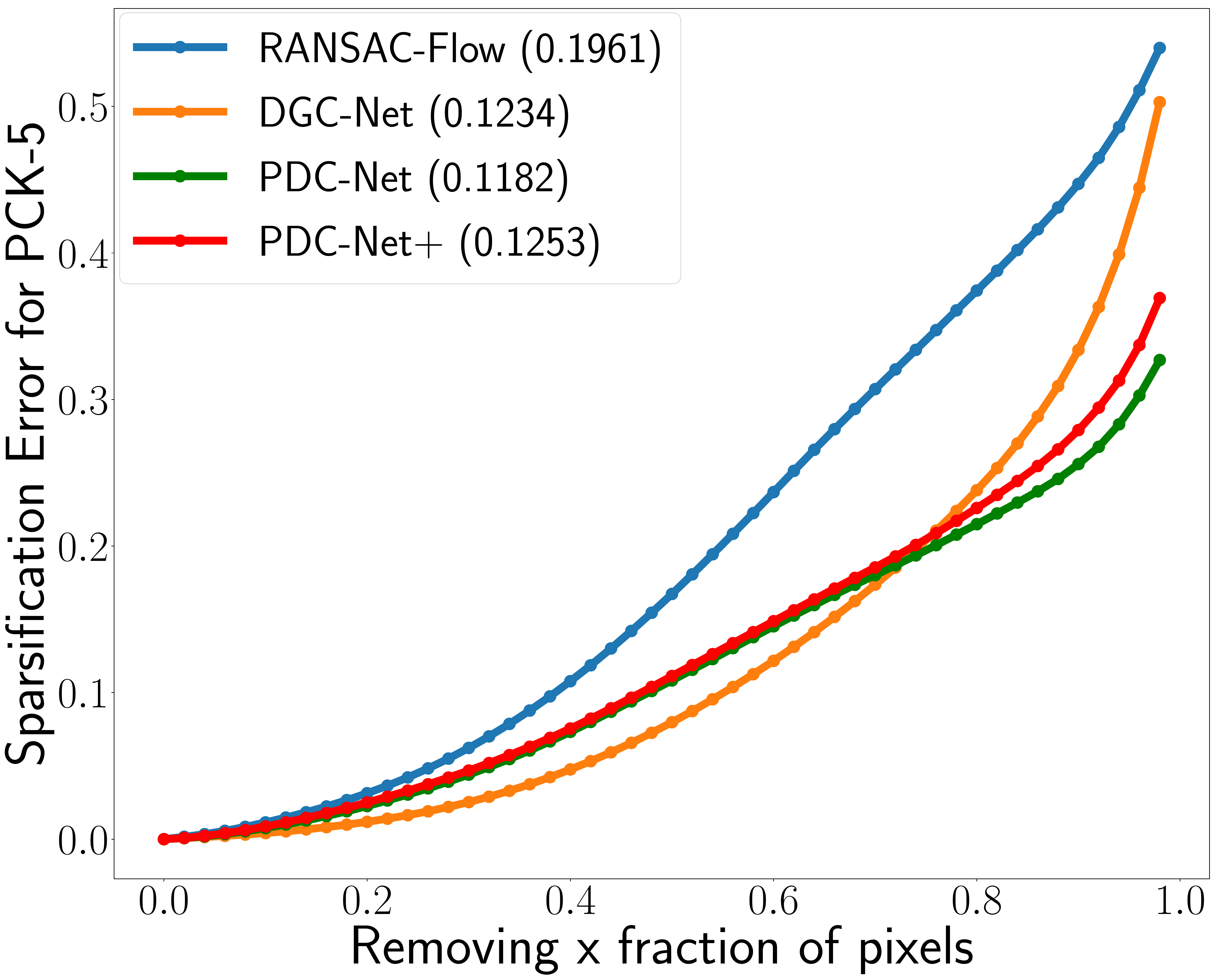}}~%
\caption{Sparsification Error plots for AEPE (left) and PCK-5 (right) on RobotCar.  Smaller AUSE (in parenthesis) is better. }
\label{fig:sparsification-robotcar}
\end{figure}

Here, we present sparsification errors curves, computed on the RobotCar dataset. As in the main paper, Sec.~\ref{subsec:uncertainty-est}, we compare our probabilistic approach PDC-Net+, to dense geometric methods providing a confidence estimation, namely DGC-Net~\cite{Melekhov2019}, RANSAC-Flow~\cite{RANSAC-flow} and the initial PDC-Net~\cite{pdcnet}. Fig.~\ref{fig:sparsification-robotcar} depicts the sparsification error curves on RobotCar. As on MegaDepth, PDC-Net+ obtains better uncertainty than RANSAC-Flow and DGC-Net, but slighly worse than PDC-Net.

\begin{table}[t]
\centering
\caption{Metrics evaluated over scenes of ETH3D with different intervals between consecutive pairs of images (taken by the same camera). Note that, in this work, we computed the PCK per image and we further average the PCKs values over each sequence. This is different than in~\cite{pdcnet}, where the PCKs were computed per sequence instead. High PCK and low AEPE are better. }\label{tab:ETH3d-details}
\resizebox{0.49\textwidth}{!}{%
\begin{tabular}{lccccccc}
\toprule
& 3 & 5 & 7 & 9 & 11 & 13 & 15 \\ \midrule
& \multicolumn{7}{c}{\textbf{AEPE}}  \\ \midrule
LIFE & 1.83 & 2.14 &2.45 &2.8 & 3.42 &4.6 & 8.6 \\
COTR & 1.71 & 1.92 & 2.16 & 2.47 & 2.85 & 3.23 & 3.76 \\
RANSAC-Flow (MS) & 1.79 & 1.94 & 2.09 & 2.25 & \textbf{2.42} & \textbf{2.64} & \textbf{2.87} \\
GLU-Net-GOCor* & 1.68 & 1.92 & 2.18 & 2.43 & 2.89 & 3.32 & 4.27\\
PDC-Net (H) &1.57& 1.77 &1.99 &2.21 &2.49& 2.72& 3.11 \\
PDC-Net (MS) & 1.59 &1.8 & 1.97 & 2.23 & 2.49 & 2.72 & 3.2   \\
\textbf{PDC-Net+ (H)} & \textbf{1.56} &  \textbf{1.74} & \textbf{1.96} & 2.18&  2.48&  2.73 & 3.22 \\
\textbf{PDC-Net+ (MS)} & 1.58 & 1.76 & \textbf{1.96} & \textbf{2.16} & 2.49 & 2.73 & 3.24\\
 \midrule
& \multicolumn{7}{c}{\textbf{PCK-1 (\%)}}  \\\midrule
 
LIFE & 51.93  & 45.98  & 41.46  & 38.05  & 35.17  & 32.85  & 30.35   \\
COTR & - & - & - & - & - & - & -\\
RANSAC-Flow (MS) & 58.33  & 54.71  & 51.56  & 48.64  & 46.14  & 44.0  &   41.75 \\
GLU-Net-GOCor* & 59.4  & 55.15 &51.18 &47.88 &44.46 &41.79 &38.9 \\
PDC-Net (H) & 62.73& 59.44 &56.21 &53.49 &50.86 &48.72 &46.5 \\
PDC-Net (MS) & 62.45 & 59.22 &56.1&  53.29 &50.79& 48.67& 46.37 \\
\textbf{PDC-Net+ (H)} & \textbf{63.12} & \textbf{59.93} & \textbf{56.81} & \textbf{54.12} & \textbf{51.59} & \textbf{49.55} & \textbf{47.32} \\
\textbf{PDC-Net+ (MS)} & 62.95 &59.76 &56.64 &54.02 &51.5  &49.38 &47.24 \\
 \midrule
& \multicolumn{7}{c}{\textbf{PCK-5 (\%)}}  \\\midrule
LIFE & 92.31 &90.54 &88.77 &87.32 &85.74 &84.1 & 81.74 \\
COTR & - & - & - & - & - & - & -\\
RANSAC-Flow (MS) & 91.85 &90.69 &89.45 &88.43& 87.48 &86.39 &85.32 \\
GLU-Net-GOCor* & 93.03 &92.12 &91.04 &90.19 &88.97 &87.8 & 85.92 \\
PDC-Net (H) & 93.52 &92.72 &91.95 &91.24& 90.45 &89.86 &88.98 \\
PDC-Net (MS) & 93.51 &92.72 &91.97 &91.22 &90.48 &89.85& 88.9  \\
\textbf{PDC-Net+ (H)} & \textbf{93.54} & 92.78 & \textbf{92.04}  & 91.30 & \textbf{90.60}  & 89.9 & \textbf{89.03}  \\
\textbf{PDC-Net+ (MS)} & 93.50 & \textbf{92.79} & \textbf{92.04} & \textbf{91.35} & \textbf{90.60} &  \textbf{89.97} & 88.97 \\ 
\bottomrule
\end{tabular}%
}
\end{table}

\subsection{Detailed results on ETH3D}

In Tab.~\ref{tab:ETH3d-details}, we show the detailed results on ETH3D, corresponding to Fig.~\ref{fig:ETH3D} of the main paper. Note that COTR~\cite{COTR} only provides results in terms of AEPE. 

\subsection{Detailed results on HP-240}

\begin{figure}[t]
\centering
\includegraphics[width=0.48\textwidth]{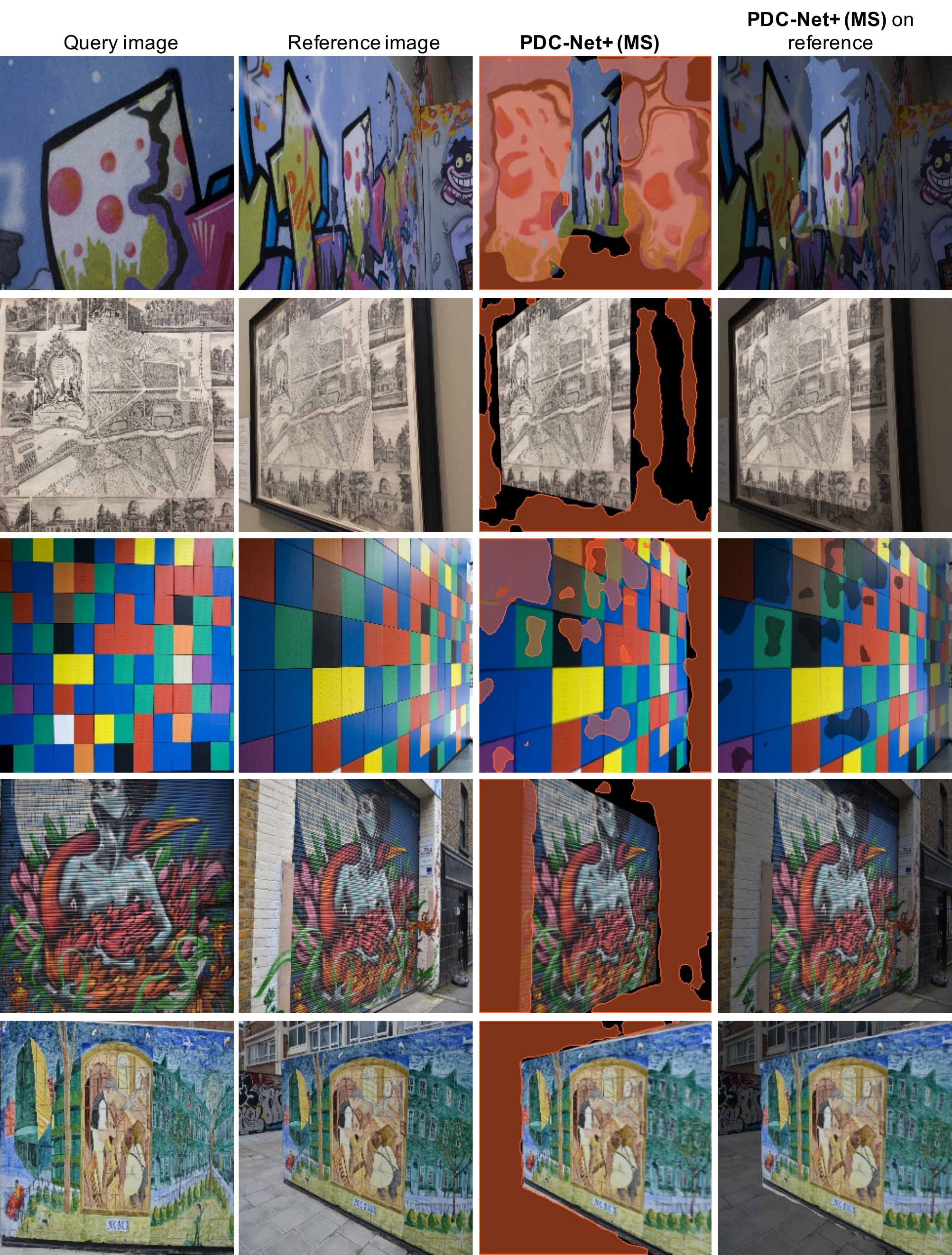}
\caption{Visual examples of PDC-Net+ applied to images of the HPatches dataset~\cite{Lenc}. We warp the query image according to PDC-Net+ and show predicted unreliable or inaccurate matching regions in red. In the last column, we overlay the reference image with the warped query from PDC-Net+, in the identified accurate matching regions (lighter color).}
\label{fig:hp-visual}
\end{figure}

In Tab.~\ref{tab:hp-details}, we show the detailed results on HP-240, corresponding to Tab.~\ref{tab:hp-240} of the main paper. We also show qualitative results of PDC-Net+ applied to some image pairs in Fig.~\ref{fig:hp-visual}. 

\begin{table*}[t]
\centering
\caption{Additional ablation study. We analyse the impact of the number of components $M$ used in the constrained mixture model, on the probabilistic BaseNet (PDC-Net-s).  All metrics are computed with a single forward-pass of the network.
}
\resizebox{0.99\textwidth}{!}{%
\begin{tabular}{lccc|ccc|cc} 
\toprule
& \multicolumn{3}{c}{\textbf{KITTI-2015}} & \multicolumn{3}{c}{\textbf{MegaDepth}} & \multicolumn{2}{c}{\textbf{YFCC100M}}\\
  & EPE  & F1 (\%)  & AUSE & PCK-1 (\%)  & PCK-5 (\%) & AUSE & mAP @5\textdegree & mAP @10\textdegree  \\ \midrule
$M=2$; $\sigma^2_1=1.0$, $2.0 < \sigma^2_2 < \beta_2^+=s^2$ &  6.61 & 31.67 & \textbf{0.208} & 31.83 & \textbf{66.52} & \textbf{0.204} & 33.05 & 44.48 \\
$M=3$; $\sigma^2_1=1.0$, $2.0 < \sigma^2_2 < \beta_2^+=s^2$, $\sigma^2_3 = s^2$ &  \textbf{6.41} & \textbf{30.54} & 0.212 & \textbf{31.89} & 66.10 & 0.214 & \textbf{34.90} & \textbf{45.86} \\ 
\bottomrule
\end{tabular}%
}
\label{tab:ablation-sup}
\end{table*}

\subsection{Qualitative results}

\begin{figure*}
\centering%
\includegraphics[width=0.95\textwidth]{ 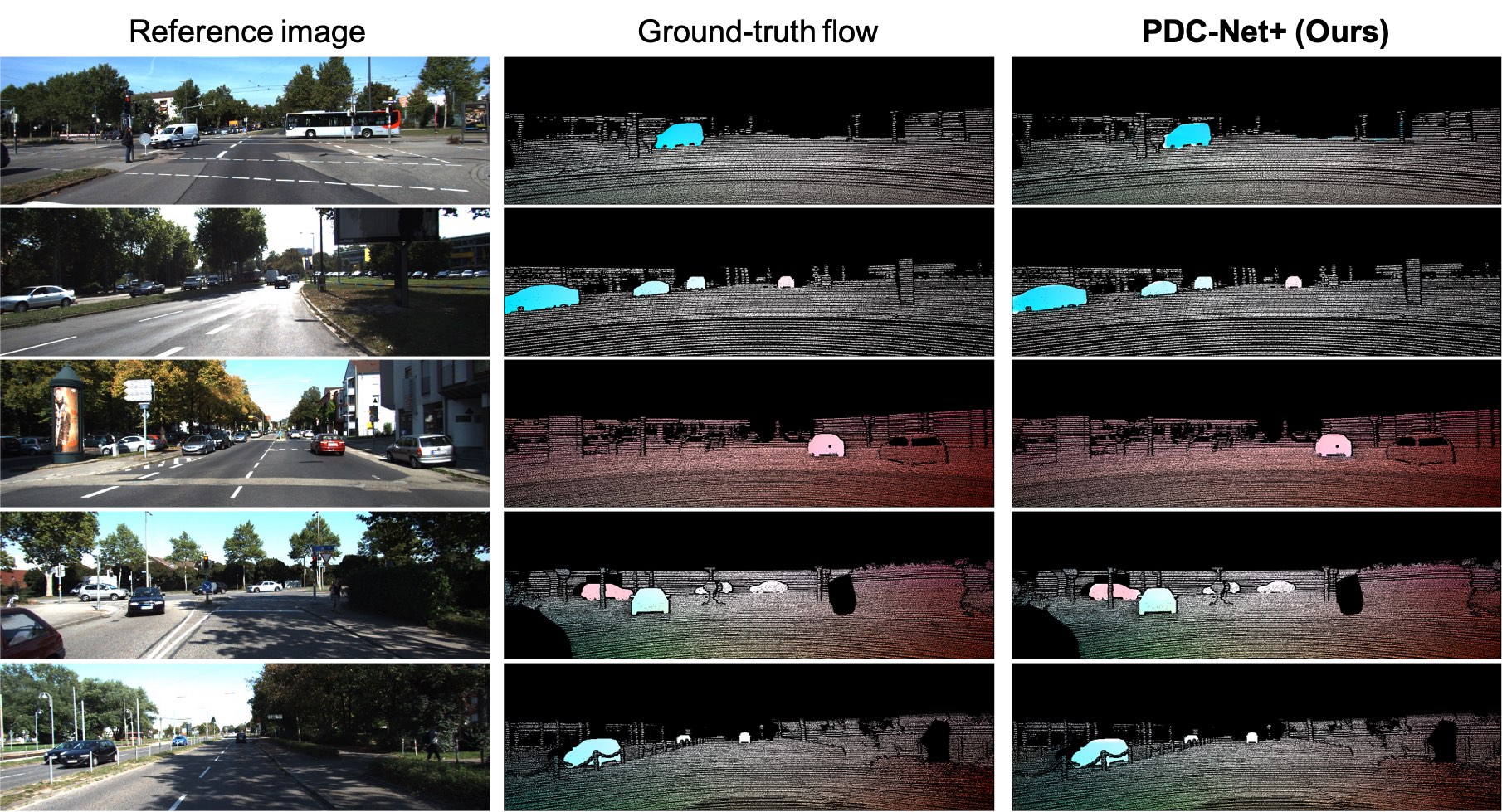}
\caption{Qualitative examples of our approach PDC-Net+ applied to images of KITTI-2015. We plot directly the estimated flow field for each image pair. }
\label{fig:kitti-qual}
\end{figure*}

\begin{figure*}
\centering%
\includegraphics[width=0.95\textwidth]{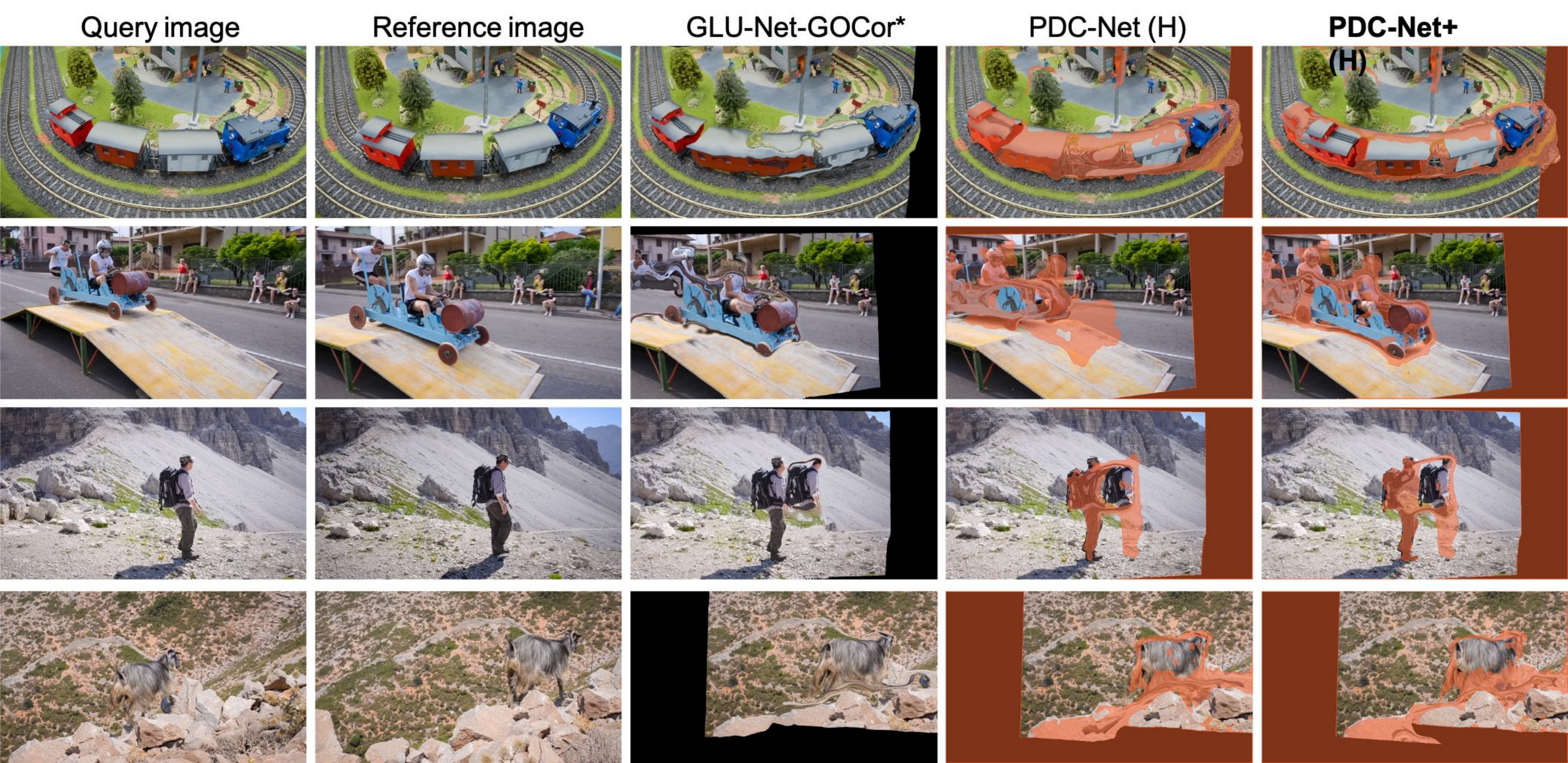}
\caption{Comparison of baseline GLU-Net-GOCor*, PDC-Net and PDC-Net+ applied to images of the segmentation DAVIS dataset~\cite{Pont-Tuset_arXiv_2017}. Scenes feature a moving object taken from a moving camera. PDC-Net and PDC-Net+ also predict a confidence map, according to which the regions represented in red, are unreliable or inaccurate matching regions. }
\label{fig:davis-qual}
\end{figure*}

\parsection{KITTI and data with moving objects}
Here, we first present qualitative results of our approach PDC-Net+ on the KITTI-2015 dataset in Fig.~\ref{fig:kitti-qual}. PDC-Net+ clearly identifies the independently moving objects, and does very well in static scenes with only a single moving object, which are particularly challenging since not represented in the training data. 
We also compare the baseline GLU-Net-GOCor*, PDC-Net and PDC-Net+ on images of the segmentation dataset, DAVIS~\cite{Pont-Tuset_arXiv_2017} in Fig.~\ref{fig:davis-qual}. Each image pair features a moving object, with a motion independent from the back-ground (camera motion). PDC-Net+ better matches the moving object compared to PDC-Net. This is thanks to our enhanced self-supervised pipeline (Sec.~\ref{sec:training-strategy}), integrating multiple independently moving object and to our introduced injective mask (Sec.~\ref{sec:injective-mask}). 

\begin{figure*}
\centering%
\vspace{-25mm}
\includegraphics[width=0.75\textwidth]{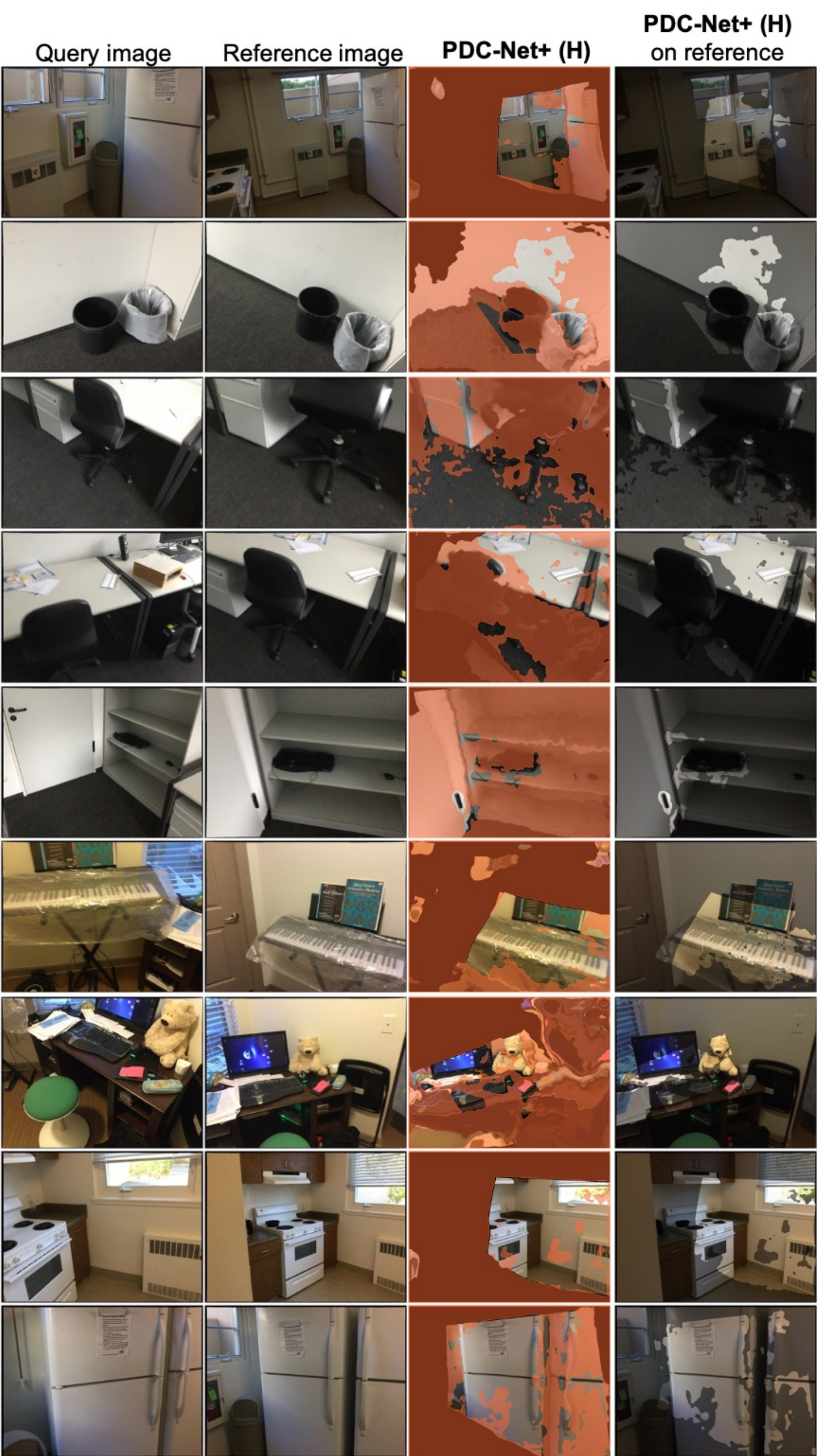} \\
\vspace{-2mm}\caption{Qualitative examples of our approach PDC-Net+ applied to images of the ScanNet dataset~\cite{scannet}. In the 2$^{nd}$ column, we visualize the query images warped according to the flow fields estimated by PDC-Net+. PDC-Net+ also predicts a confidence map, according to which the regions represented in red, are unreliable or inaccurate matching regions. In the last column, we overlay the reference image with the warped query from PDC-Net+, in the identified accurate matching regions (lighter color).}
\vspace{-20mm}
\label{fig:scannet}
\end{figure*}

\begin{figure*}
\centering%
\vspace{-25mm}
\includegraphics[width=0.70\textwidth]{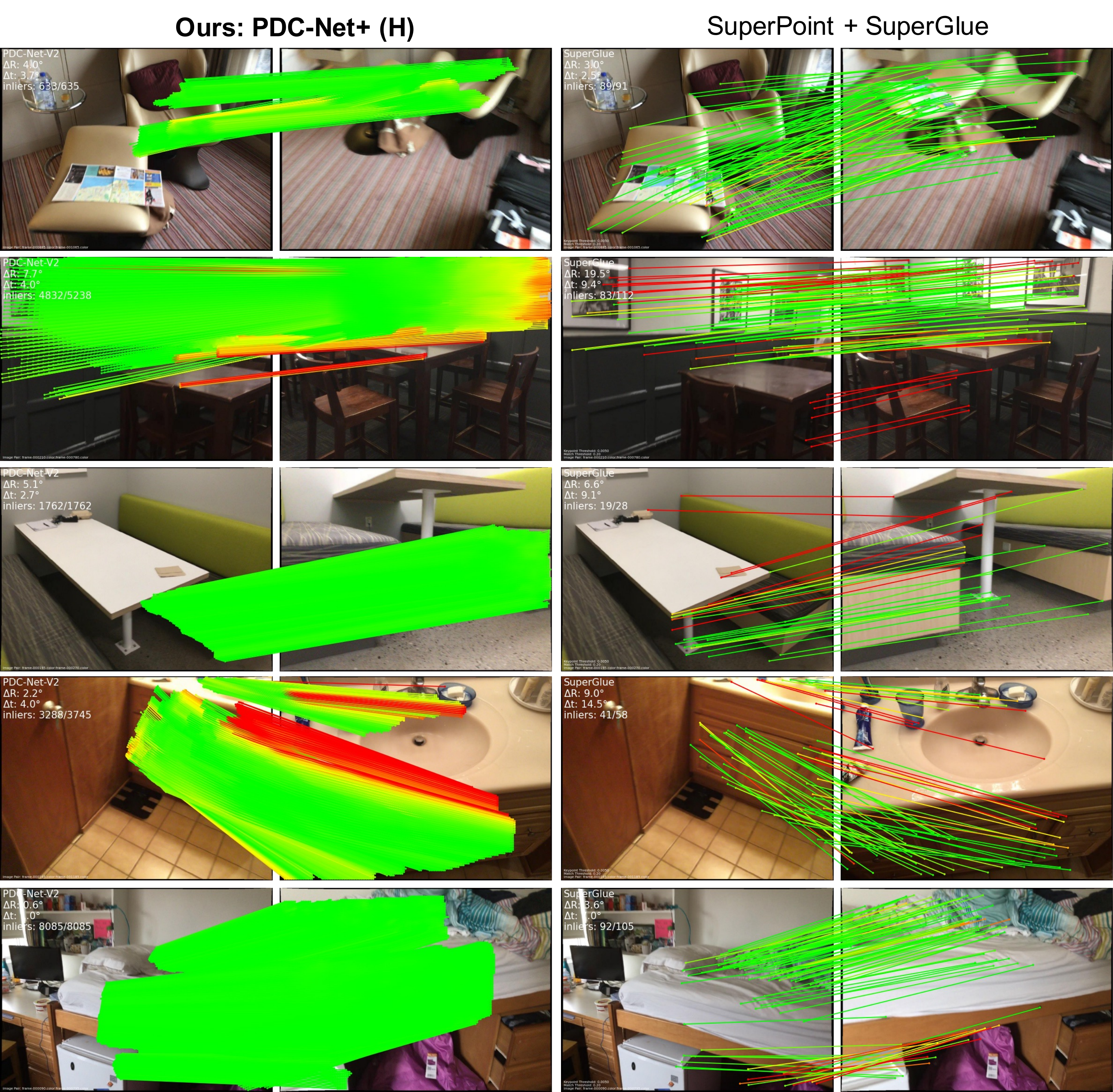} \\
\vspace{1mm}
\includegraphics[width=0.70\textwidth]{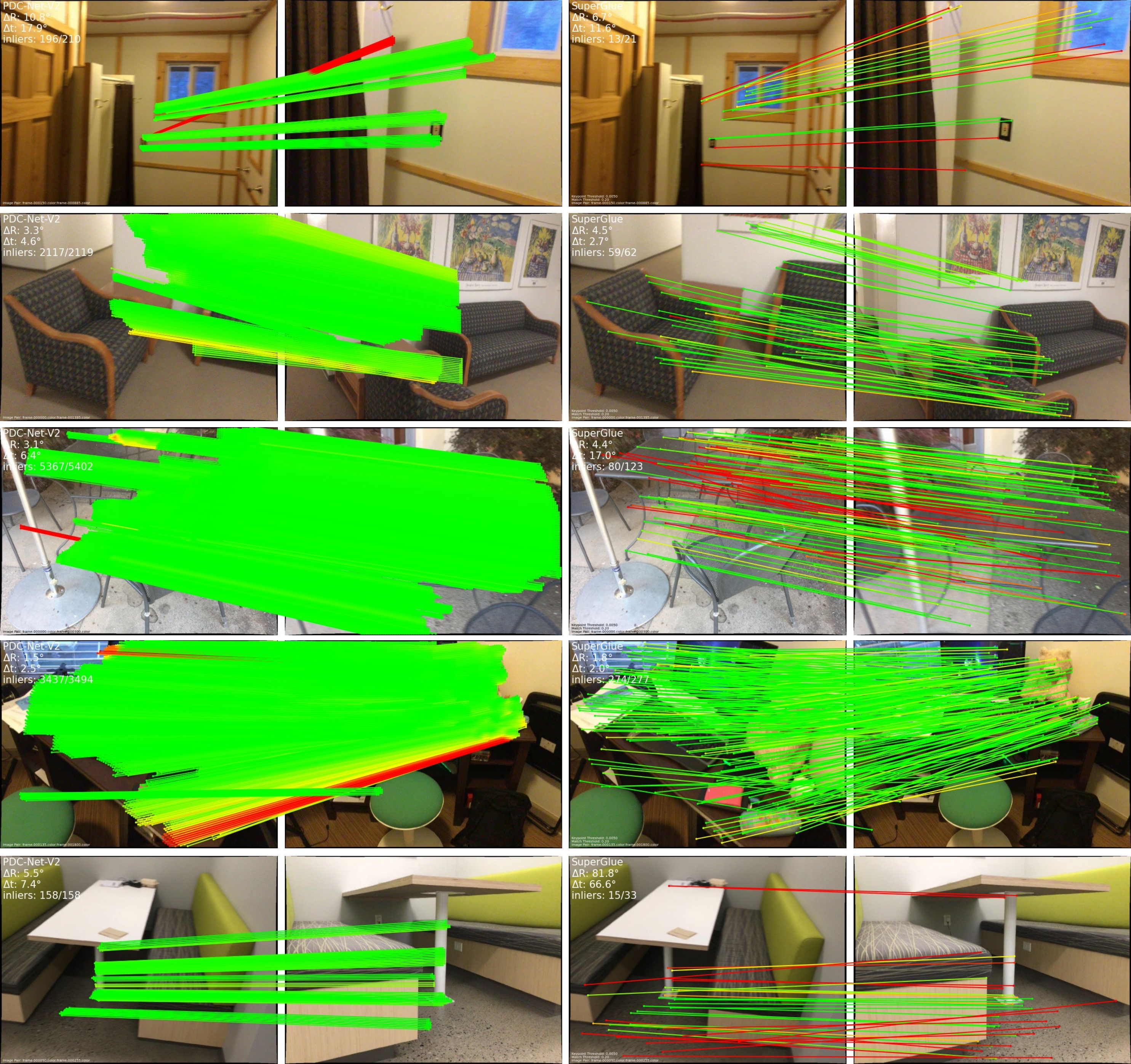} \\
\caption{Comparison of matches found by our PDC-Net+ (left) and Superpoint + SuperGlue (right) on images of ScanNet. Correct matches are green lines and mismatches are red lines. The pose errors are indicated in the top left corner. }
\vspace{-20mm}
\label{fig:scannet-matches-sup}
\end{figure*}

\parsection{ScanNet} We show the performance of PDC-Net+ (trained on MegaDepth) on the ScanNet dataset in Fig.~\ref{fig:scannet}. It can handle very large view-point changes. Notably, it also deals with homogeneous and textureless regions, very common in this dataset. 
We also provide additional comparisons between the matches found by our approach PDC-Net+ and SuperPoint + SuperGlue in~\ref{fig:scannet-matches-sup}. Our approach finds substantially more correct matches than SuperGlue. This is particularly due to the detector, which struggle to find repeatable keypoints in textureless regions. On the contrary, we do not suffer from such limitation. 

\begin{figure*}
\centering%
\vspace{-5mm}\includegraphics[width=0.99\textwidth]{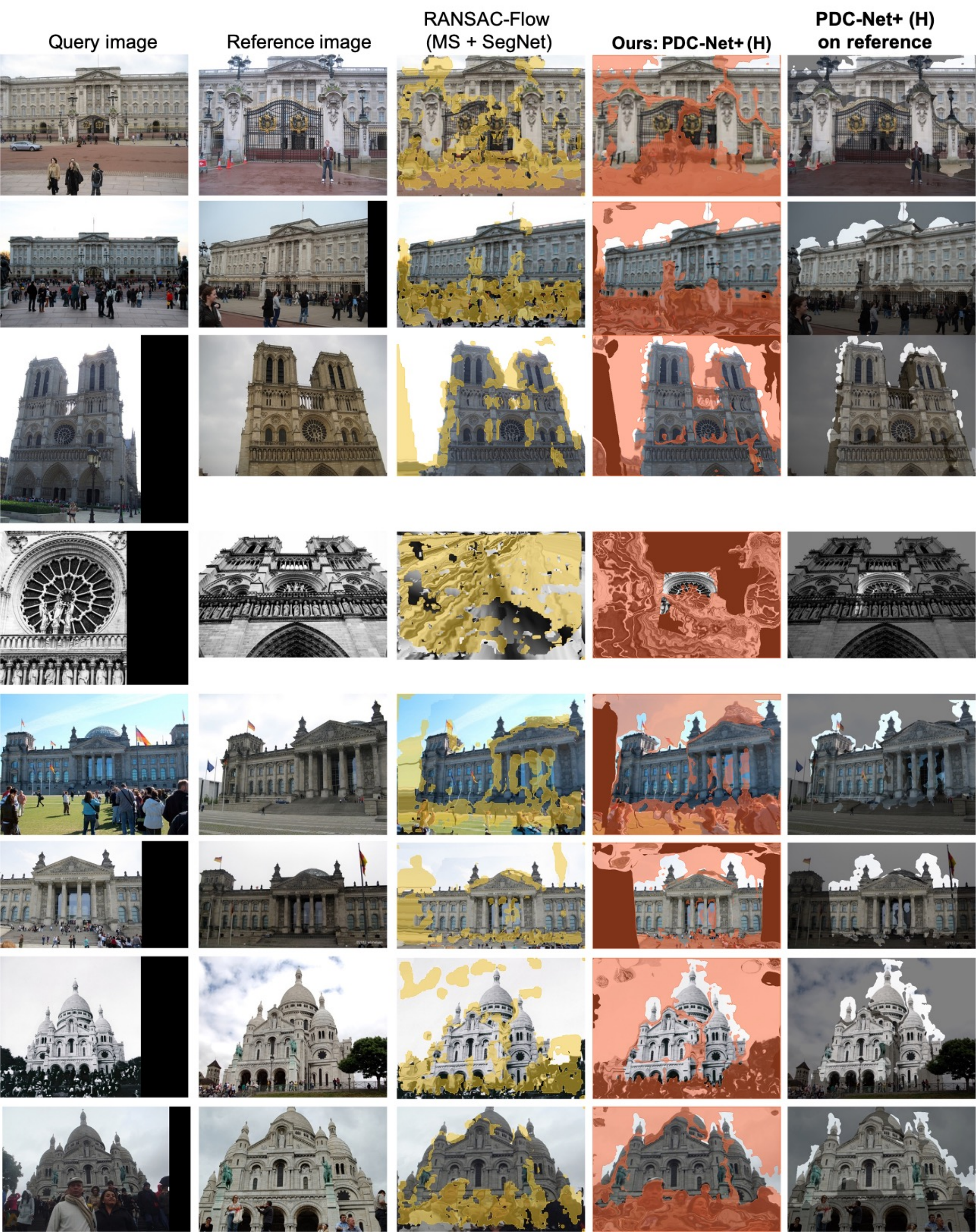}
\vspace{-2mm}
\caption{Visual comparison of RANSAC-Flow and our approach PDC-Net+ on image pairs of the YFCC100M dataset~\cite{YFCC}. In the 3$^{rd}$ and 4$^{th}$ columns, we visualize the query images warped according to the flow fields estimated by the RANSAC-Flow and PDC-Net+ respectively. Both networks also predict a confidence map, according to which the regions represented in respectively yellow and red, are unreliable or inaccurate matching regions. In the last column, we overlay the reference image with the warped query from PDC-Net+, in the identified accurate matching regions (lighter color). }
\label{fig:YCCM-qual}
\end{figure*}

\parsection{YFCC100M} In Fig.~\ref{fig:YCCM-qual}, we visually compare the estimated confidence maps of RANSAC-Flow~\cite{RANSAC-flow} and our approach PDC-Net+ on the YFCC100M dataset. Our confidence maps can accurately \textit{segment} the object from the background (sky). On the other hand, RANSAC-Flow predicts confidence maps, which do not exclude unreliable matching regions, such as the sky. Using these regions for pose estimation for example, would result in a drastic drop in performance, as evidenced in Tab.~\ref{tab:YCCM} of the main paper. 
Note also the ability of our predicted confidence map to identify \emph{small accurate flow regions}, even in a dominantly failing flow field. This is the case in the fourth example from the top in Fig.~\ref{fig:YCCM-qual}.

\begin{figure*}
\centering%
\vspace{-16mm}\includegraphics[width=0.93\textwidth]{ 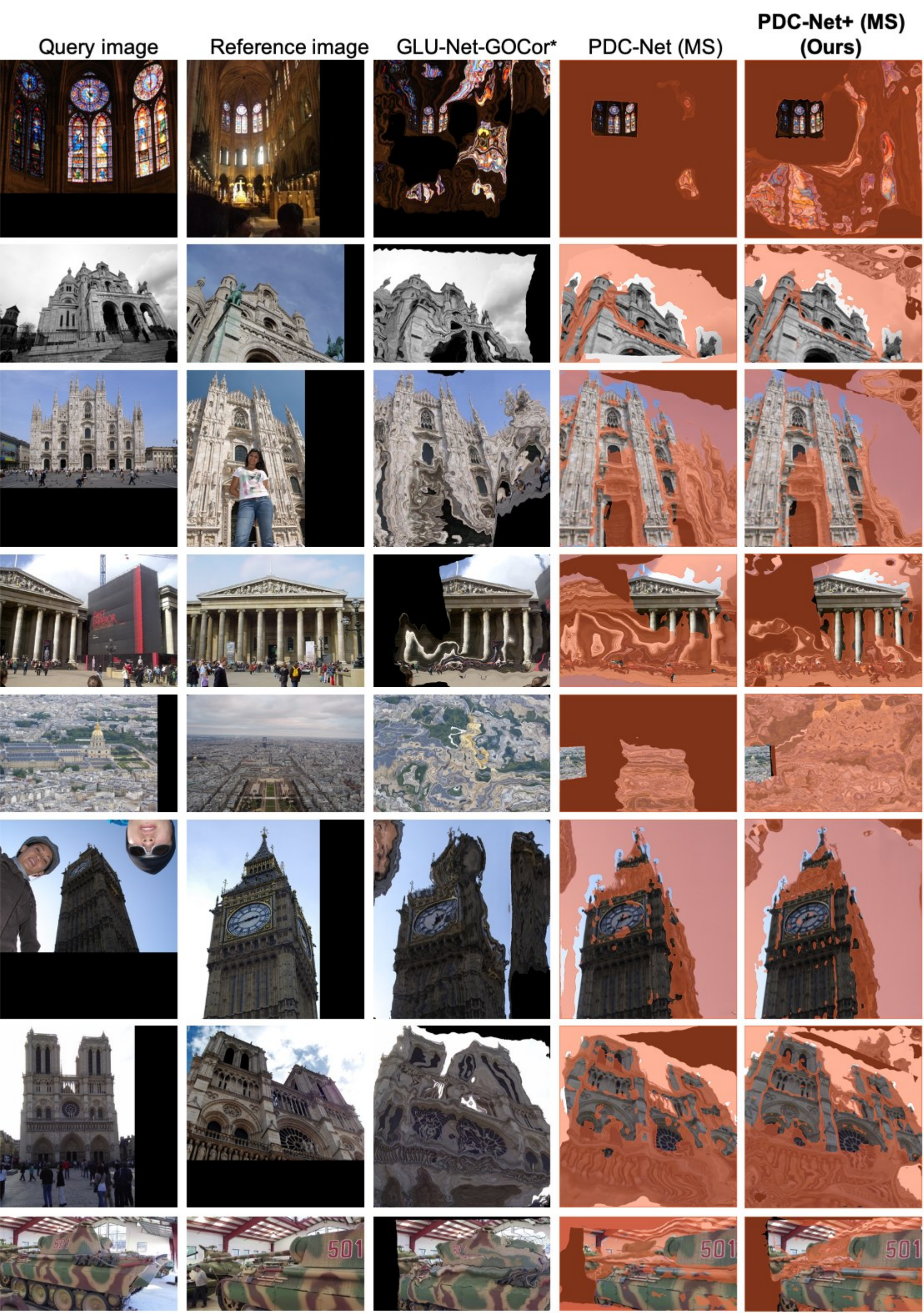}
\vspace{-2mm}\caption{Qualitative examples of our approach PDC-Net+, compared to PDC-Net and corresponding non-probabilistic baseline GLU-Net-GOCor*, applied to images of the MegaDepth dataset~\cite{megadepth}. We visualize the query images warped according to the flow fields estimated by the GLU-Net-GOCor*, PDC-Net and PDC-Net+. The probabilistic networks also predict a confidence map, according to which the regions represented in red, are unreliable or inaccurate matching regions.}
\label{fig:mega-1}
\end{figure*}

\begin{figure*}
\centering%
\vspace{-15mm}\includegraphics[width=0.93\textwidth]{ 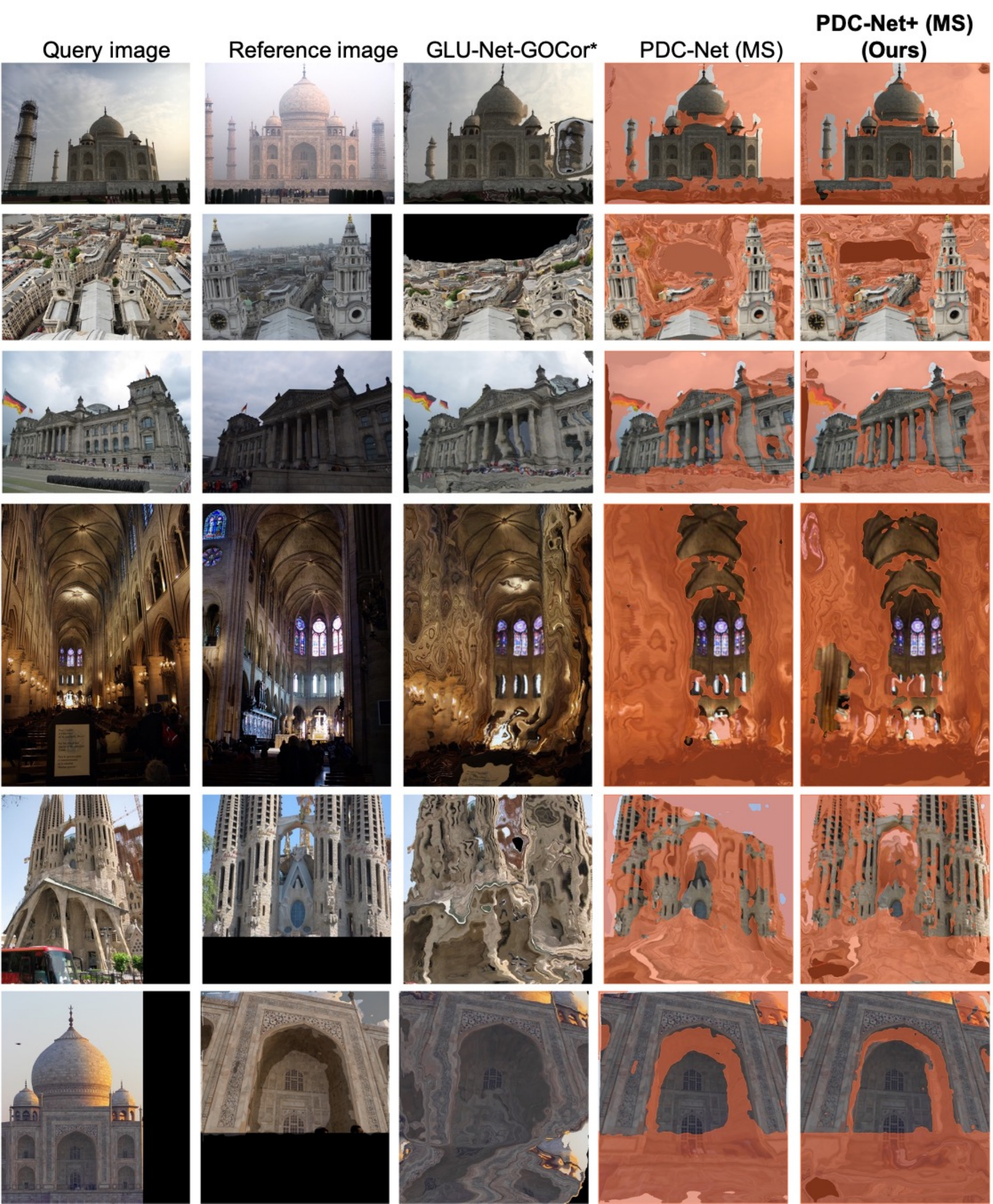}
\vspace{-2mm}\caption{Qualitative examples of our approach PDC-Net+, compared to PDC-Net and corresponding non-probabilistic baseline GLU-Net-GOCor*, applied to images of the MegaDepth dataset~\cite{megadepth}. We visualize the query images warped according to the flow fields estimated by the GLU-Net-GOCor*, PDC-Net and PDC-Net+. The probabilistic networks also predict a confidence map, according to which the regions represented in red, are unreliable or inaccurate matching regions.}
\label{fig:mega-2}
\end{figure*}

\begin{figure*}
\centering%
\vspace{-13mm}\includegraphics[width=0.93\textwidth]{ 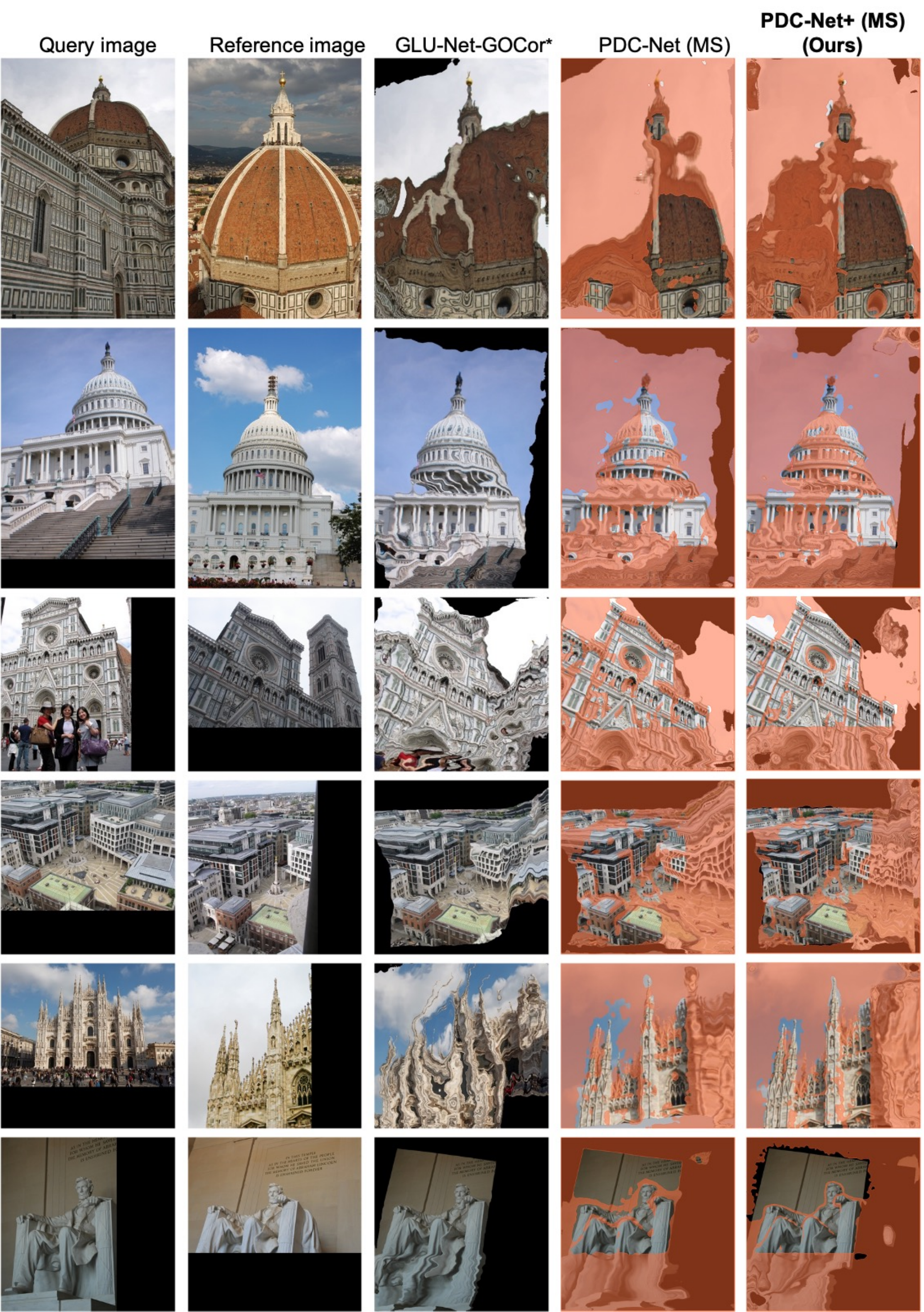}
\vspace{-2mm}\caption{Qualitative examples of our approach PDC-Net+, compared to PDC-Net and corresponding non-probabilistic baseline GLU-Net-GOCor*, applied to images of the MegaDepth dataset~\cite{megadepth}. We visualize the query images warped according to the flow fields estimated by the GLU-Net-GOCor*, PDC-Net and PDC-Net+. The probabilistic networks also predict a confidence map, according to which the regions represented in red, are unreliable or inaccurate matching regions.}
\label{fig:mega-3}
\end{figure*}

\parsection{MegaDepth} In Fig.~\ref{fig:mega-1},~\ref{fig:mega-2},~\ref{fig:mega-3}, we qualitatively compare our approach PDC-Net+ to the initial PDC-Net and to the baseline GLU-Net-GOCor* on images of the Megadepth dataset. We additionally show the performance of our uncertainty estimation on these examples. While PDC-Net+ and PDC-Net generally produce similar outputs, PDC-Net+ is more accurate on some examples. 
By overlaying the warped query image with the reference image at the locations of the identified accurate matches, we observe that our probabilistic formulation produces \emph{highly precise correspondences}. Our uncertainty estimates successfully identify accurate flow regions and also correctly exclude in most cases homogeneous and sky regions. These examples show the benefit of confidence estimation for high quality image alignment, useful \eg in multi-frame super resolution~\cite{WronskiGEKKLLM19}. Texture or style transfer (\eg for AR) also largely benefit from it. 

\begin{figure*}
\centering%
\vspace{-10mm}\includegraphics[width=0.9\textwidth]{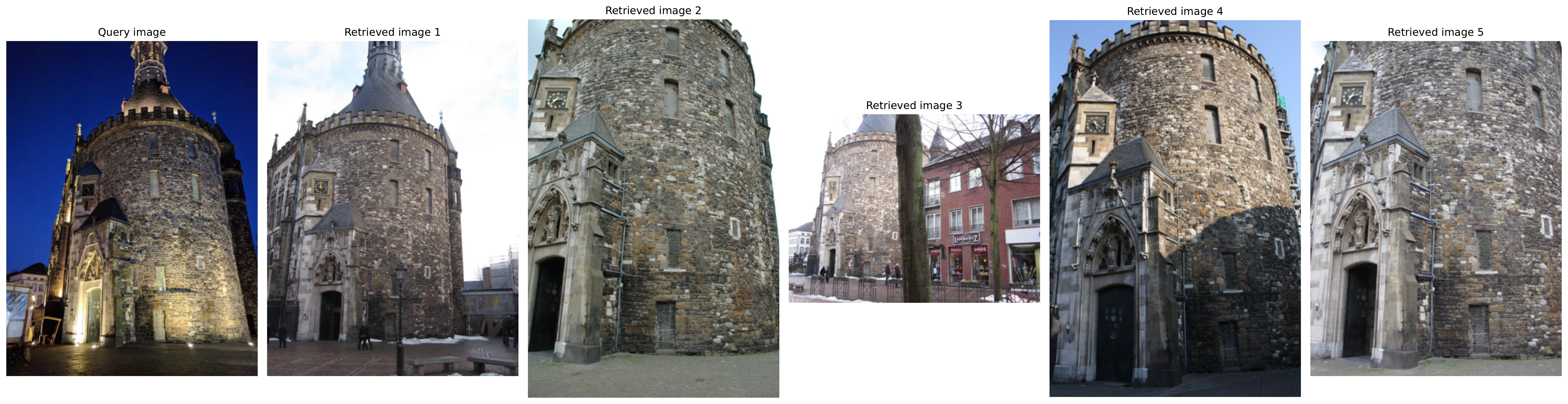} \\
\includegraphics[width=0.9\textwidth]{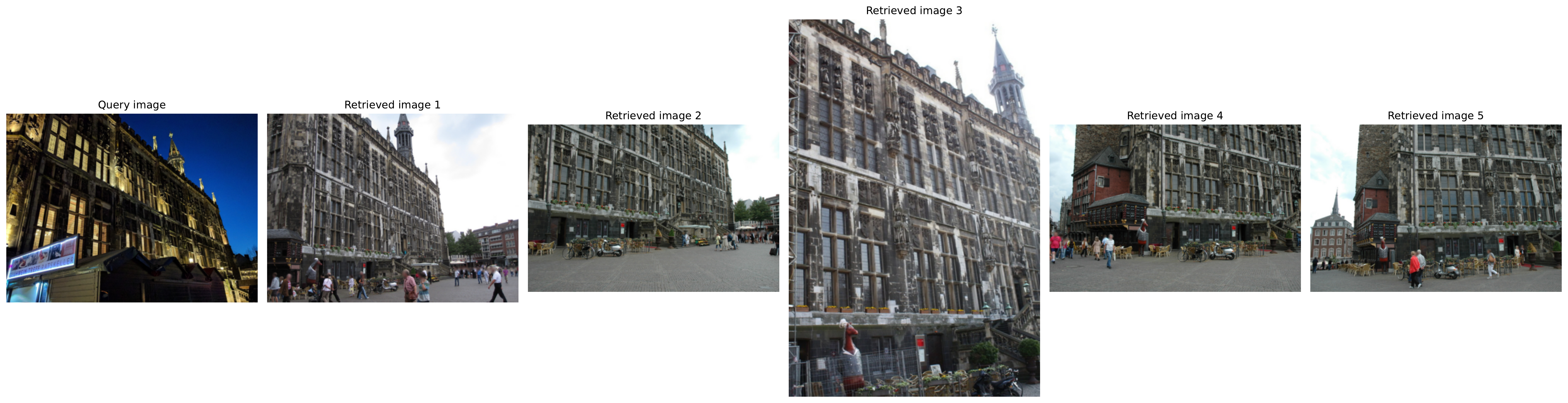} \\
\includegraphics[width=0.9\textwidth]{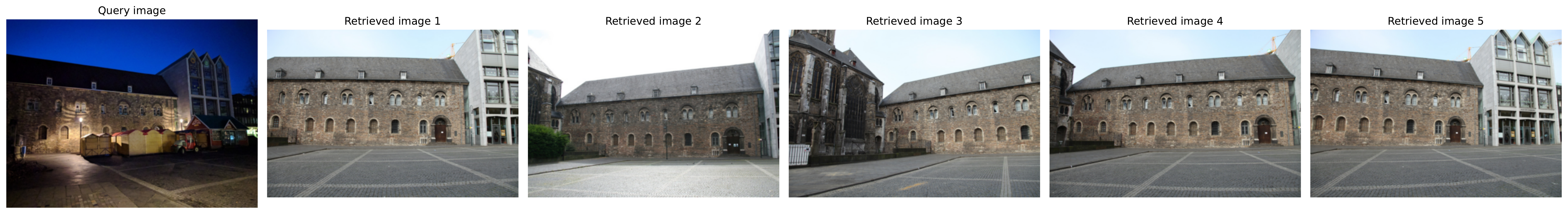} \\
\includegraphics[width=0.9\textwidth]{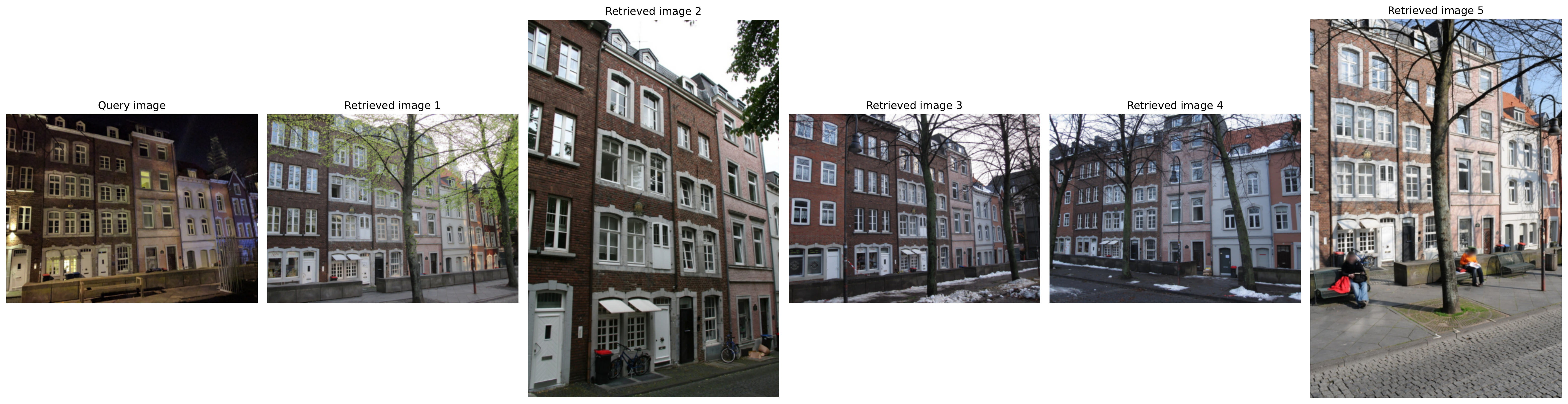} \\
\includegraphics[width=0.9\textwidth]{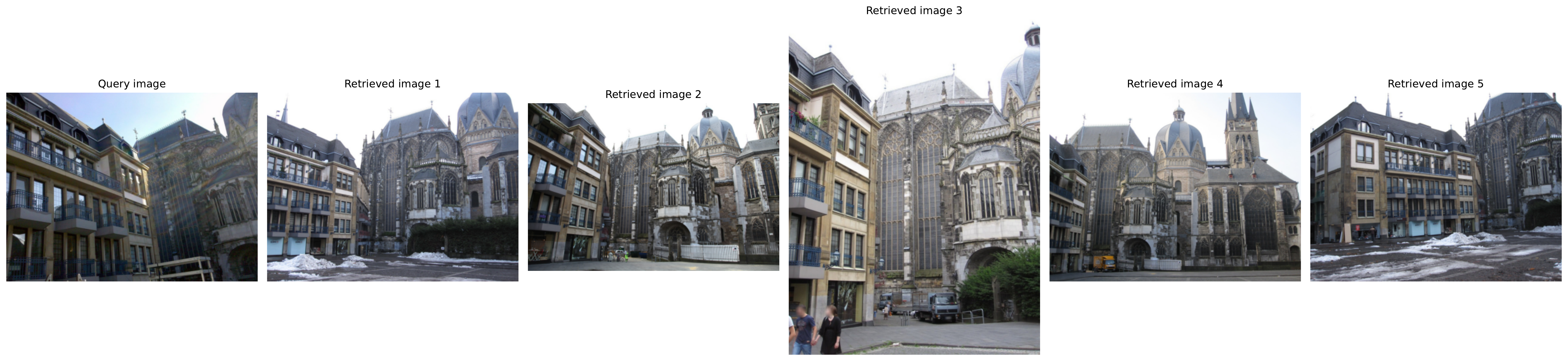} \\
\includegraphics[width=0.9\textwidth]{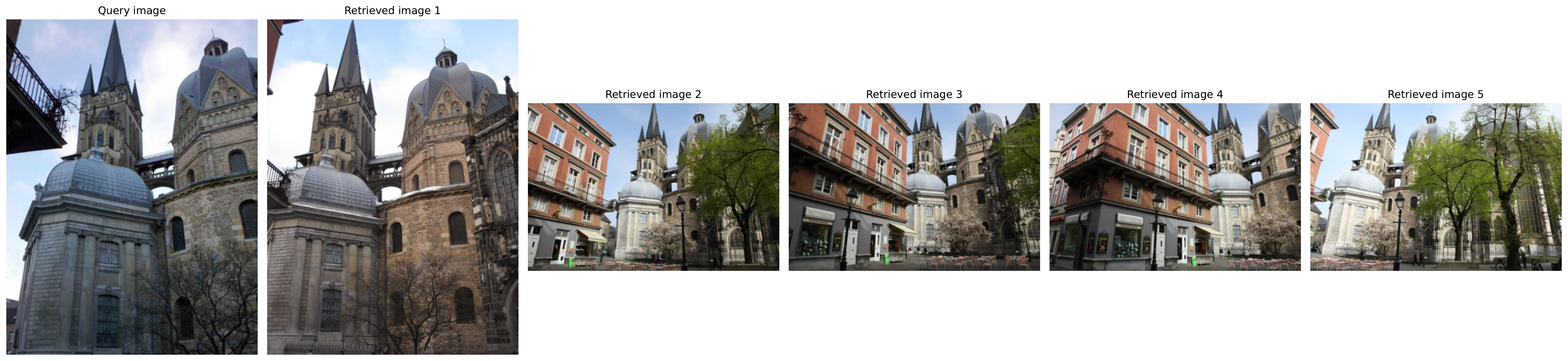}
\includegraphics[width=0.9\textwidth]{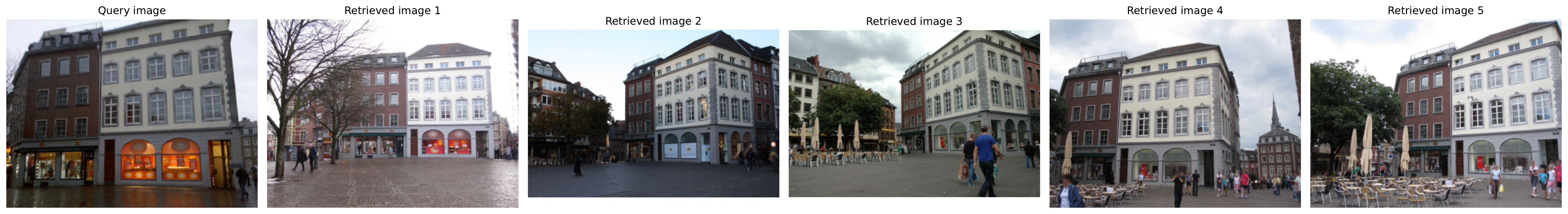}
\vspace{-2mm}
\caption{Retrieved images by PDC-Net+ for multiple night or day queries of the Aachen dataset.}
\vspace{-10mm}
\label{fig:aachen-retrieved}
\end{figure*}

\parsection{Image retrieval on Aachen} In Fig.~\ref{fig:aachen-retrieved}, we present the five closest retrieved images according to PDC-Net+, for multiple day and night queries of the Aachen dataset. Since we do not have access to the ground-truth query poses, we cannot verify which retrieved image is the correct closest to the query. However, for all examples, the retrieved images look very similar to the queries, even for difficult night examples.

\begin{table}[b]
\centering
\caption{Metrics evaluated over the scenes of HP-240. High PCK and low AEPE are better . }\label{tab:hp-details}
\resizebox{0.48\textwidth}{!}{%
\begin{tabular}{lcccccc}
\toprule
& I & II & III & IV & V & all \\ \midrule
& \multicolumn{6}{c}{\textbf{AEPE}}  \\ \midrule
DGC-Net & 1.74&5.88&9.07&12.14&16.5&9.07 \\
GLU-Net &0.59&4.05&7.64&9.82&14.89&7.4 \\
GLU-Net-GOCor & 0.78& 3.63& 7.64& 9.06 &  11.98&6.62 \\
LIFE &0.83 &  3.1&4.59&4.92&8.09&4.3  \\
RANSAC-Flow (MS) &0.52&\textbf{2.13}&4.83&5.13&6.36&3.79 \\
GLU-Net-GOCor* &0.74&3.33&6.02&7.66&7.57&5.06 \\
PDC-Net (H)&0.32&2.56&5.98&5.7&7.01&4.32 \\
PDC-Net (MS)&0.35&2.57&\textbf{3.02}&\textbf{5.06}&6.82&\textbf{3.56} \\
\textbf{PDC-Net+ (H)} & \textbf{0.33}&2.66&4.35&6.69&7.43&4.29 \\
\textbf{PDC-Net+ (MS)} & 0.36&2.67&3.29&5.51&\textbf{6.1}&3.59 \\ \midrule
& \multicolumn{6}{c}{\textbf{PCK-1} (\%)}  \\ \midrule
DGC-Net
&70.29&53.97&52.06&41.02&32.74&50.01 \\
GLU-Net
&87.89&67.49&62.31&47.76&34.14&59.92 \\
GLU-Net-GOCor
&84.72&64.43&60.07&48.87&34.12&58.45 \\
LIFE
&79.05&64.42&61.95&54.29&47.07&61.36 \\
RANSAC-Flow (MS)
&88.24&80.27&79.33&74.64&69.63&78.42 \\
GLU-Net-GOCor*
&82.81&68.16&64.98&58.5&49.97&64.8\\
PDC-Net (H)
&96.15&90.41&84.72&81.9&76.65&85.97 \\
PDC-Net (MS)
&95.91&\textbf{90.78}&88.95&83.78&77.6&87.4 \\
\textbf{PDC-Net+ (H)}
&95.87&90.75&87.89&82.16&76.14&86.56 \\
textbf{PDC-Net+ (MS)}
&\textbf{96.28}&90.44&\textbf{89.05}&\textbf{84.06}&\textbf{79.32}&\textbf{87.83} \\
\midrule
& \multicolumn{6}{c}{\textbf{PCK-5} (\%)}  \\ \midrule
DGC-Net
&93.7&82.43&77.58&71.53&61.78&77.4 \\
GLU-Net
&99.14&92.39&85.87&78.1&61.84&83.47 \\
GLU-Net-GOCor
&98.49&92.25&87.23&81.&70.49&85.89 \\
LIFE
&98.75&94.27&92.12&89.55&85.&91.94 \\
RANSAC-Flow (MS)
&99.35 & 97.65&\textbf{96.34}&93.8&93.17&96.06 \\
GLU-Net-GOCor*
&98.97&94.13&88.19&86.59&83.3&90.24 \\
PDC-Net (H)
&\textbf{99.91}&\textbf{97.99}&91.39&92.56&91.11&94.59 \\
PDC-Net (MS)
&99.82&97.98&96.24&94.44&92.44&96.18 \\
\textbf{PDC-Net+ (H)}
&99.84&97.94&94.73&91.83&90.52&94.97 \\
\textbf{PDC-Net+ (MS)}
&99.8&97.92&96.22&\textbf{94.46}&\textbf{93.38}&\textbf{96.36} \\
\bottomrule
\end{tabular}%
}
\end{table}

\section{Additional ablation study}
\label{sec-sup:ablation}

Finally, we provide additional ablative experiments. As in Sec.~\ref{subsec:ablation-study} of the main paper, we use BaseNet as base network to create the probabilistic models, as described in Sec.~\ref{sec-sub:arch-details}. Similarly, all networks are trained on solely the first training stage of the initial PDC-Net~\cite{pdcnet}. 

\parsection{Number of components of the constrained mixture} We compare $M=2$ and $M=3$ Laplace components used in the constrained mixture in Tab.~\ref{tab:ablation-sup}. In the case of $M=3$, the first two components are set similarly to the case $M=2$, \ie as $\sigma^2_1=1.0$ and $2.0 \leq \sigma^2_2 \leq \beta_2^+=s^2$ where $\beta_2^+$ is fixed to the image size used during training (256 here). The third component is set as $\sigma^2_3 = \beta_3^+ = \beta_3^- = \beta_2^+$. The aim of this third component is to identify outliers (such as out-of-view pixels) more clearly. The 3 components approach obtains a better Fl value on KITTI-2015 and slightly better pose estimation results on YFCC100M. However, its performance on MegaDepth and in terms of pure uncertainty estimation (AUSE) slightly degrade. As a result, for simplicity we adopted the version with $M=2$ Laplace components. Note, however, that more components could easily be added.

\end{document}